\documentclass[preprint,12pt]{elsarticle}

\usepackage{amssymb}
\usepackage{amsmath}
\usepackage{algorithm}%
\usepackage{algorithmicx}%
\usepackage{algpseudocode}
\floatname{algorithm}{Algorithm}

\usepackage{tabularx} % 用于 tabularx 环境
\usepackage{booktabs} % 可选，提供更美观的表格线
\usepackage{graphicx} % 插入图表
\usepackage{float}    % 控制图片位置H
\usepackage{microtype} % 自动化排版
\usepackage{subcaption} % 引入子图标题宏包
\usepackage{microtype} % 自动处理换行问题，避免文字溢出

\usepackage{xcolor}
\usepackage{hyperref}
\definecolor{ccr}{RGB}{33,150,209}
\hypersetup{
	hypertex=true,
	colorlinks=true,
	linkcolor=ccr,
	anchorcolor=ccr,
	citecolor=ccr
}
\begin{document}
\begin{sloppypar}
\begin{frontmatter}
	


\title{MRG: A Multi-Robot Manufacturing Digital Scene Generation Method Using Multi-Instance Point Cloud Registration}


\author[1]{Songjie Han} %% Author name
\author[1]{Yinhua Liu\corref{cor1}}  % 标注通讯作者
\ead{liuyinhua@usst.edu.cn}
\author[1]{Yanzheng Li}
\author[2]{Hua Chen}
\author[3]{Dongmei Yang}

\cortext[cor1]{Corresponding author.}
% \ead{liuyinhua@usst.edu.cn}

\affiliation[1]{organization={School of Mechanical Engineering},%Department and Organization
	addressline={University of Shanghai for Science and Technology}, 
	city={Shanghai},
	postcode={200093}, 
	country={China}}
\affiliation[2]{organization={Sino-German College of Intelligent Manufacturing},%Department and Organization
	addressline={Shenzhen Technology University}, 
	city={Shenzhen},
	postcode={518118}, 
	country={China}}
\affiliation[3]{organization={SAIC General Motors Corporation},%Department and Organization
	city={Shanghai},
	postcode={201208}, 
	country={China}}

%% Abstract
\begin{abstract}
	A high-fidelity digital simulation environment is crucial for accurately replicating physical operational processes. However, inconsistencies between simulation and physical environments result in low confidence in simulation outcomes, limiting their effectiveness in guiding real-world production. Unlike the traditional step-by-step point cloud "segmentation-registration" generation method, this paper introduces, for the first time, a novel Multi-Robot Manufacturing Digital Scene Generation (MRG) method that leverages multi-instance point cloud registration, specifically within manufacturing scenes. Tailored to the characteristics of industrial robots and manufacturing settings, an instance-focused transformer module is developed to delineate instance boundaries and capture correlations between local regions. Additionally, a hypothesis generation module is proposed to extract target instances while preserving key features. Finally, an efficient screening and optimization algorithm is designed to refine the final registration results. Experimental evaluations on the Scan2CAD and Welding-Station datasets demonstrate that: (1) the proposed method outperforms existing multi-instance point cloud registration techniques; (2) compared to state-of-the-art methods, the Scan2CAD dataset achieves improvements in MR and MP by 12.15\% and 17.79\%, respectively; and (3) on the Welding-Station dataset, MR and MP are enhanced by 16.95\% and 24.15\%, respectively. This work marks the first application of multi-instance point cloud registration in manufacturing scenes, significantly advancing the precision and reliability of digital simulation environments for industrial applications.
\end{abstract}



%% Keywords
\end{frontmatter}



\section{Introduction}
\label{sec1}
In recent years, digital twin technology has experienced widespread adoption across modern manufacturing sectors, particularly in the areas of welding, assembly, and quality inspection \cite{bib1}. Through the construction of high-fidelity digital models, this technology facilitates the precise simulation of real-world production environments, thereby optimizing manufacturing processes and forecasting potential failures \cite{bib2}. A large field-of-view three-dimensional (3D) scanner is employed to acquire high-quality ambient point clouds, which serve as the foundation for generating a highly reliable digital simulation environment. The generation process involves several stages: 1) utilizing a 3D scanner to capture the spatial configuration of the industrial robot’s working shops; 2) preprocessing point cloud data to eliminate noise and enhance data quality; 3) segmenting the industrial robot’s point cloud data; and 4) employing point cloud registration technology to align the corresponding robot’s digital model with the spatial pose of the point cloud data \cite{bib3}.

To facilitate high-precision simulations within digital industrial manufacturing environments, point cloud registration technology has become indispensable. Its objective is to estimate the transformation relationships between point clouds from different locations and to align them spatially with high precision. Specifically, this involves identifying and establishing spatial mappings between two point clouds, thereby integrating their respective coordinate systems. This technology constitutes a critical research domain within industrial big data analysis and has found extensive applications in reverse engineering \cite{bib4}, simultaneous localization and mapping \cite{bib5}, pose estimation \cite{bib6}, and intelligent manufacturing \cite{bib7}.

Over the past few decades, researchers have developed a variety of effective point cloud registration techniques. Traditional geometric methods include the Iterative Closest Point (ICP) algorithm \cite{bib8}. Fast Global Registration (FGR) \cite{bib9}, the Normal Distribution Transform (NDT) \cite{bib10,bib11}, and their variants are also prominent. In recent years, deep learning (DL)-based methods, such as PointNetLK \cite{bib12}, CoFiNet \cite{bib13}, and GeoTransformer \cite{bib14}, have been increasingly proposed. However, the step-by-step method for generating digital industrial manufacturing simulation environments has the following drawbacks: 1) As a crucial front-end step in point cloud registration, segmentation accuracy significantly impacts scene generation quality, and its robustness is inadequate; 2) The step-by-step method results in prolonged generation times;
\begin{figure}[H]
	\centering
	\includegraphics[width=\textwidth]{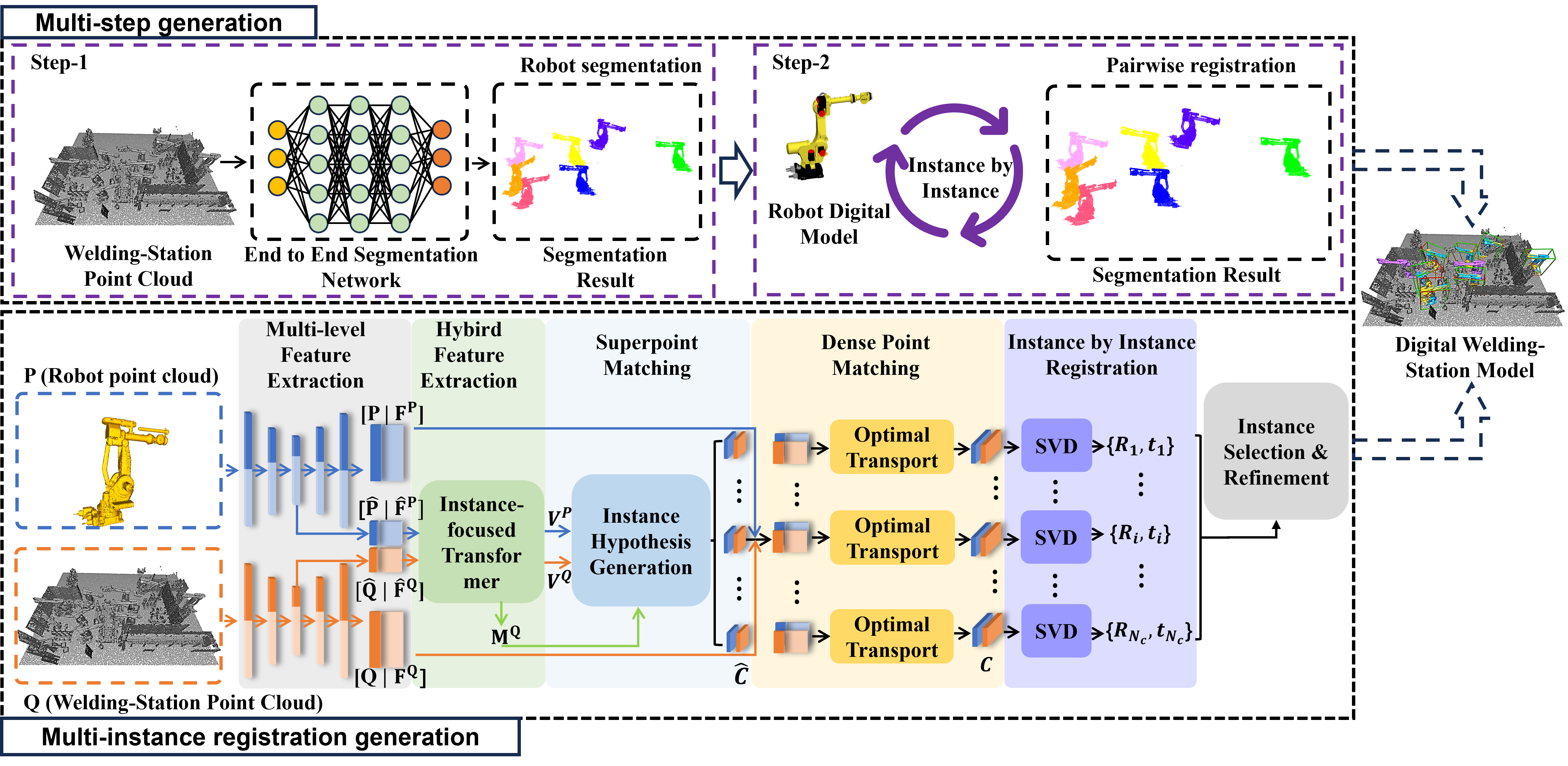}
	\caption{Comparison of different methods for generating digital welding station. The target instance point clouds in the station are shown in blue, the source point cloud after the transformation estimate is shown in yellow, point clouds similar to the target instances are shown in purple, and outlier points are shown in gray. The green bounding boxes represent the actual poses of the instances in the target point clouds, and the red bounding boxes represent the predicted poses. Instances surrounded by both red and green bounding boxes indicate successful detection, while those surrounded only by green bounding boxes indicate missed detection.}
	\label{fig:comparison_different_generation}
\end{figure}

In view of the aforementioned challenges, step-by-step generation methods struggle to meet the accuracy and efficiency requirements of digital manufacturing scene generation. With the rapid advancement of 3D scanning technologies, such as LiDAR and RGB-D cameras, multi-instance point cloud registration methods have gradually become a research focus. Unlike paired point cloud registration methods, multi-instance point cloud registration methods require the identification of corresponding point clusters belonging to different instances and the estimation of multiple transformation relationships. Fig. \ref{fig:comparison_different_generation} illustrates the differences between the two digital simulation environment generation methods. Multi-instance point cloud registration methods are primarily categorized into multi-model fitting methods and DL-based methods. The fundamental concept of multi-model fitting methods \cite{bib15, bib16, bib17} is to sample a series of hypotheses and then subsequently conduct preference analysis or consensus analysis. These methods rely on efficient hypothesis sampling; however, in industrial manufacturing scenes with dense point clouds and a high number of outliers, the efficiency and robustness of the algorithms can be significantly reduced. To this end, researchers leverage DL algorithms such as Predator \cite{bib18} for point cloud feature extraction and combine them with the robust representation capabilities of a distance consistency matrix to propose a DL-based multi-instance point cloud registration method \cite{bib19}. The impact of outliers on the extraction of instance correspondences is effectively reduced, and the efficiency of instance correspondence point clustering is improved. However, the DL-based method also has limitations when applied to registration tasks in digital manufacturing scenes. For example, these methods often overlook the correlations between local areas of industrial robots, resulting in an insufficient mapping accuracy.

To address the aforementioned research gap, we propose a DL-based multi-instance point cloud registration method, Multi-Robot Registration (MRG), which aims to solve the model generation challenges in digital manufacturing environments. This method considers the sensing range of point clouds in different regions and integrates a coarse-to-fine mapping extraction strategy to substantially enhance the accuracy of instance extraction. Additionally, the interconnections among various local regions of the industrial robot are thoroughly taken into account, thereby enhancing registration accuracy. The specific contributions of this paper are outlined as follows:
\begin{enumerate}
	\item[1)] We propose a novel multi-instance point cloud registration method that directly extracts instance correspondences and estimates transformation parameters without relying on multi-model fitting. 
	\item[2)] We have developed a novel geometric transformer module that, while limiting context features to the instance range, extends the receptive field of the point cloud and enhances connections between point clouds in different local areas. 
	\item[3)] We introduce a novel hypothesis generation and optimization strategy that eliminates redundant candidates while retaining key features of the robot, thereby generating the final registration result. 
\end{enumerate}

The remainder of this paper is organized as follows: Section \ref{sec2} reviews the relevant research progress in point cloud registration and multi-instance point cloud registration; Section \ref{sec3} introduces the general framework of MRG. Section \ref{sec4} evaluates the performance of the proposed method and presents the experimental results. Section \ref{sec5} summarizes the research presented in this paper.

%------------------------------Related works--------------------------------
\section{Related works}
\label{sec2}
Based on the number of objects in the scene and the registration requirements, point cloud registration algorithms can be classified into pairwise point cloud registration methods and multi-instance registration methods. The latter can be further subdivided into multi-model fitting methods and DL-based methods.
\subsection{Pairwise Point Cloud Registration}
\label{subsec1}
Point cloud registration has a long history, and most current methods focus on pairwise registration. These methods are generally divided into three sub-tasks: point matching, outlier rejection, and transformation estimation. Traditional point matching methods rely on manually designed descriptors \cite{bib20, bib21}, to capture local information; however, these methods are sensitive to noise and outliers. To enhance the robustness of matching algorithms, DL-based methods have gradually replaced traditional methods. Zeng et al. \cite{bib22} proposed the pioneering 3DMatch algorithm, which takes local voxel blocks as input and employs a 3D Convolutional Neural Network (CNN) to learn local geometric features, thereby generating robust and highly discriminative 3D descriptors. Deng et al. \cite{bib23} proposed an unordered network architecture based on PointNet \cite{bib24} that utilizes a novel N-tuple loss function to inject global contextual information into local descriptors, thereby enhancing their representation capability. Gojcic et al. \cite{bib25} aligned Smoothed Density Value (SDV) with the Local Reference Frame (LRF) to address the rotation invariance issue of descriptors. They used a twin deep learning architecture for efficient point cloud matching.

Accurate correspondences are essential for enhancing the accuracy and efficiency of pose estimation, making a robust outlier elimination module crucial. Random Sample Consensus (RANSAC) \cite{bib26} and its variants \cite{bib27, bib28} are widely regarded as the most stable traditional outlier elimination methods. These algorithms fit the model by randomly selecting data subsets. The model assesses data consistency to distinguish inliers from outliers. Deep learning methods, such as SACF-Net \cite{bib29} and DGR \cite{bib30}, treat outlier elimination as a binary classification task, outputting confidence scores by constructing the largest possible group based on spatial consistency. After obtaining accurate correspondences, the microweighted Procrustes method \cite{bib31}, based on Singular Value Decomposition (SVD), is typically used for the final pose transformation. With the advancement of end-to-end architectures, several models \cite{bib11, bib32} can directly output the pose transformation matrix.

Although the aforementioned pairwise registration methods perform well in many scenes, their effectiveness diminishes in digital manufacturing scene generation for multi-instance industrial robots. In these scenes, identifying all transformation matrices that achieve the necessary pose estimation accuracy becomes challenging.
\subsection{Multi-model Fitting}
\label{subsec2}
Multi-model fitting methods aim to fit multiple models from noisy data, such as fitting multiple planes in a point cloud, estimating a fundamental matrix in motion segmentation, or determining a rigid transformation matrix in multi-instance point cloud registration. Existing multi-model fitting methods are categorized into RANSAC-based and cluster-based methods. RANSAC-based methods \cite{bib16, bib32} primarily adopt the hypothesis-verification method for sequential fitting of multiple models. For instance, Sequential RANSAC \cite{bib32} detects instances by repeatedly running RANSAC to recover a single instance and subsequently removing its inlier points from the input. CONSAC \cite{bib16} was the first to introduce the DL model into multi-model fitting, utilizing a network similar to PointNet \cite{bib24} to guide sampling. However, efficiency significantly decreases with large-scale inputs. Cluster-based methods \cite{bib15, bib17} sample numerous hypotheses and group input points based on residuals under these hypotheses. For example, RansaCov \cite{bib15} transforms the multi-model fitting problem into a maximum coverage problem and provides two approximate solving strategies. T-linkage \cite{bib17} initializes a broad range of hypotheses through point sampling and preference vector clustering to remove outliers.
\subsection{Multi-instance Point Cloud Registration based on Deep Learning}
\label{subsec3}
Multi-instance registration methods based on deep learning require the estimation of multiple transformations for various instances in both the source and target point clouds. This process not only requires filtering outliers in the noisy correspondences but also clustering the remaining correspondences into individual instances. The current methods \cite{bib19, bib33} primarily focus on utilizing the depth representation of correspondences for clustering, followed by the iterative estimation of transformations for each individual instance. Yuan et al. \cite{bib33} proposed a multi-instance point cloud registration framework leveraging contrastive learning. Through contrastive learning, a discriminative representation of high-dimensional correspondence relationships was acquired. Finally, they clustered high-dimensional features using a spectral clustering algorithm with a specific pruning strategy. However, the learned high-dimensional features have a limited impact on the overall results, and the clustering process of these features is time-consuming. In addition to employing contrastive learning, Tang et al. \cite{bib19} directly groups the noisy correspondence sets into distinct clusters using a distance-invariant matrix, based on the global spatial consistency of the point cloud’s rigid transformation \cite{bib34}. However, when multiple instances are involved, the reliability of the distance-invariant matrix is compromised due to dense noisy correspondences and similarity between outliers and inlier points.

In summary, existing registration methods have made significant progress in addressing the challenges of paired registration and multi-instance registration. However, these methods are not specifically designed for generating of digital industrial manufacturing scenes and often overlook the consideration of local, fine-grained characteristics among industrial robots. Therefore, this paper primarily aims to propose a multi-instance point cloud registration framework specifically designed for generating digital models of robots in industrial manufacturing scenes. The framework is designed to eliminate noise interference, investigate the correlation between local regions of the robot, and enhance the accuracy of pose estimation.
%% Use \subsection commands to start a subsection.
%\subsection{Example Subsection}
%\label{subsec1}
%% Use \subsubsection, \paragraph, \subparagraph commands to 
%% start 3rd, 4th and 5th level sections.
%% Refer following link for more details.
%% https://en.wikibooks.org/wiki/LaTeX/Document_Structure#Sectioning_commands
\section{Method}
\label{sec3}
Given the source point cloud $P=\{p_i\in\mathbb{R}^3|i=1,...,N\}$ and the target point cloud $Q=\{q_i\in\mathbb{R}^3|i=1,...,M\}$. The source point cloud contains one instance of the 3D model, and the target point cloud contains $J$ instances of the same model. The purpose of multi-instance point cloud registration is to recover $J$ rigid transformations from two point clouds: $\{R_j\in SO(3),t_j\in\mathbb{R}^3\}_{j=1}^J$. Given the number of instances $J$ in the target point cloud and the predicted value of the point set $C$, the process of solving the rigid transformations can be formulated as an optimization problem.
\begin{equation}
	\label{equation_1}
	\min_{\left\{R_j,t_j\right\}_{j=1}^J}\frac{1}{J}\sum_{j=1}^J\sum_{(p_{j,i},q_{j,i})\in C_j^{gt}}\frac{1}{\left|C_j^{gt}\right|}\left\|q_{j,i}-R_jp_{j,i}-t_j\right\|^2
\end{equation}
where \( C_j^{\text{gt}} \) represents the true inlier set of the \( j \)-th instance, and \( \left| C_j^{\text{gt}} \right| \) represents the number of inliers.
To illustrate the proposed methodology, the remainder of this section is organized as follows: an overview of the proposed architecture is described first (Section \ref{method_subsec1}). Subsequently, the four key modules of the architecture are described in detail: the instance-focused transformer module (Section \ref{method_subsec2}), the instance hypothesis generation module (Section \ref{method_subsec3}), the instance filtering and optimization module (Section \ref{method_subsec4}), and the loss function module (Section \ref{method_subsec5}).
\subsection{Overview}
\label{method_subsec1}
\begin{figure}[htbp]
	\centering
	\includegraphics[width=\textwidth]{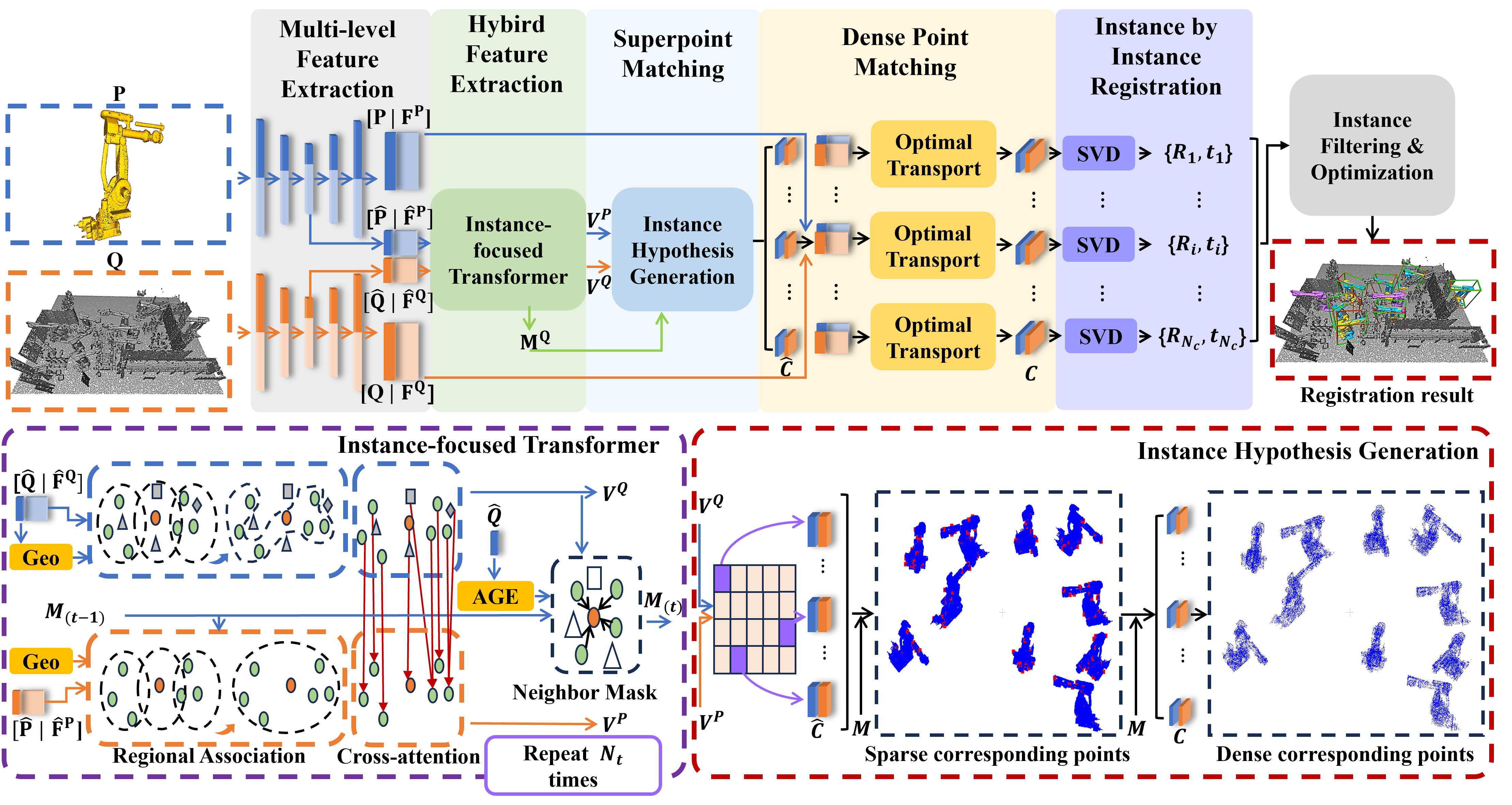}
	\caption{The pipeline of the proposed MRG for multi-instance point cloud registration. It takes putative correspondences and the original point cloud as input, and outputs $N_{c}$ rigid transformations.}
	\label{fig:pipeline}
\end{figure}
MRG employs a coarse-to-fine strategy \cite{bib13} based on a keypoint-free registration method for extracting correspondences. The comprehensive process of our method is illustrated in Fig. \ref{fig:pipeline}. To enhance the computational efficiency and feature representation for scene-level point clouds, we use the backbone network \cite{bib35} to progressively downsample the two point clouds \( P \) (source point cloud) and \(Q \) (target point cloud), as well as their downsampled counterparts \(\hat{P} \) and \( \hat{Q} \), and extract multi-level features represented by \( F^{P} \), \( F^{Q} \), \( \hat{F}^{P} \), and \( \hat{F}^{Q} \). At the coarse level, the instance-focused transformer module processes the multi-level features of the source and target point clouds, yielding highly discriminative fused features and a neighbor mask matrix. Considering the complex geometric structures of instances in industrial manufacturing scenes, MRG uses an instance hypothesis generation module to process the fused features and the neighbor mask matrix, generating uniformly distributed coarse correspondences on instance surfaces. At the fine level, accurate coarse correspondences are extended to dense correspondences by incorporating the neighbor mask matrix. Finally, a simple and effective instance filtering and optimization algorithm generates the final registrations.
\subsection{Instance-Focused Transformer}
\label{method_subsec2}
\begin{figure}[htbp]
	\centering
	\includegraphics[width=\textwidth]{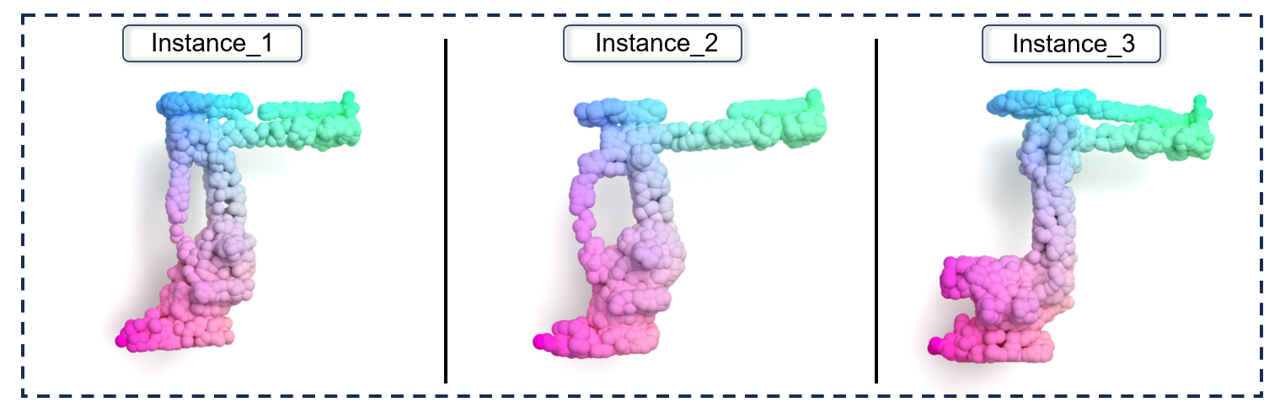}
	\caption{Schematic characterization of different robot instances.}
	\label{fig:fine_grained}
\end{figure}

\begin{figure}[htbp]
	\centering
	\includegraphics[width=\textwidth]{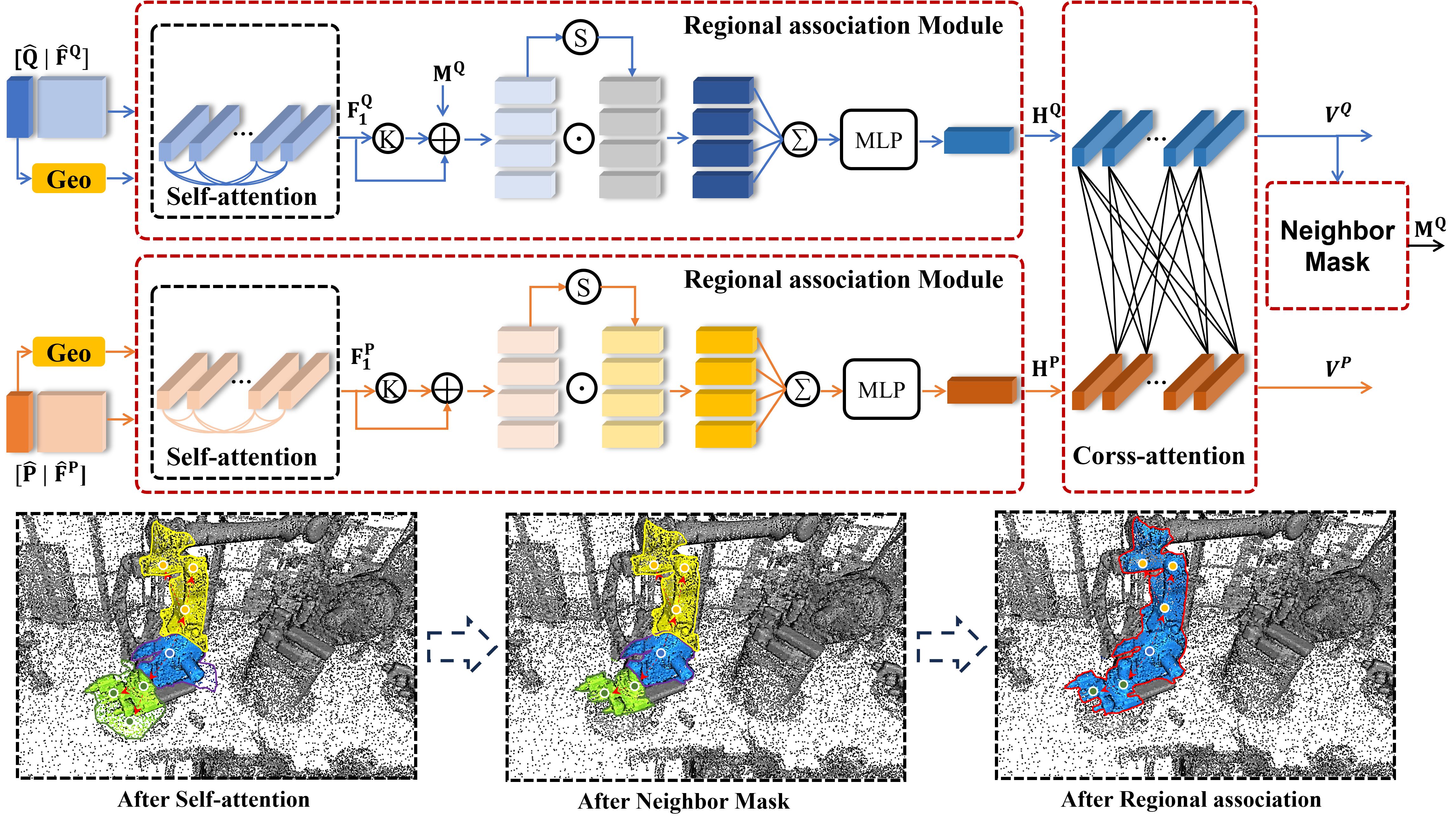}
	\caption{Structure of the Instance-Focused module.}
	\label{fig:instance_focused_transformer}
\end{figure}

Accurate correspondence is essential for high-precision pose estimation. The feature similarity matrix represents the correlation between source and target point clouds. Transformers have been demonstrated to effectively capture contextual information within individual point clouds while facilitating cross-feature fusion between paired point clouds \cite{bib36}. However, in industrial manufacturing, target instance features are susceptible to contamination, and robotic joint features often exhibit significant similarity. Consequently, the feature similarity matrix may become unreliable, as illustrated in Fig. \ref{fig:fine_grained}. To address this, we propose a novel instance-focused transformer module. This module restricts the contextual encoding of target instance point clouds to intra-instance information and enhances associations between different local structures within each instance. The specific structural details of this module are illustrated in Fig. \ref{fig:instance_focused_transformer}.

\textbf{Regional association module.} The instance-focused transformer module comprises three key components: a regional association module, a cross-attention module, and a neighbor mask module. Given a superpoint $\hat{q}_i \in \hat{Q}$, its $K$-nearest neighbors are represented as: $\hat{N}_i^Q = \{\hat{q}_{i,1}, \hat{q}_{i,2}, \hat{q}_{i,3}, \dots, \hat{q}_{i,k}\}$. The input feature matrix is $\hat{F}^Q \in \mathbb{R}^{|\hat{Q}| \times d}$, and the neighbor mask matrix is $M^Q \in \mathbb{R}^{|\hat{Q}| \times k}$. Here, $|\cdot|$ represents the number of elements in the set, and $d$ represents the feature dimension. The regional association module outputs $H^Q \in \mathbb{R}^{|\hat{Q}| \times d}$, with elements $h_i^Q \in H^Q$ computed as follows:
\begin{equation}
	\label{attention_scores}
	e_{i,j}=\frac{\left(\hat{f}_i^QW^Q\right)\left(\hat{f}_{i,j}^QW^K+g_{i,j}W^R\right)^T}{\sqrt{d}}+m_{i,j}^Q
\end{equation}
\begin{equation}
	f_1^Q=\sum_{j=1}^k\frac{\exp(e_{i,j})}{\sum_{l=1}^k\exp(e_{i,l})}(\hat{f}_{i,j}^QW^V)
\end{equation}
\begin{equation}
	h_i^Q=MLP_1(sum\left(\sum_{i=1}^k\left(f_2^Q\odot softmax(f_2^Q)\right)\right))
\end{equation}
where $e_{i,j}$ represents the attention score between the superpoint and its neighboring point, $W^Q$, $W^K$, $W^V$, and $W^R \in \mathbb{R}^{d \times d}$ represent the projection weights for the query, key, value, and geometric embedding, respectively, and $g_{i,j} \in \mathbb{R}^{1 \times d}$ represents the geometric structure of the embedding \cite{bib14}. $m_{i,j}^Q$ represents the correlation score between superpoint $\hat{q}_i$ and its neighboring point $\hat{q}_{i,j}$, where $m_{i,j}^Q$ = 0 if $\hat{q}_{i}$ and $\hat{q_{i,j}}$ belong to the same instance, otherwise, $m_{i,j}^Q$ = $-\infty$. $f_1^Q\in\mathbb{R}^{1\times d}$ represents the output features, $f_2^Q\in\mathbb{R}^{1\times k\times2d}$ represents the fusion features filtered by $M^Q$, computed as shown below:
\begin{equation}
	f_2^Q=concat(repeat(f_1^Q,k),f_3^Q)
\end{equation}
where $f_3^Q\in\mathbb{R}^{1\times k\times d}$ represents the $K$-nearest neighbor feature filtered by $M^Q$, $concat(\cdot)$ represents an aggregate operation, and $repeat(\cdot)$ represents a repeat operation. For the source point cloud $P$, we omit the mask term calculation in Equation \ref{attention_scores}.

$\textbf{Cross-attention module}$. Inspired by \cite{bib13, bib14}, we capture the correlations between the source and target point clouds using a cross-attention module subsequent to encoding the internal geometric context of each instance. Given input features $H^P$ and $H^Q$, the cross-attention module outputs $V^P\in\mathbb{R}^{|P|\times d}$. The calculation for $v_i^P\in V^P$ is shown as follows:
\begin{equation}
	v_i^P=\sum_{j=1}^{|\widehat{Q}|}\frac{\exp(e_{i,j})}{\sum_{k=1}^{|\widehat{Q}|}\exp(e_{i,k})}(h_k^QW^V)
\end{equation}
where the attention score $e_{i,j}$ is calculated as follows:
\begin{equation}
	e_{i,j}=\frac{\left(h_i^PW^Q\right)\left(h_i^QW^K\right)^T}{\sqrt{d}}
\end{equation}
where $W^Q$, $W^K$, and $W^V\in\mathbb{R}^{d \times d}$ are the respective projection weights for the query, key and value. The same computation applies for $V^Q$. Benefiting from the cross-attention block, the superpoint features in one point cloud become aware of the geometric structure of the other, thereby facilitating the modeling of geometric consistency between two point clouds.

$\textbf{Neighbor mask module}$. Finally, we propose a neighbor mask module based on the self-attention mechanism to mitigate interference from noisy instances in the target point cloud. By dynamically adjusting the attention weights, the module effectively suppresses the influence of irrelevant noisy neighbors. However, the enhancement of prediction accuracy is hindered by insufficient feature discrimination. Therefore, we introduce geometric features including normal vectors, curvature, and geodesic distances to enhance instance differentiation. Specifically, in complex scenes, normal vectors represent the directionality of local surfaces, curvature describes the bending of local structures, curvature describes the bending of local structures, and superpoints of different instances exhibit greater distances in geodesic space. Comprehensive utilization of these characteristics significantly enhances prediction accuracy.
%\begin{figure}[H]
%	\centering
%	\begin{minipage}[t]{0.5\textwidth}
%		\centering
%		\includegraphics[width=0.9\linewidth]{pictures/Figure5(a)_neighbor_mask.png} % 调整图片宽度
%		\caption*{(a)}
%	\end{minipage}%
%	\begin{minipage}[t]{0.5\textwidth}
%		\centering
%		\includegraphics[width=0.4\linewidth]{pictures/Figure5(b)_standard_attention.png} % 调整图片宽度
%		\caption*{(b)}
%	\end{minipage}
%	\caption{Structure of the Neighbor mask module. (a) is the Neighbor mask module, and (b) is the standard self-attention module.}
%	\label{fig:neighbor_mask_module}
%\end{figure}
\begin{figure}[htbp]
	\centering
	\includegraphics[width=\textwidth]{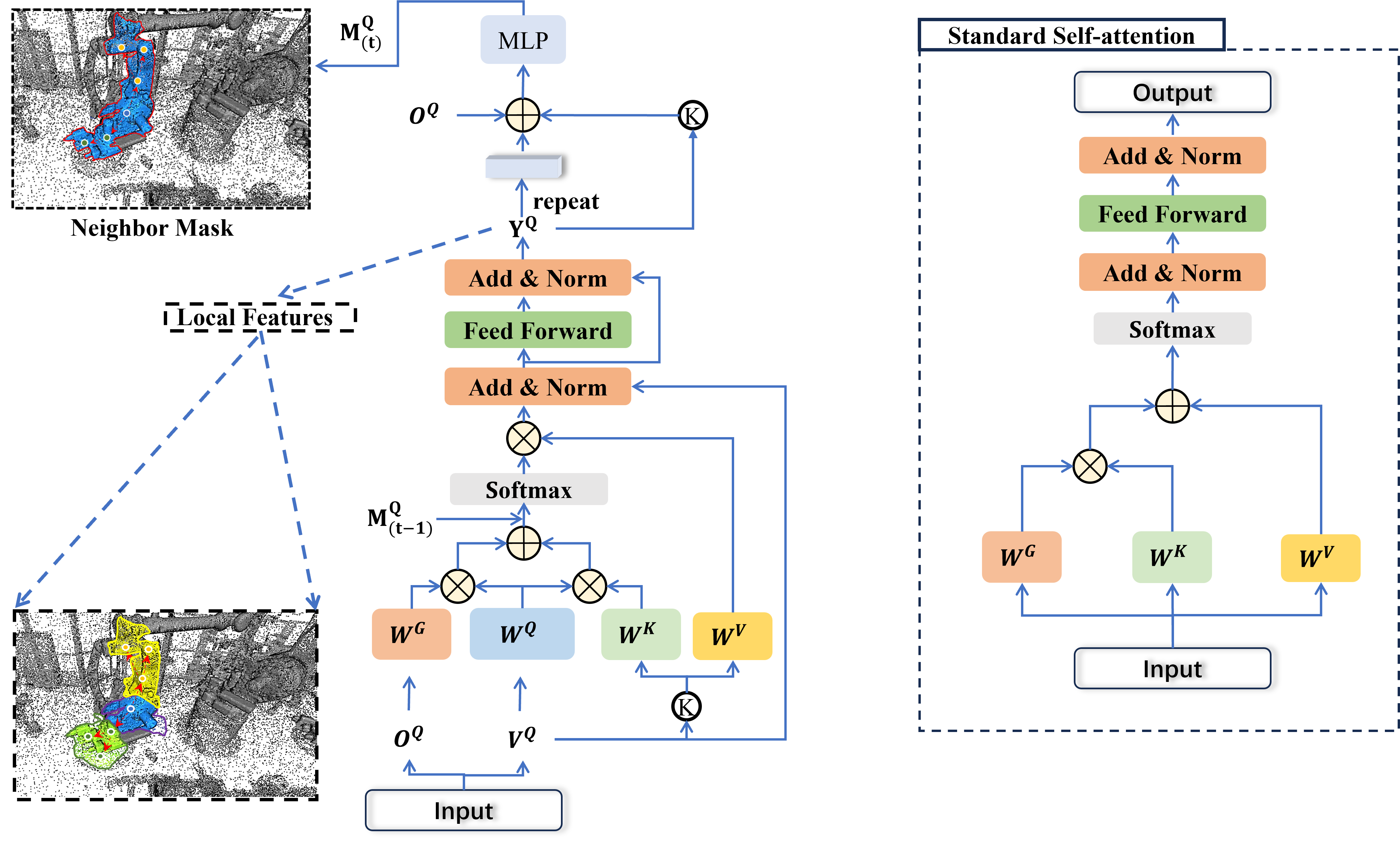}
	\caption{Structure of the Neighbor mask module. The left is the Neighbor mask module, and the right is the standard self-attention module.}
	\label{fig:neighbor_mask}
\end{figure}

The forward propagation schematic of the neighbor mask module is illustrated in Fig. \ref{fig:neighbor_mask}. The geometric structural embeddings in Equation \ref{attention_scores} are replaced with aggregated geometric features to enhance the discriminative capabilities among instances. Finally, we employ an \textit{MLP} to predict whether $\hat{q}_{i} \in \hat{Q}$ and its neighbor $\hat{q}_{i,j} \in \hat{N}_{i}^{Q}$ belong to the same instance. The confidence score $a_{i,j} \in A$ is calculated as follows:
\begin{equation}
	o_{i,j}^Y=MLP_2\left(cat(f_{normal},f_{curve},f_{gde})\right)
\end{equation}
\begin{equation}
	a_{i,j}=sigmoid\left(MLP_3\left(concat(\hat{y}_{i,j}-\hat{y}_i,o_{i,j}^Y)\right)\right)
\end{equation}
where $o_{i,j}^Y$ represents the embedded geometric features, while $f_{normal}$, $f_{curve}$, and $f_{gde}$ represent the normal vector, curvature, and geodesic features, respectively. Finally, the confidence score matrix $A \in \mathbb{R}^{|\hat{Q}| \times k}$ is transformed into $M_{(t)}^Q$ through a threshold function, as detailed below:
\begin{equation}
	m_{i,j}^Q =
	\begin{cases} 
		-\infty, & \text{if } a_{i,j} < \tau, \\ 
		0, & \text{otherwise}.
	\end{cases}
\end{equation}
where $\tau$ represents the confidence threshold. By utilizing instance masks, our model can effectively learn the overlapping contextual information within instances and extract precise correspondences that encompass more instances. Additionally, the neighbor mask module contributes to extracting dense point correspondences, as described in Section \ref{method_subsec3}.
\subsection{Instance Hypothesis Generation}
\label{method_subsec3}
After obtaining the combined source and target point cloud features, a feature similarity matrix is typically generated. Subsequently, the top $N_l$ pairs with the highest matching scores are selected to form sparse superpoint matches \cite{bib13, bib14}. However, in industrial manufacturing scenes, as instances and robot geometries become more complex, target point clouds with low matching degrees are often frequently overlooked. This oversight results in partial loss of instance point clouds and decreased pose estimation accuracy. 

To address this issue, this paper proposes an effective instance hypothesis generation module. By combining the neighbor mask matrix and feature similarity matrix, we restore the geometric structures of most target instances, even with low feature matching scores. Refer to Algorithm \ref{algo1} for detailed steps.
\renewcommand{\thealgorithm}{1:}
\begin{algorithm}[H]
	\caption{Instance Hypothesis Generation Module}
	\label{algo1}
	\begin{algorithmic}[1]
		\Require Hybrid features $(V^P, V^Q)$, neighborhood mask $M^Q$, positions $(\hat{P}, \hat{Q})$, $Top_k$ parameter $N_l$
		\Ensure Sparse correspondences $\hat{C}$, match scores $\hat{S}$
		\State Compute Euclidean distance $E \gets E(V^P, V^Q)$, and matching scores $S \gets \exp(-E)$
		\State Select top-$N_l$ correspondences $(\hat{C}_P, \hat{C}_Q) \gets Top_k(S, N_{l})$
		\State Initialize $T^Q \gets 0$, candidate list $L_Q \gets [\,]$
		\State Initialize candidate index list $D \gets [\,]$
		\For{$\hat{c}_i \in \hat{C}_P$ \textbf{where} $T^Q[\hat{c}_i] == 0$}
			\State Set $T^Q[\hat{c}_i] \gets 1$ and add $\hat{c}_i$ to $D$
			\While{$D$ is not empty}
				\State $d \gets D.pop()$
				\If{$\text{sum}(M_{\hat{c}_i}^Q[d]) == K$}
					\State Set $T^Q[d] \gets 1$, add $d$ to $L_Q$
					\State Add all neighbor indices of $d$ to $D$
				\EndIf
			\EndWhile
		\EndFor
		\State $L_Q \gets \text{Unique}(L_Q)$; $L_P \gets \arg\min(S[j])$ for $j \in \text{len}(L_Q)$
		\If{$\text{len}(L_Q) < N_l$}
			\State Based on $\hat{C}_P, \hat{C}_Q, S$, compute $\hat{S}$
			\State \textbf{return} $\hat{C} \gets (\hat{C}_P, \hat{C}_Q), \hat{S}$
		\EndIf
		\State \textbf{return} $\hat{C}, \hat{S} \gets \text{FS}(L_P, L_Q, S)$
	\end{algorithmic}
\end{algorithm}
where $E(\cdot)$ represents the Euclidean distance computation, $Unique(\cdot)$ performs element deduplication, and $FS(\cdot)$ represents the Farthest Point Sampling algorithm \cite{bib37}.

In the Instance Hypothesis Generation Module, feature differences between the source and target point clouds are calculated. The top $N_l$ pairs with the lowest scores are selected as reliable candidate correspondences. To capture a larger portion of the instance’s surface structure, a neighborhood mask matrix is utilized. By incorporating region growing, the candidate points are iteratively expanded. To enhance computational efficiency and reduce memory consumption, the Farthest Point Sampling (FPS) algorithm is applied to downsample the correspondences and scores.

Given the feature similarities among various robot models, sparse superpoint correspondences must be refined into dense point correspondences at a finer scale. This process facilitates the acquisition of richer geometric information, enhances diversity among robot instances, and improves both the accuracy and robustness of pose estimation. Previous methods \cite{bib13} employed an optimal transport layer to directly match points within the local regions of two corresponding point clouds. However, locally extracted correspondences originate from tightly clustered groups, potentially leading to unstable pose estimation. To address this issue, this paper integrates a coarse-level instance mask matrix with the point-to-node partition strategy \cite{bib13}, thereby refining sparse instance candidate points into dense correspondences that uniformly cover the instances. Given a superpoint $\hat{q}_i \in \hat{Q}$, its assigned local region is $\mathcal{G}_i^Q$. The following outlines the specific refinement strategy:
\begin{equation}
	f\left(\hat{q}_i, \hat{q}_{i,j}, \hat{g}_{j,l}^Q\right) =
	\begin{cases}
		\text{True},  & \text{if } Cls(\hat{q}_i) = Cls(\hat{q}_{i,j}) = Cls(\hat{g}_{j,l}^Q), \\
		\text{False}, & \text{otherwise}.
	\end{cases}
\end{equation}
where $\hat{g}_{j,l}^Q \in \mathcal{G}_i^Q$ represents a local region point assigned to the superpoint $\hat{q}_j$, and $Cls(\cdot)$ represents the instance class. The function $f(\cdot)$ determines whether the local region point $\hat{g}_{j,l}^Q$ and the superpoint $\hat{q}_j$ belong to the same instance. Specifically, for each sparse superpoint correspondence $\hat{C}$, we collect their neighbor points $\hat{N}_i^P$ and $\hat{N}_i^Q$, and based on the instance mask matrix $M^Q$ and the point-to-node partition strategy, we remove local region points originating from different instances within $\mathcal{G}_i^Q$, thereby expanding the set of candidate instance points. Any point located in the overlapping local regions of $\hat{N}_i^P$ and $\hat{N}_i^Q$ suggests that $P$ may appear in $Q$, forming a candidate instance $\mathcal{A}_{m}$.

Finally, following a procedure similar to that described in \cite{bib14}, we employ an optimal transport layer to extract instance correspondences from $\mathcal{A}_{m}$, represented as $C_m$.
\subsection{Instance Filtering and Optimization}
\label{method_subsec4}
After completing these steps, the correspondence achieves high accuracy. The subsequent step involves partitioning these correspondences into subsets belonging to different instances and determining their final rigid transformations through the application of weighted SVD as outlined in Equation \ref{equation_1}. This division constitutes a clustering problem, where the number of instances corresponds to the number of clusters. Previous methods \cite{bib33} employed spectral clustering on high-dimensional features. However, clustering high-dimensional features is computationally intensive. Therefore, this paper introduces a simple yet effective method for filtering and optimizing instances. First, using the mean absolute distance error \cite{bib38}, we determine whether different local regions belong to the same instance. The specific calculation method is as follows:
\begin{equation}
	sim_{i,j}=1-\frac{ADD(T_i,T_j)}{r}
\end{equation}
where $sim_{i,j}$ similarity between two transformation relationships, $r$ represents the normalization factor, and $ADD(\cdot)$ represents the computation of the mean absolute distance error.

Secondly, based on the overlap ratio within the local regions of each instance, we rank the corresponding transformation matrices. The method for calculating the overlap ratio is detailed below:
\begin{equation}
	P_{op}=\left\{p_i^A|d(p_i^A,P^Q)\leq d_{op}^{th},p_i^A\in P^A\right\}
\end{equation}
where $P^A$ represents the transformed source point cloud, $P^Q$ represents the instance point corresponding to $P^A$, $d(p_i^A, P^Q)$ represents the point-to-plane distance, and the calculation method is detailed as follows:
\begin{equation}
	d\left(p_i^A,P^Q\right)=\min_{p_j^Q\in P^Q}\left|\left|p_i^A-p_j^Q\right|\right|
\end{equation}
where $d_{op}^{th}$ represents the distance threshold, which is utilized to determine whether two points fall within the overlapping regions of $P^A$ and $P^Q$. The method for calculating the inlier ratio is detailed as follows:
\begin{equation}
	overlap=\frac{|P_{op}|}{|P^{A}|}
\end{equation}

Finally, an iterative inlier selection method, as proposed in \cite{bib14}, is employed to gradually improve the pose estimation of instances until no new transformation relationships between instances emerge.
\subsection{Loss function}
\label{method_subsec5}
Regarding the loss functions, we employ three distinct loss functions to train MRG: 1) Overlap-aware Circle Loss; 2) Negative Log-likelihood Loss; 3) Neighbor Mask Loss. The overall loss function is detailed as follows:
\begin{equation}
	\mathcal{L}=\mathcal{L}_{circle}+\mathcal{L}_{nll}+\mathcal{L}_{mask}
\end{equation}

\textbf{Overlap-aware Circle Loss}. To supervise the superpoint features of the instance-focused module output (superpoint feature features), we adopt the method proposed in \cite{bib14}, utilizing the Overlap-aware Circle Loss, which weights the loss of each superpoint (patch) matching pair based on its overlap ratio. Given the set of local patches $\mathcal{U}$, it consists of the patches in $Q$ which have at least one positive patch $P$. For each patch $\mathcal{G}_i^Q \in \mathcal{U}$, we define the positive patches $P$, which share at least a 10\% overlap with $\mathcal{G}_i^Q$, as $\mathcal{E}_p^i$, and the negative patches, which do not overlap with $\mathcal{G}_i^Q$, as $\mathcal{E}_n^i$. The Overlap-aware Circle Loss on $Q$ is then calculated as follows:
\begin{equation}
	\mathcal{L}_{circle}^Q=\frac{1}{|\mathcal{U}|}\sum_{\mathcal{G}_i^Q\in\mathcal{U}}\log\left[1+\sum_{\mathcal{G}_i^P\in\mathcal{E}_p^i}e^{\lambda_i^j\beta_p^{i,j}\left(d_i^j-\triangle_p\right)}\cdot\sum_{\mathcal{G}_k^P\in\mathcal{E}_n^i}e^{\beta_n^{i,k}(\triangle_n-d_i^k)}\right]
\end{equation}
where $d_i^j = \| \widehat{f}_i^Q - \widehat{f}_i^P \|_2$ represents the distance in feature space, $\lambda_i^j = (lap_i^j)^{1/2}$, and $lap_i^j$ is the overlap ratio between $\mathcal{G}_i^P$ and $\mathcal{G}_i^Q$. The weights $\beta_p^{i,j} = \gamma(d_i^j - \Delta_p)$ and $\beta_n^{i,k} = \gamma(\Delta_n - d_i^k)$ are determined individually for each positive and negative example, using the margin hyperparameters $\Delta_p = 0.1$ and $\Delta_n = 1.4$. The loss $\mathcal{L}_{circle}^P$ on $P$ is computed in the same way. The overall loss is $\mathcal{L}_{circle}=(\mathcal{L}_{circle}^Q+\mathcal{L}_{circle}^P)/2$.

\textbf{Negative Log-likelihood Loss}. In accordance with \cite{bib39}, we employ a Negative Log-likelihood Loss on the assignment matrix $\bar{H}_i$ for each ground-truth superpoint correspondence $\hat{C}_i^{gt}$. For each $\hat{C}_i$, we calculate the inlier ratio between matched patches using each ground-truth transformation. Subsequently, we select the transformation corresponding to the highest inlier ratio to estimate a set of ground-truth point correspondences $\hat{C}_i$ within a matching radius $\mathrm{r}$. The point matching loss for $\hat{C}_i^{gt}$ is calculated as follows:
\begin{equation}
	\mathcal{L}_{nll,i}=-\sum_{(x,y\in \hat{C}_i^{gt})}log\bar{h}_{x,y}^i-\sum_{x\in \mathcal{F}_i}log\bar{h}_{x,m_i+1}^i-\sum_{y\in \mathcal{J}_i}log\bar{h}_{n_i+1,y}^i
\end{equation}
where $\mathcal{F}_i$ and $\mathcal{J}_i$ are the unmatched points in the two matched patches. The final loss is the average loss over all sampled superpoint matches: $\mathcal{L}_{nll}=\frac{1}{N_{g}}\sum_{i=1}^{N_{g}}\mathcal{L}_{p,i}$.

\textbf{Neighbor Mask Loss}. Following \cite{bib40}, the neighbor mask prediction loss consists of the Binary Cross-Entropy Loss and the Dice Loss with Laplace smoothing, which is defined as follows:
\begin{equation}
	\mathcal{L}_{mask,i}=BCE(m_i,m_i^{gt})+1-2\frac{m_i\cdot m_i^{gt}+1}{|m_i|+|m_i^{gt}|+1}
\end{equation}
where $m_i$ and $m_i^{gt}$ are the predicted and the ground-truth instance masks, respectively. The final loss is the average loss over all superpoints: $\mathcal{L}_{mask}=\frac{1}{N_{m}}\sum_{i=1}^{N_{m}}\mathcal{L}_{mask,i}$.
\section{Experiments}
\label{sec4}
\subsection{Experimental Setup}
\label{experimental setup}
\textbf{Datasets}. We collected a Welding-Station dataset from a real-world multi-robot welding workshop, capturing detailed 3D scans of industrial welding environments. The dataset includes different models of robots, various welding stations, electrical equipment, and fixtures. This diversity reflects the complexity and variability of real industrial settings. Each scene contains accurate annotations of object instances and their rigid transformations. For our experiments, we sampled the target point clouds from workshop scenes and the source point clouds from corresponding sources. We allocated 70\% of the data for training, 10\% for validation, and 20\% for testing.

To comprehensively evaluate the performance of our method, we conducted experiments on the publicly available Scan2CAD dataset \cite{bib41}. The Scan2CAD dataset precisely aligns object instances from ScanNet \cite{bib42} with CAD models from ShapeNet \cite{bib37}. It provides accurate rigid transformation annotations for multiple real-world scanned scenes, each containing 2 to 5 identical CAD instances. We fully utilize the annotation information and conduct experiments by separately sampling the target point clouds from the scene point clouds and the source point clouds from CAD models. After obtaining 2,175 sets of point clouds, we used 1,523 scenes for training, 326 scenes for validation, and 326 scenes for testing. We utilize the fine-tuned Predator \cite{bib18} for point matching to establish the initial putative correspondence set.

\textbf{Metrics}. We follow the evaluation procedure of PointCLM \cite{bib33}, where the rotation error is defined as
\begin{equation}
	RE=arccos[(Tr(R_{gt}^TR_{est})-1)/2]
\end{equation}
and translation error is defined as
\begin{equation}
	TE=\left\|t_{est}-t_{gt}\right\|_2
\end{equation}

For the Welding-Station dataset, we established strict thresholds to evaluate registration success, aligning with the high-precision requirements of industrial environments. A registration is considered successful if the Rotation Error (RE) $\leq 15^\circ$ and the Translation Error (TE) $\leq 0.2$ m. For the public Scan2CAD dataset \cite{bib41}, we adhered to the commonly used evaluation standards in the field to ensure result comparability. A registration is considered successful if RE $\leq 15^\circ$ and TE $\leq 0.1$ m. We employed Mean Recall (MR), Mean Precision (MP), and their harmonic mean (MF) as evaluation metrics, which are defined as follows:
\begin{equation}
	MR=\frac{1}{N_{pair}}\sum_{i=1}^{N_{pair}}\frac{M_i^{suc}}{M_i^{gt}}
\end{equation}
\begin{equation}
	MP=\frac{1}{N_{pair}}\sum_{i=1}^{N_{pair}}\frac{M_i^{suc}}{M_i^{pred}}
\end{equation}
\begin{equation}
	MF=\frac{2\times MR\times MP}{MR+MP}
\end{equation}
where $N_{pair}$ represents the number of paired point clouds, $M^{suc}$ represents the number of successful registration instances, $M^{gt}$ represents the actual number of instances and $M^{pred}$ represents the number of predicted transformations.

\textbf{Implementation Details}. The experiments in this paper were conducted on a computing platform with an Intel(R) Core(R) i5-12400F processor running at 4.5 GHz and an NVIDIA RTX 3060 with 12 GB of memory. The code was developed using CUDA version 11.7 and PyTorch version 1.13.0. The Adam optimizer was employed to update the model parameters, with an initial learning rate of $10^{-4}$, a momentum of 0.98, and a weight decay of $10^{-6}$. The learning rate decayed exponentially with a decay rate of 0.05 after each epoch. We trained the network for 100 epochs, and all point clouds were downsampled to a voxel size of 0.05 m. The default value for k-nearest neighbors was set to 32 for both the Scan2CAD dataset and the Welding-Station dataset. We determined whether each point and its $K$-nearest neighbors belong to the same instance based on a confidence threshold $\tau$. Points with confidence scores exceeding 0.5 are considered part of the same instance; those with lower scores are not. The default value for the normalization factor $r$ is set to the instance diameter. The distance threshold $d_{op}^{th}$ is set to 1.5 $pr$, where $pr$ is the resolution of a point cloud \cite{bib43}. The similarity threshold $\theta$ is set to 0.8 on the Welding-Station dataset and 0.7 on the Scan2CAD dataset.

\textbf{Baseline Methods}. We compared MRG against three multi-model fitting methods(RansaCov \cite{bib15}, CONSAC \cite{bib16}, T-linkage \cite{bib17}) and two state-of-the art multi-instance point cloud registration methods (ECC \cite{bib19}, PointCLM \cite{bib33}). We tuned all methods to achieve optimal performance on the evaluation dataset within reasonable time and memory consumption constraints. To ensure a fair comparison, all methods used the same assumed correspondences as input.
\subsection{Evaluation on Real-World Welding-Station Dataset}
\begin{figure}[!ht]
	\centering
	\includegraphics[width=\textwidth]{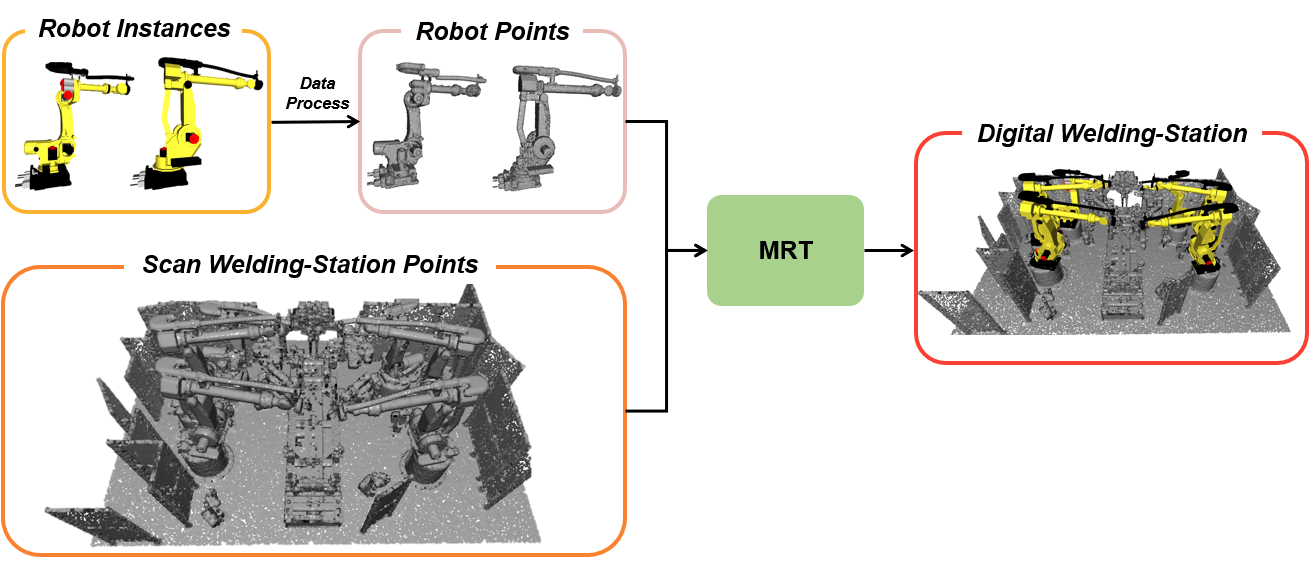}
	\caption{Process for welding station digital manufacturing scene generation.}
	\label{fig:generation_process}
\end{figure}
To clearly illustrate the process of generating a digital shop model, we utilize an automotive body welding station as an example. This workstation comprises two distinct robot models, resulting in six robot instances. To generate the source point cloud, we sample points from each triangular face of the robot CAD models. The source point cloud is then aligned with and matched to the corresponding robot instances in the welding station. Through this registration process, we construct a digital manufacturing scene model for the body welding station. Fig. \ref{fig:generation_process} illustrates the complete workflow of the digital scene generation process.
\begin{table}[!ht]
	\caption{Multi-instance registration results on Welding-Station dataset. $\uparrow$ means the larger the better, while $\downarrow$ indicates the contrary.}
	\label{table1}
	\small
	\begin{tabularx}{\textwidth}{>{\centering\arraybackslash}X|>{\centering\arraybackslash}X>{\centering\arraybackslash}X>{\centering\arraybackslash}X>{\centering\arraybackslash}X}
		\toprule
		Method    & MR(\%)$\uparrow$         & MP(\%)$\uparrow$         & MF(\%)$\uparrow$         & Time(s)$\downarrow$       \\ \midrule
		T-linkage\cite{bib17} & 20.05          & 28.42          & 23.51          & 11.31         \\ 
		RansaCov\cite{bib15}  & 42.58          & 20.85          & 27.99          & \textbf{0.32} \\ 
		CONSAC\cite{bib16}    & 43.66          & 41.54          & 42.57          & 0.61          \\ 
		ECC\cite{bib19}       & 54.22          & {$\underline{57.15}$}          & 55.65          & 2.14          \\ 
		PointCLM\cite{bib33}  & {$\underline{62.37}$}    & 55.10    & {$\underline{58.50}$}    & {$\underline{0.53}$}    \\ 
		Ours      & \textbf{75.10} & \textbf{72.64} & \textbf{73.85} & 1.33          \\ 
		\bottomrule
	\end{tabularx}
\end{table}

As shown in Table \ref{table1}. Our MRG method outperforms all other methods. Compared to the closest competitor, it achieves improvements of 16.95\% in MR metrics, 24.15\% in MP metrics, and 20.79\% in MF metrics. Although our method requires an additional 0.8s processing time compared to the state-of-the-art method, accuracy remains paramount in industrial digital scene generation applications. By leveraging the distance-invariance matrix, ECC directly clusters noise correspondences into distinct groups. While this method significantly improves registration accuracy, it incurs substantially increased computational complexity. PointCLM proposes a two-stage method: first obtaining discriminative high-dimensional correspondence representations through contrastive learning, then applying spectral clustering to these features embeddings. This novel framework achieves simultaneous improvements in both accuracy and computational efficiency. However, experimental results indicate that the learned high-dimensional features yield only marginal improvements in overall performance. Moreover, the remaining three multi-model fitting methods demonstrate suboptimal performance in multi-instance point cloud registration tasks within welding scenes. Notably, none of these methods achieves the registration accuracy required for industrial applications.

\begin{figure}[H]
	\centering
	\begin{minipage}{0.2\textwidth}  % 文本部分
		\centering
		T-linkage\cite{bib17}
	\end{minipage}%
	\begin{minipage}{0.26\textwidth}  % 图片1部分
		\centering
		\includegraphics[width=\textwidth]{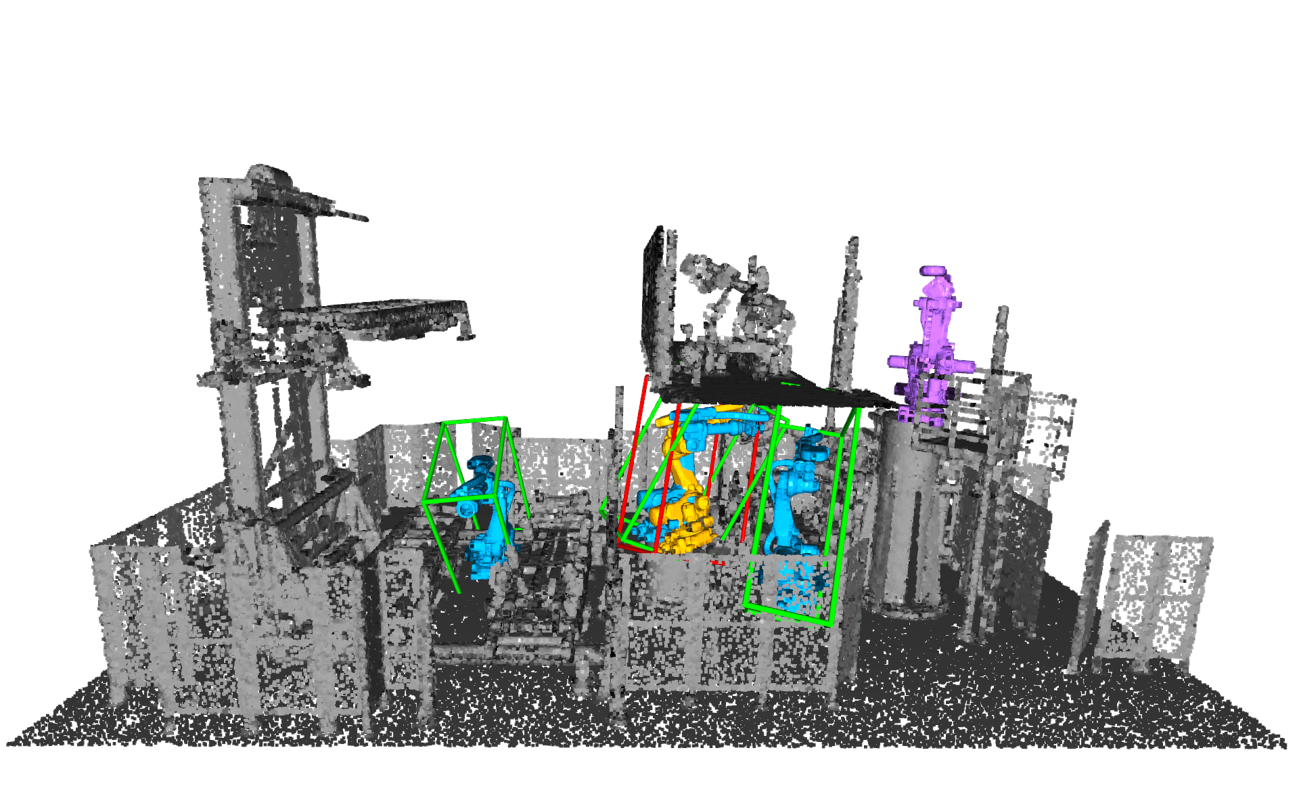}
	\end{minipage}%
	\begin{minipage}{0.26\textwidth}  % 图片2部分
		\centering
		\includegraphics[width=\textwidth]{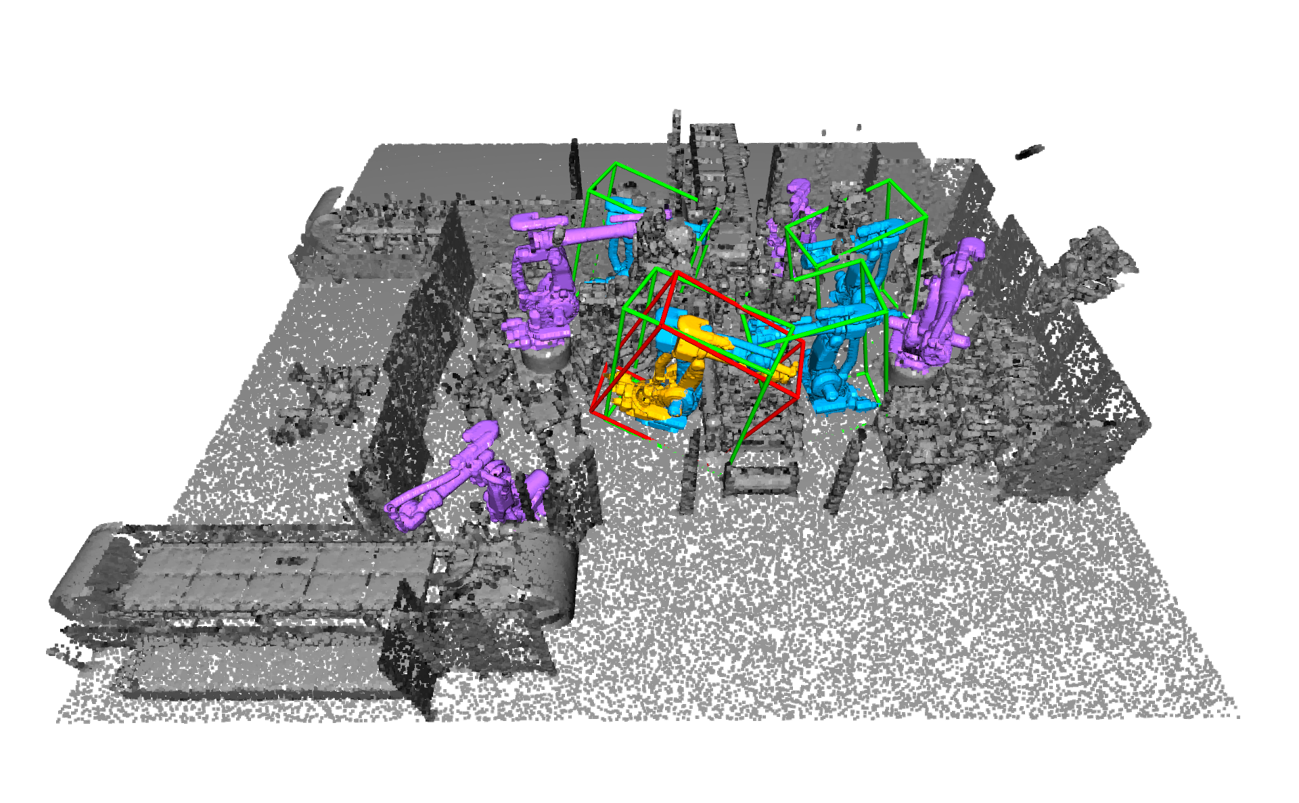}
	\end{minipage}%
	\begin{minipage}{0.26\textwidth}  % 图片3部分
		\centering
		\includegraphics[width=\textwidth]{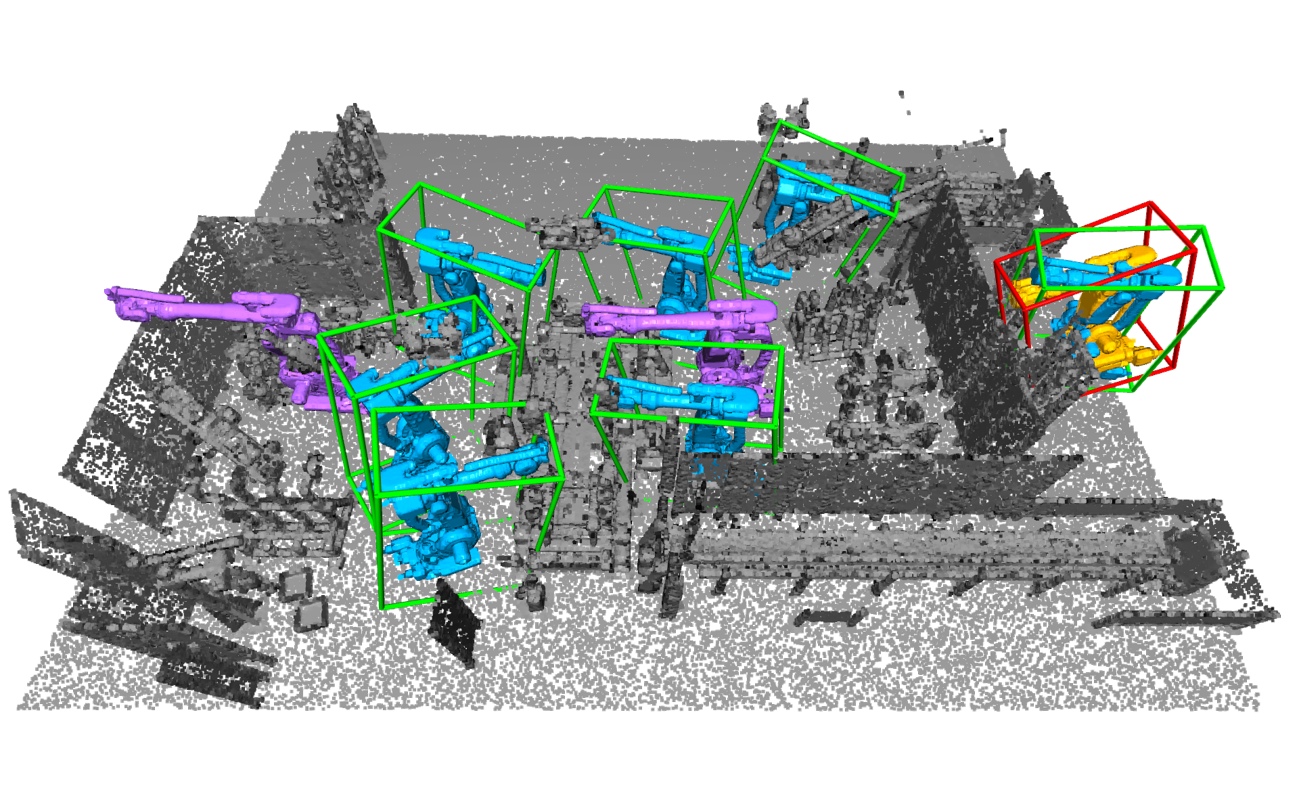}
	\end{minipage}
\end{figure}
\vspace{-10mm}
\begin{figure}[H]
	\centering
	\begin{minipage}{0.2\textwidth}  % 文本部分
		\centering
		RansaCov\cite{bib15}
	\end{minipage}%
	\begin{minipage}{0.26\textwidth}  % 图片1部分
		\centering
		\includegraphics[width=\textwidth]{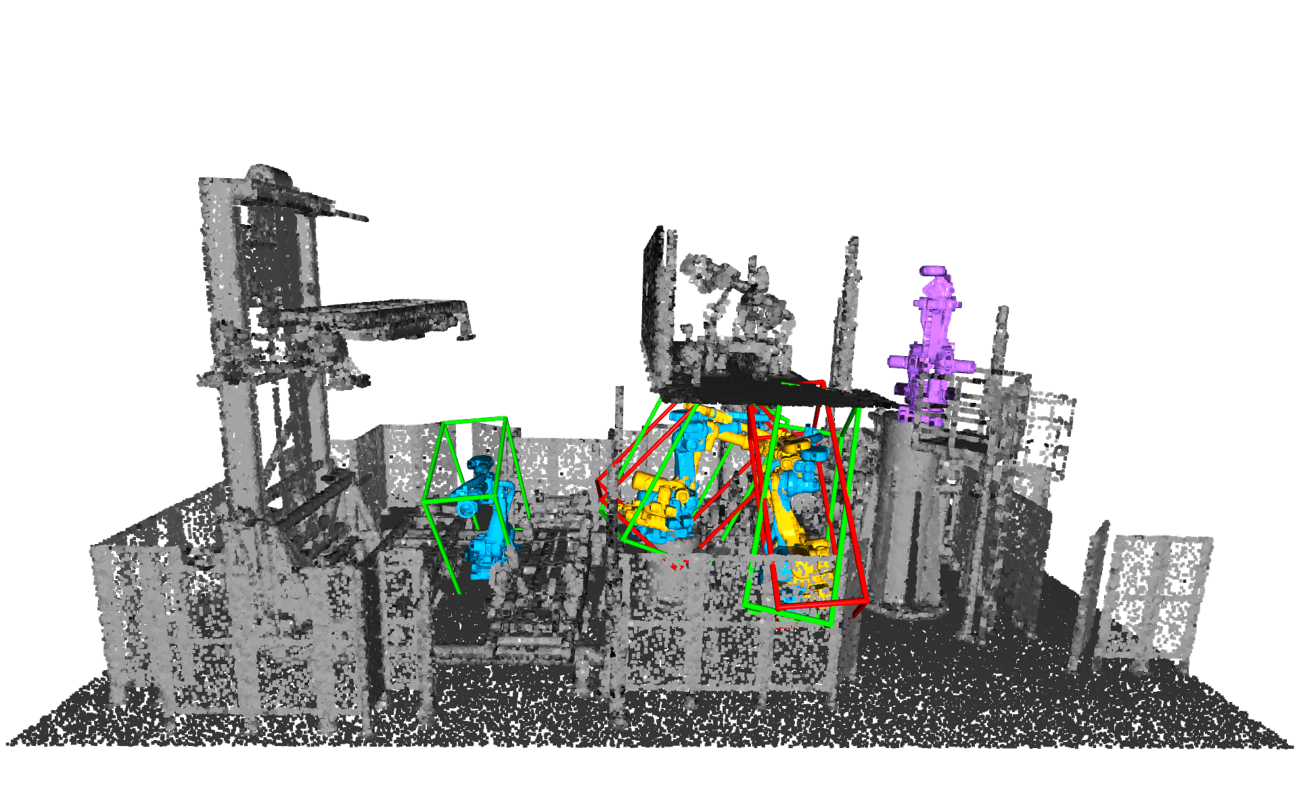}
	\end{minipage}%
	\begin{minipage}{0.26\textwidth}  % 图片2部分
		\centering
		\includegraphics[width=\textwidth]{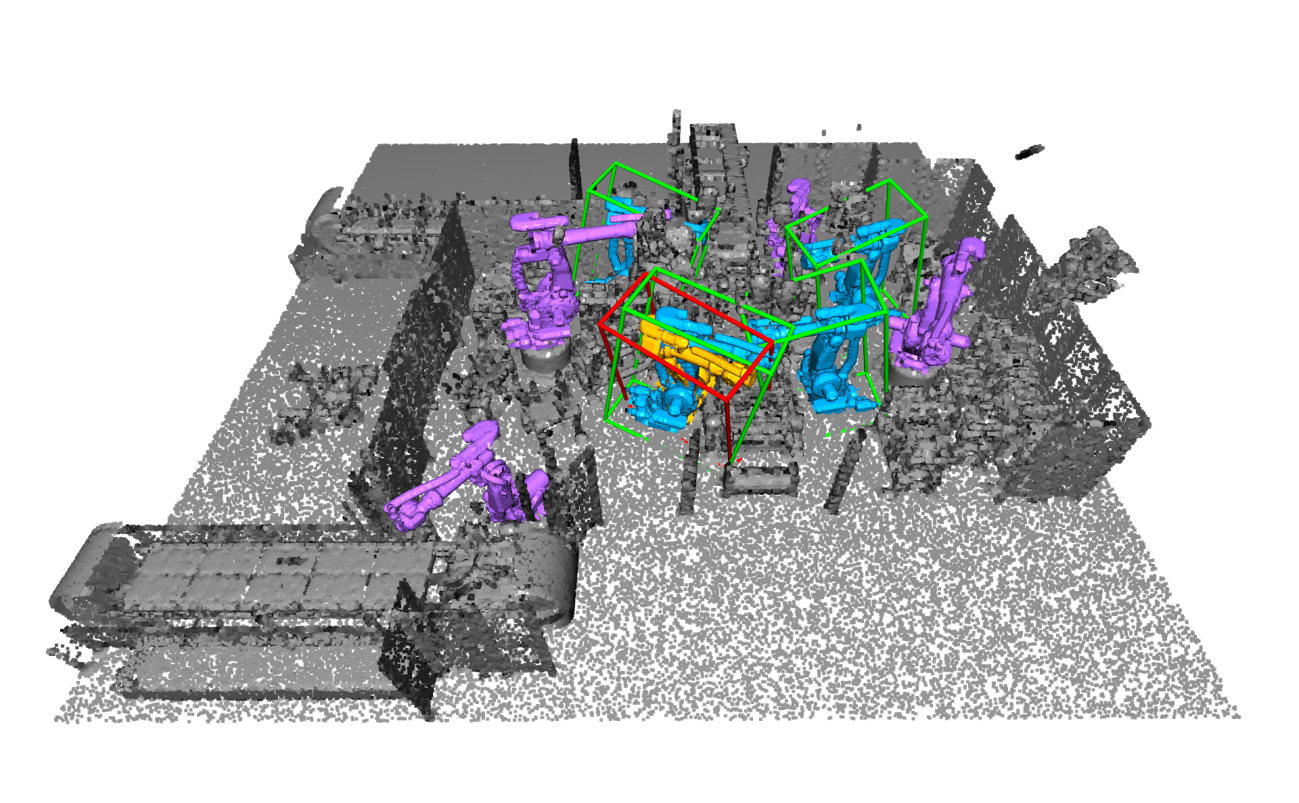}
	\end{minipage}%
	\begin{minipage}{0.26\textwidth}  % 图片3部分
		\centering
		\includegraphics[width=\textwidth]{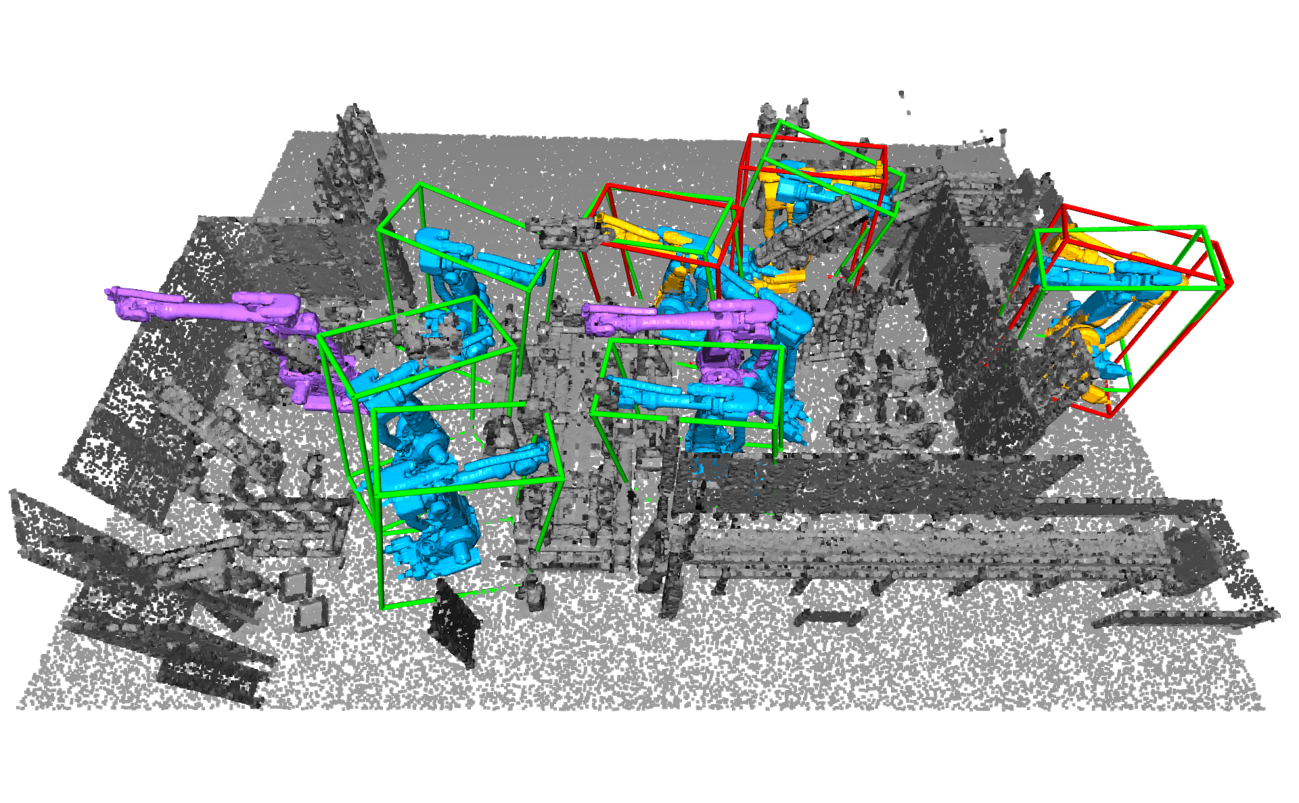}
	\end{minipage}
\end{figure}
\vspace{-10mm}
\begin{figure}[H]
	\centering
	\begin{minipage}{0.2\textwidth}  % 文本部分
		\centering
		CONSAC\cite{bib16}
	\end{minipage}%
	\begin{minipage}{0.26\textwidth}  % 图片1部分
		\centering
		\includegraphics[width=\textwidth]{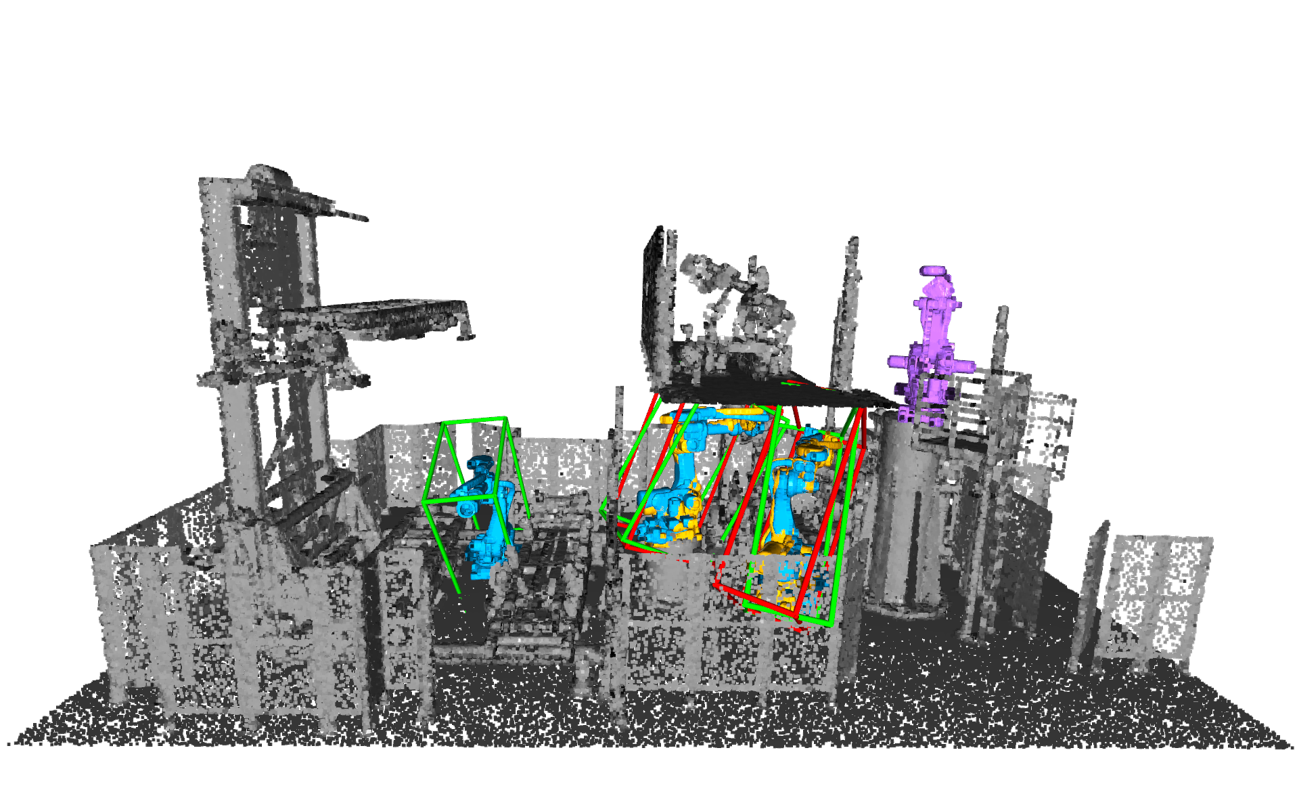}
	\end{minipage}%
	\begin{minipage}{0.26\textwidth}  % 图片2部分
		\centering
		\includegraphics[width=\textwidth]{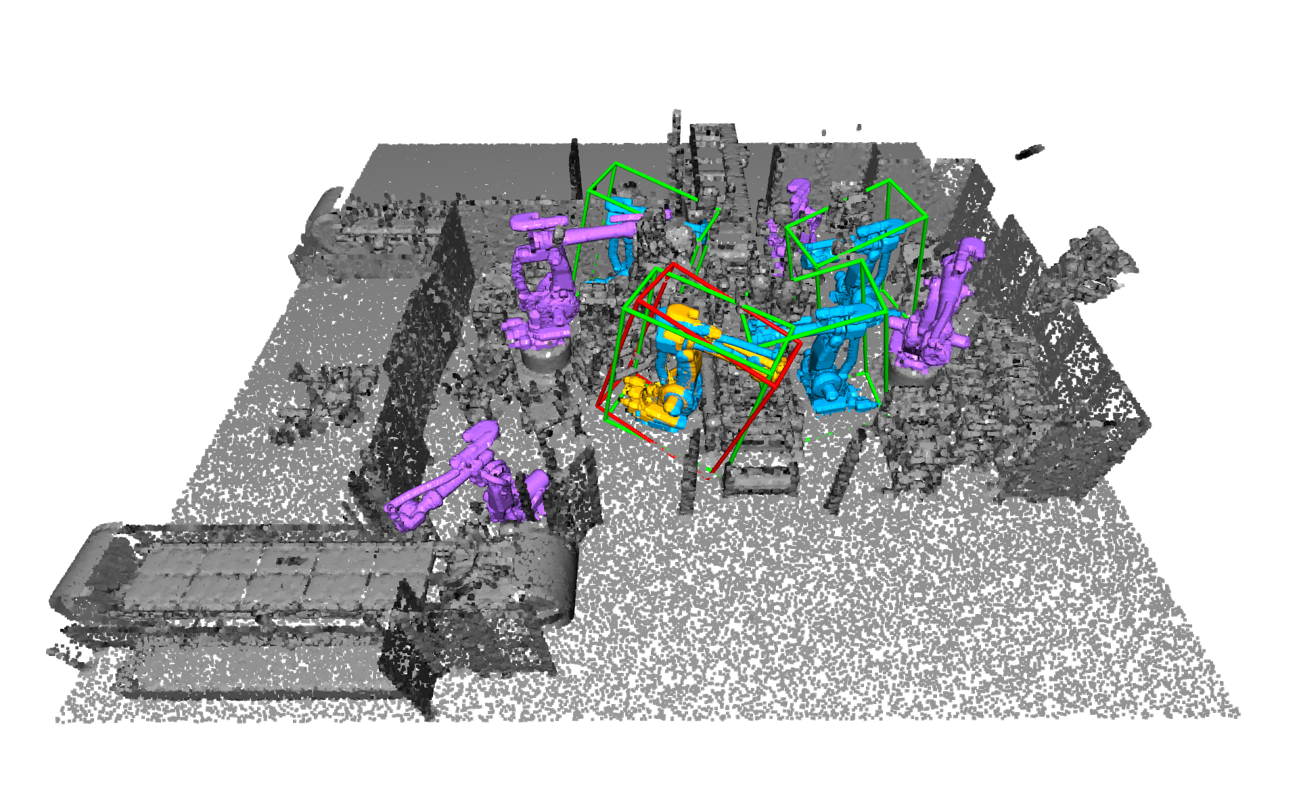}
	\end{minipage}%
	\begin{minipage}{0.26\textwidth}  % 图片3部分
		\centering
		\includegraphics[width=\textwidth]{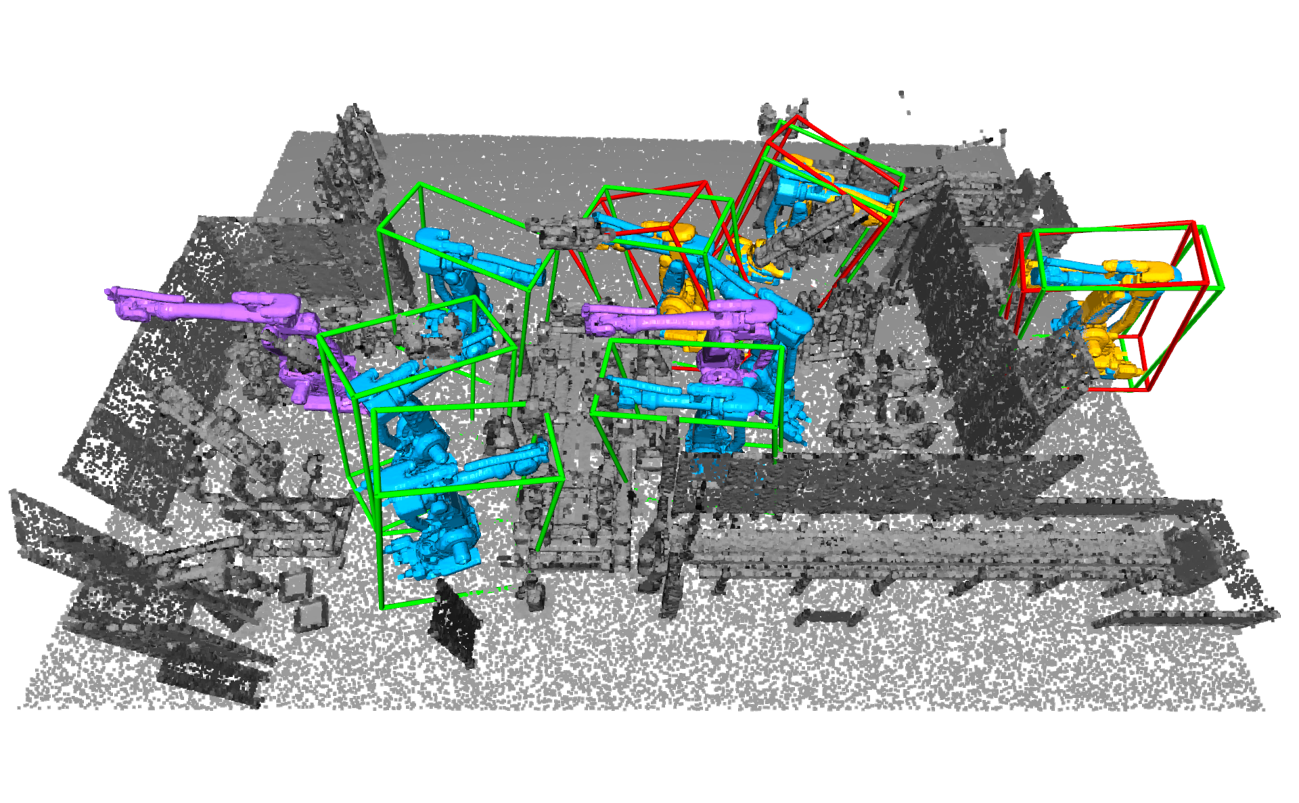}
	\end{minipage}
\end{figure}
\vspace{-10mm}
\begin{figure}[H]
	\centering
	\begin{minipage}{0.2\textwidth}  % 文本部分
		\centering
		ECC\cite{bib19}
	\end{minipage}%
	\begin{minipage}{0.26\textwidth}  % 图片1部分
		\centering
		\includegraphics[width=\textwidth]{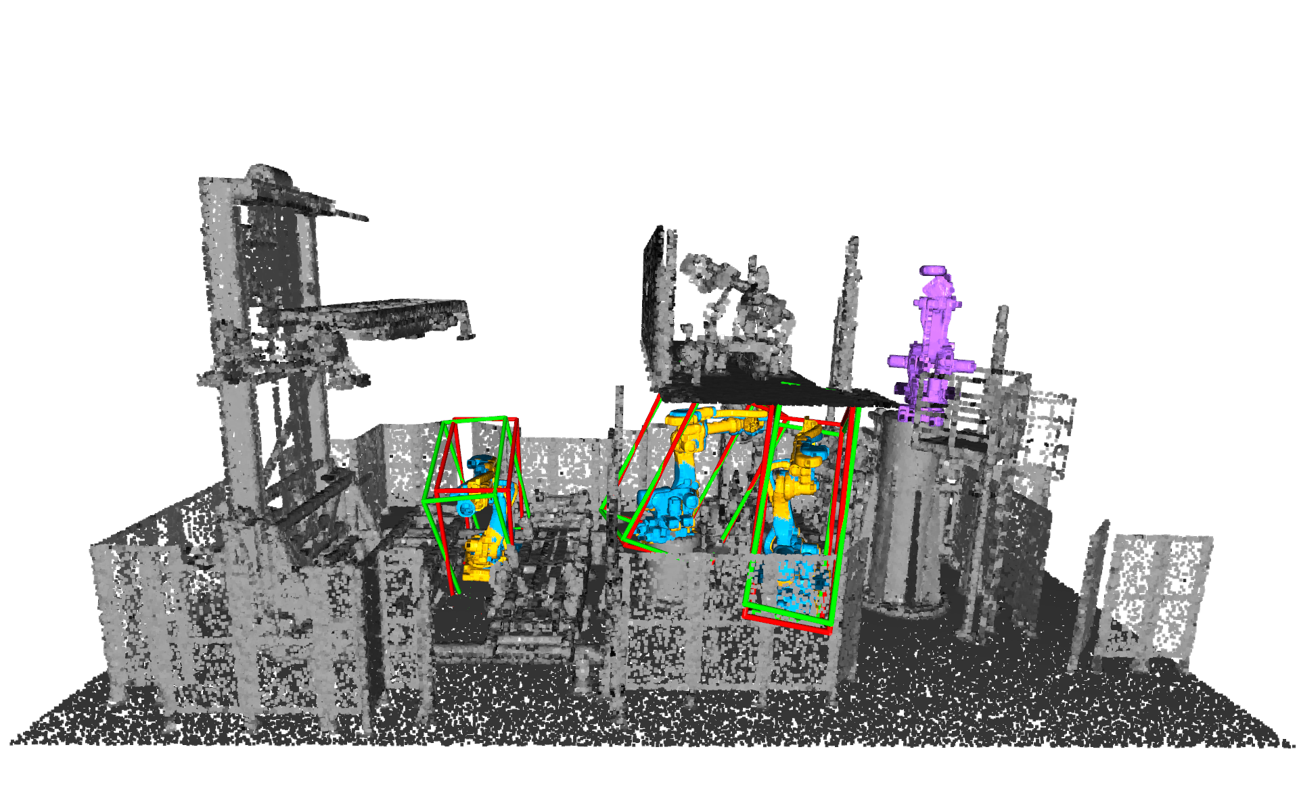}
	\end{minipage}%
	\begin{minipage}{0.26\textwidth}  % 图片2部分
		\centering
		\includegraphics[width=\textwidth]{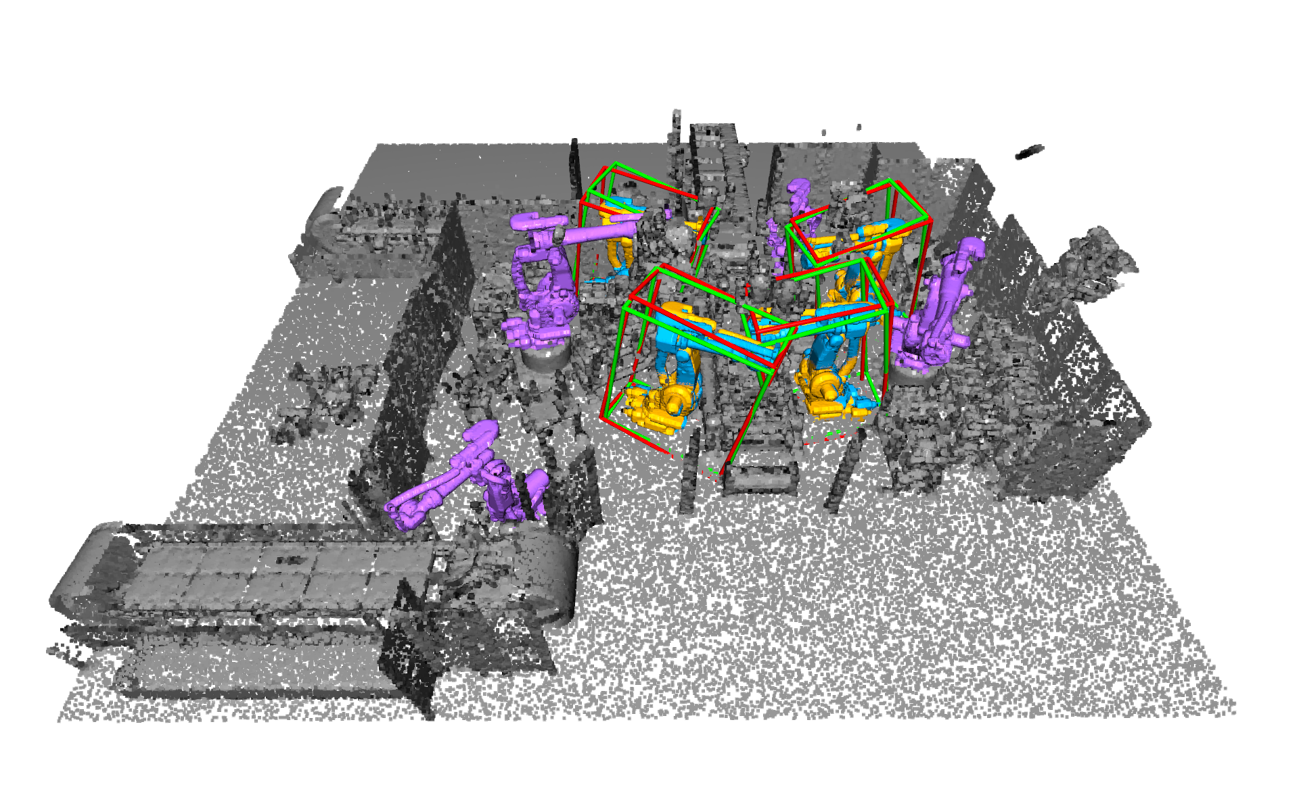}
	\end{minipage}%
	\begin{minipage}{0.26\textwidth}  % 图片3部分
		\centering
		\includegraphics[width=\textwidth]{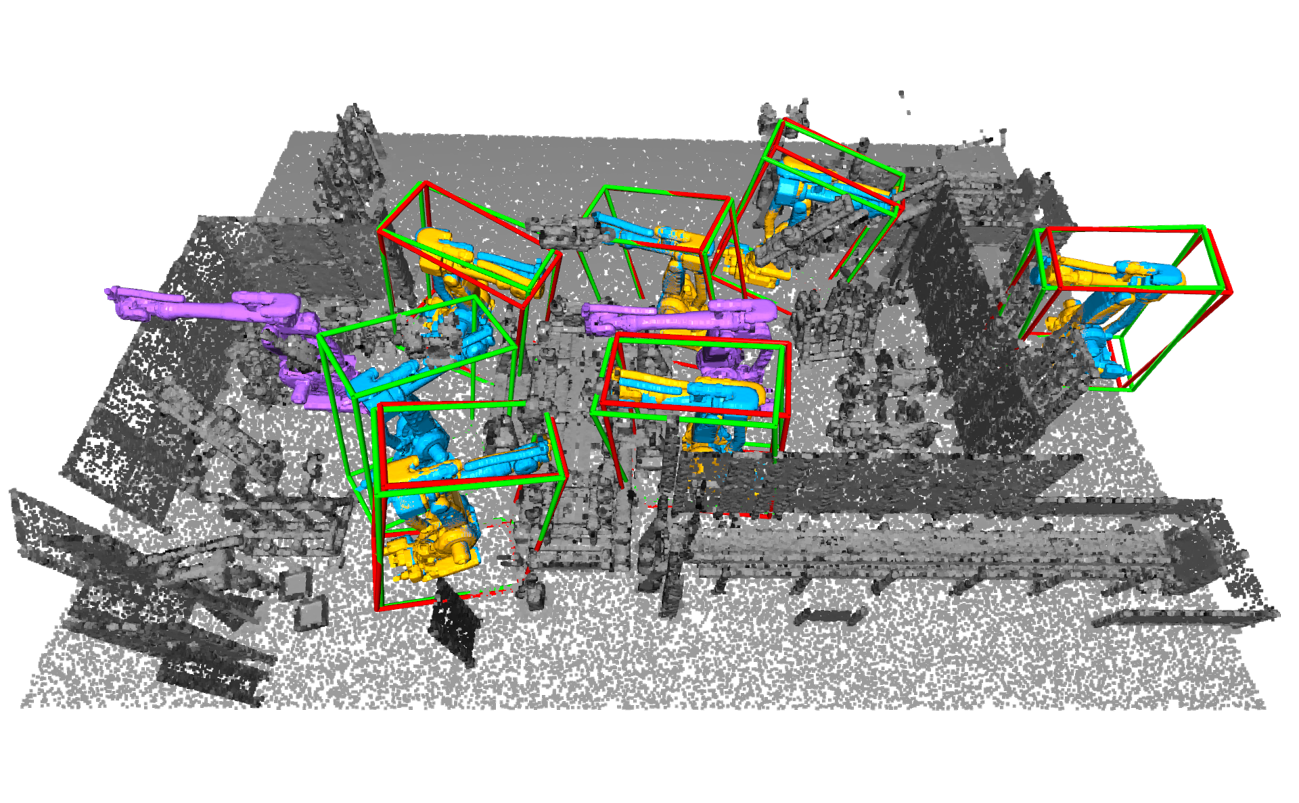}
	\end{minipage}
\end{figure}
\vspace{-10mm}
\begin{figure}[H]
	\centering
	\begin{minipage}{0.2\textwidth}  % 文本部分
		\centering
		PointCLM\cite{bib33}
	\end{minipage}%
	\begin{minipage}{0.26\textwidth}  % 图片1部分
		\centering
		\includegraphics[width=\textwidth]{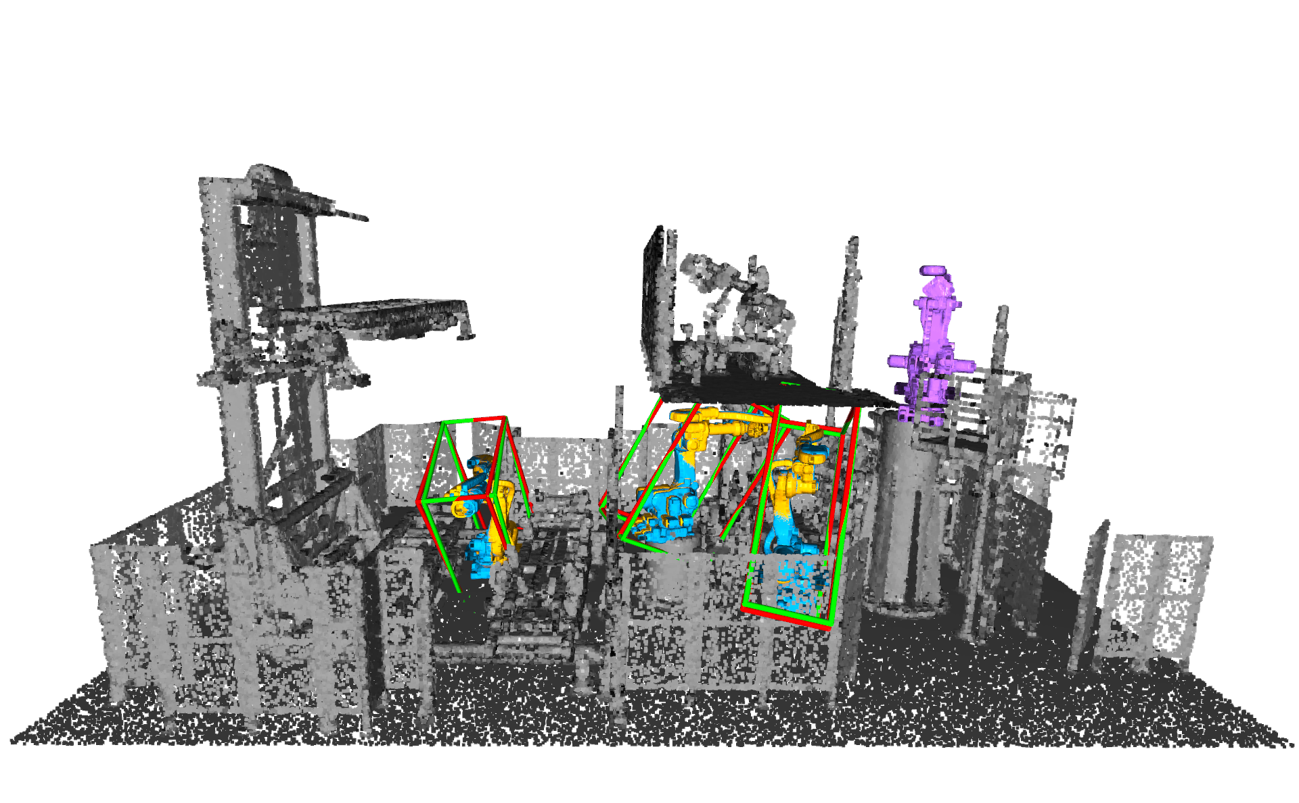}
	\end{minipage}%
	\begin{minipage}{0.26\textwidth}  % 图片2部分
		\centering
		\includegraphics[width=\textwidth]{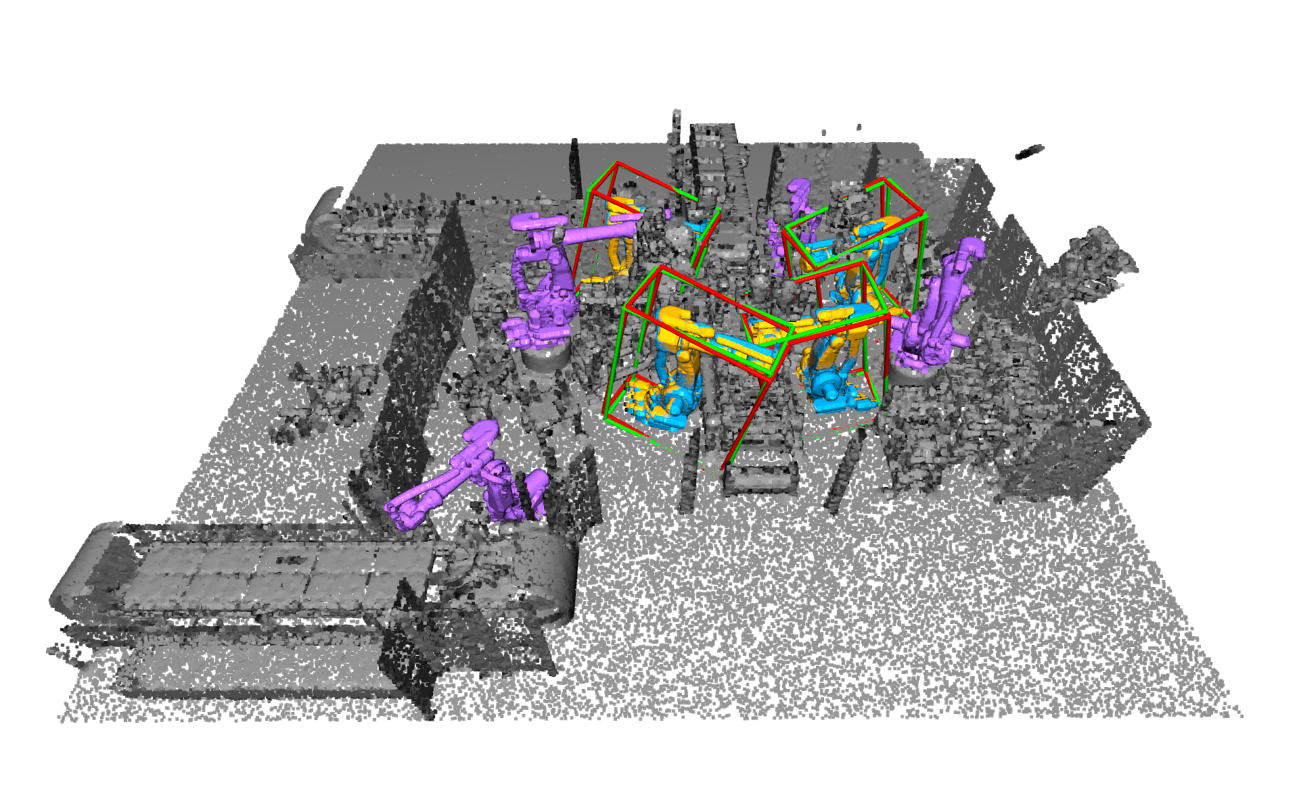}
	\end{minipage}%
	\begin{minipage}{0.26\textwidth}  % 图片3部分
		\centering
		\includegraphics[width=\textwidth]{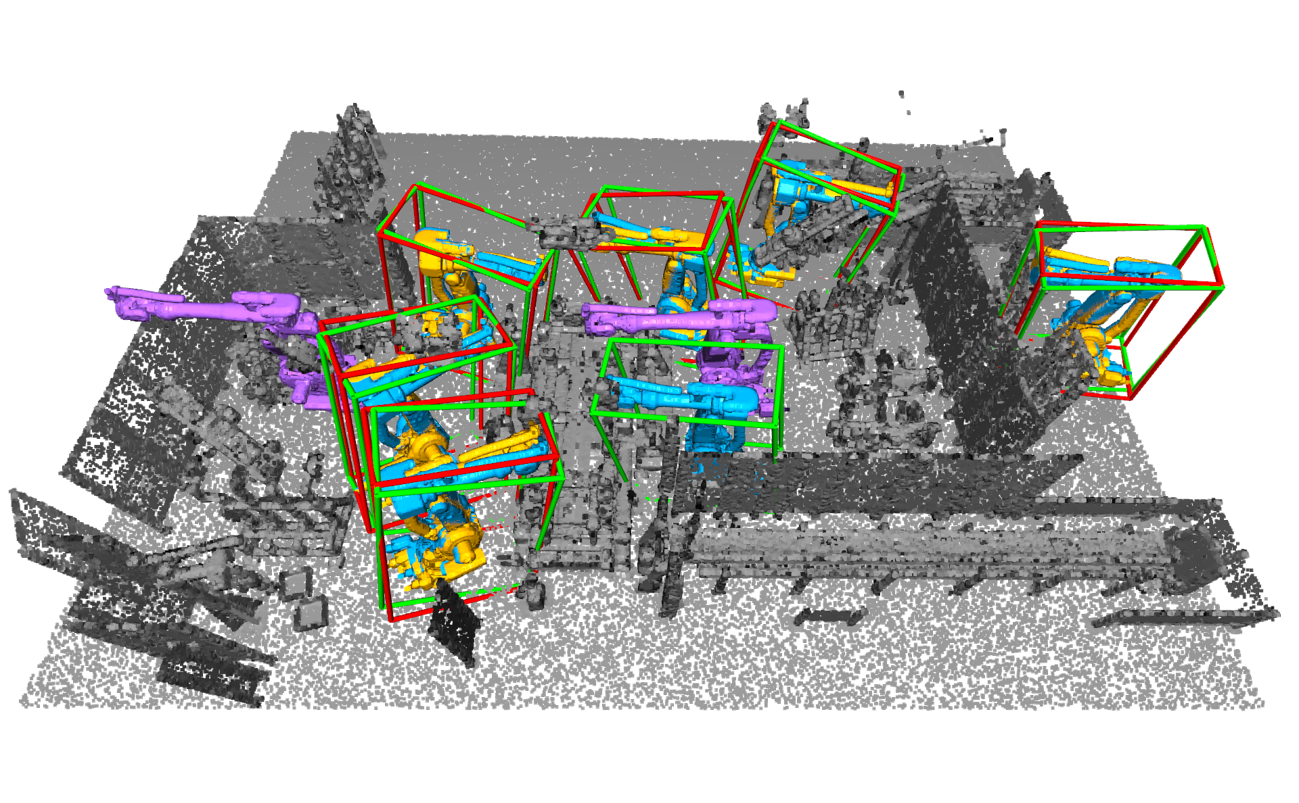}
	\end{minipage}
\end{figure}
\vspace{-10mm}
\begin{figure}[H]
	\centering
	\begin{minipage}{0.2\textwidth}  % 文本部分
		\centering
		Ours
	\end{minipage}%
	\begin{minipage}{0.26\textwidth}  % 图片1部分
		\centering
		\includegraphics[width=\textwidth]{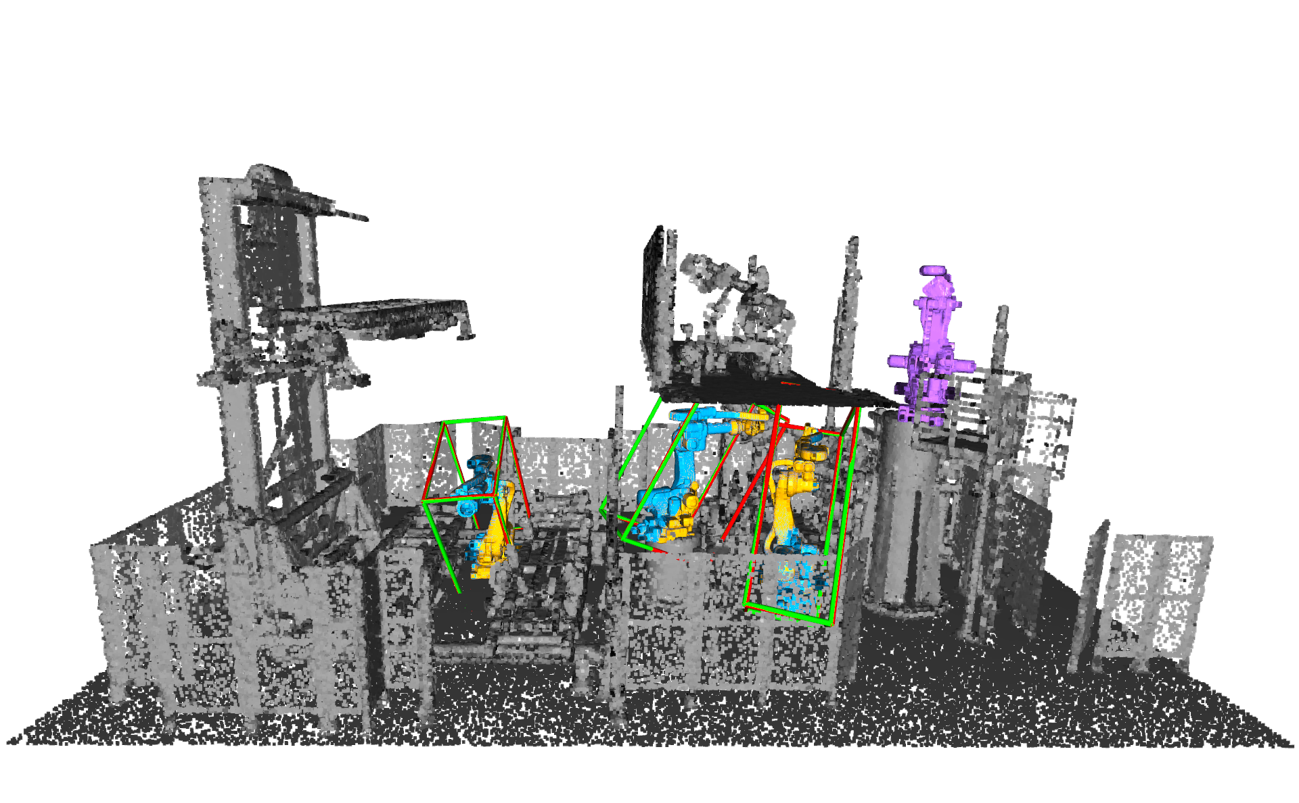}
	\end{minipage}%
	\begin{minipage}{0.26\textwidth}  % 图片2部分
		\centering
		\includegraphics[width=\textwidth]{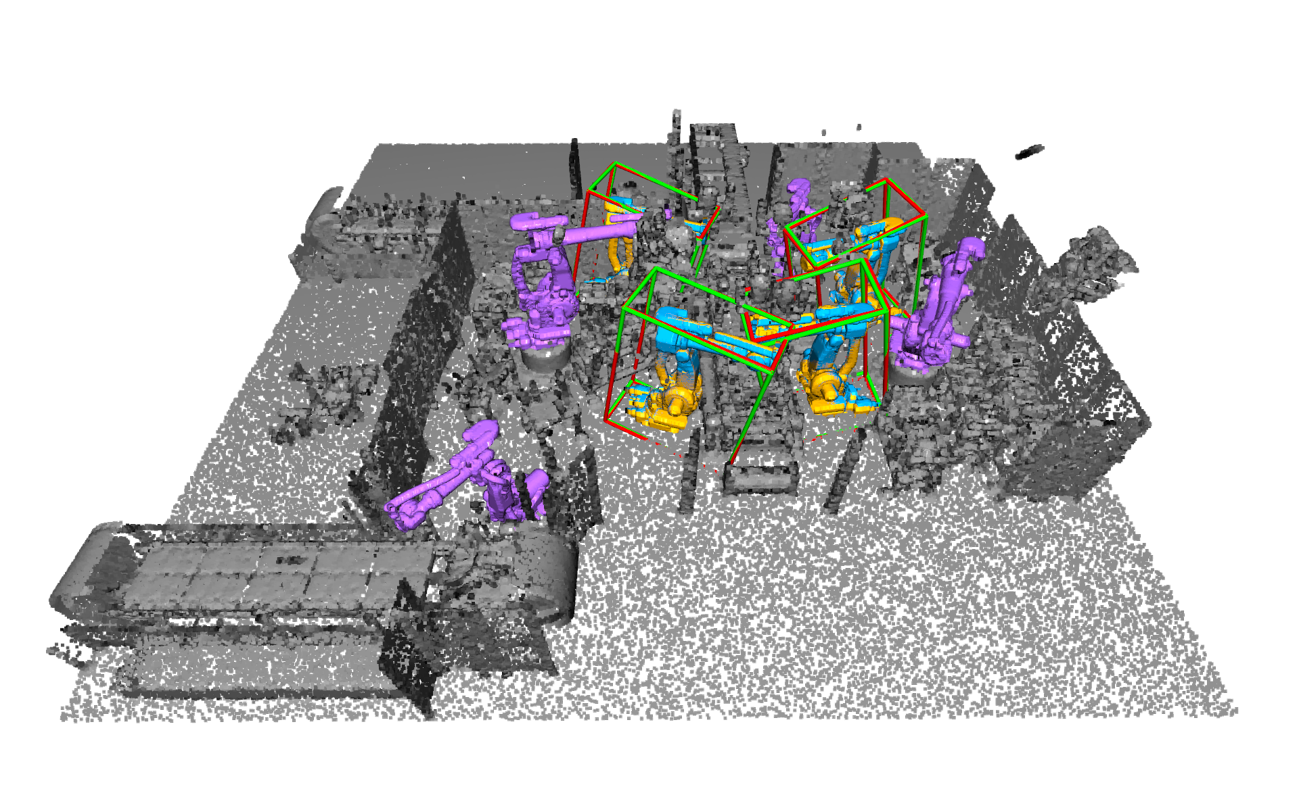}
	\end{minipage}%
	\begin{minipage}{0.26\textwidth}  % 图片3部分
		\centering
		\includegraphics[width=\textwidth]{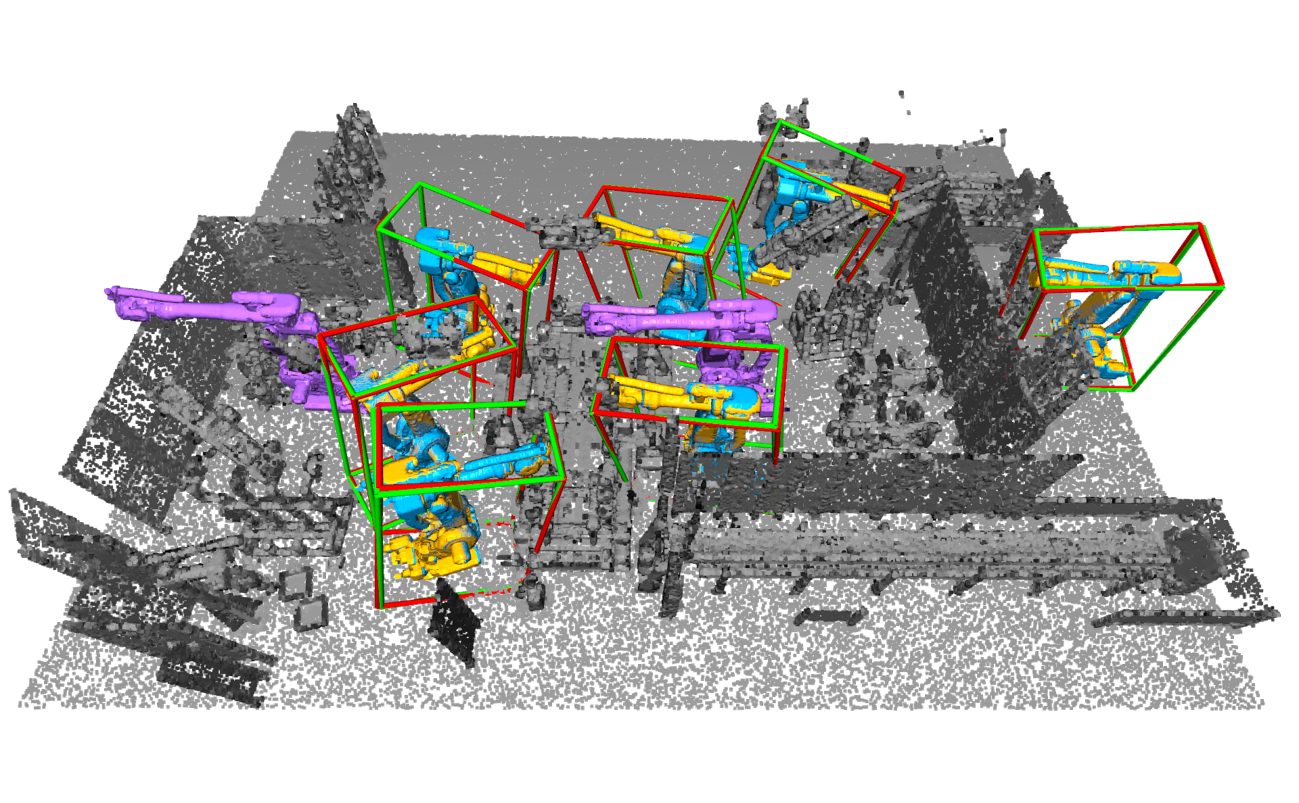}
	\end{minipage}
	\caption{Examples of multi-instance registrations on the Welding-Station dataset.}
	\label{fig:weldingstation_results}
\end{figure}

We present the visualization of the multi-instance point cloud registration results in Fig. \ref{fig:weldingstation_results}. In scenes with multiple similar instances, Our MRG accurately estimates the number of rigid transformations and the corresponding transformation poses with minimal errors. Due to the use of a powerful instance-focused geometric transformer module, our MRG effectively distinguish all instances. This module accurately estimates instance boundaries and captures the relationships among local regions within each instance. However, in welding stations with numerous outlier points. The limitations of global spatial consistency and the similarity of local structures between instances prevent PointCLM and ECC from distinguishing multiple similar instances. This results in missed detections or poor pose estimation outcomes. The other three multi-model fitting methods can only successfully register a few obvious instances but with significant errors.

\subsection{Evaluation on Scan2CAD Dataset}
In addition to conducting experiments on the Welding-Station dataset, we evaluated our method using the Scan2CAD dataset. Table \ref{tab:scan2cad_results} presents the quantitative results, demonstrating that MRG outperforms all other methods across three metrics: MR, MP, and MF. On average, MRG achieves improvements of 35.65\%, 37.31\%, and 37.49\% in these metrics. Notably, MRG outperforms PointCLM and ECC across three metrics without a significantly increasing in computational time. Compared to the other three multi-model fitting methods, PointCLM and ECC also demonstrate satisfactory performance.

\begin{table}[!ht]
	\caption{Multi-instance registration results on Scan2CAD dataset.}
	\label{tab:scan2cad_results}
	\small
	\begin{tabularx}{\textwidth}{>{\centering\arraybackslash}X|>{\centering\arraybackslash}X>{\centering\arraybackslash}X>{\centering\arraybackslash}X>{\centering\arraybackslash}X}
		\toprule
		Method    & MR(\%)$\uparrow$         & MP(\%)$\uparrow$         & MF(\%)$\uparrow$         & Time(s)$\downarrow$       \\ \midrule
		T-linkage\cite{bib17} & 33.56          & 45.69          & 38.70          & 6.67         \\ 
		RansaCov\cite{bib15}  & 57.25          & 55.24          & 56.22          & 0.25 \\ 
		CONSAC\cite{bib16}    & 58.00          & 32.33          & 41.67          & \textbf{0.10}          \\ 
		ECC\cite{bib19}       & 67.56          & 73.32          & 70.32          & 1.80          \\ 
		PointCLM\cite{bib33}  & {$\underline{81.26}$}    & {$\underline{73.44}$}    & {$\underline{77.15}$}    & {$\underline{0.12}$}    \\ 
		Ours      & \textbf{92.50} & \textbf{89.33} & \textbf{90.89} & 0.92          \\ 
		\bottomrule
	\end{tabularx}
\end{table}

\begin{figure}[H]
	\centering
	\begin{minipage}{0.2\textwidth}  % 文本部分
		\centering
		Ours
	\end{minipage}%
	\begin{minipage}{0.26\textwidth}  % 图片1部分
		\centering
		\includegraphics[width=\textwidth]{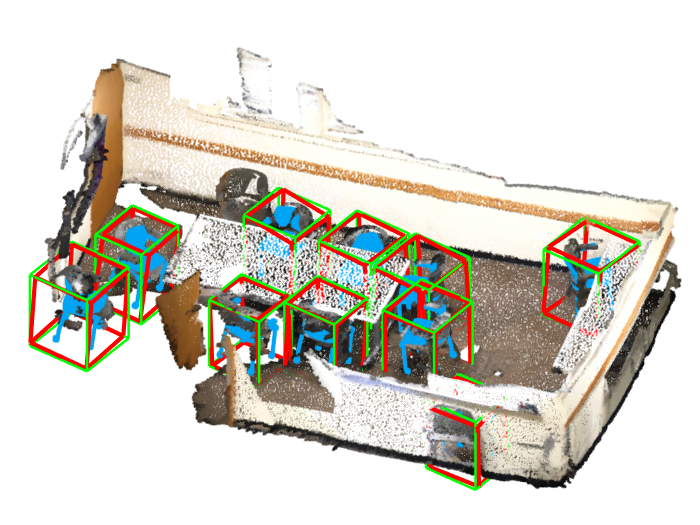}
	\end{minipage}%
	\begin{minipage}{0.26\textwidth}  % 图片2部分
		\centering
		\includegraphics[width=\textwidth]{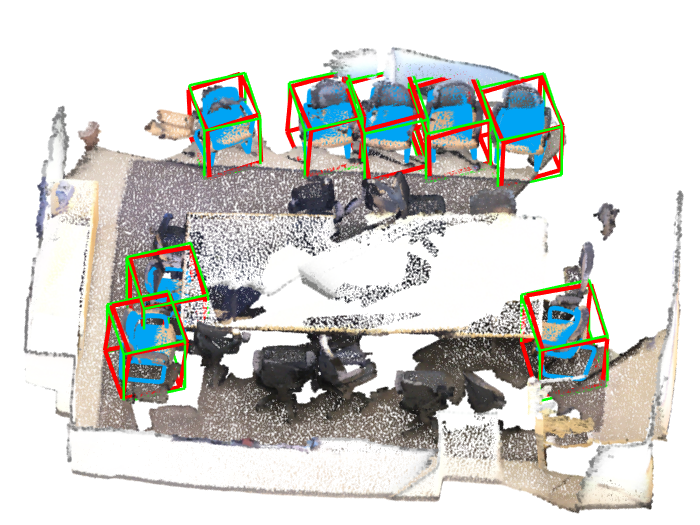}
	\end{minipage}%
	\begin{minipage}{0.26\textwidth}  % 图片3部分
		\centering
		\includegraphics[width=\textwidth]{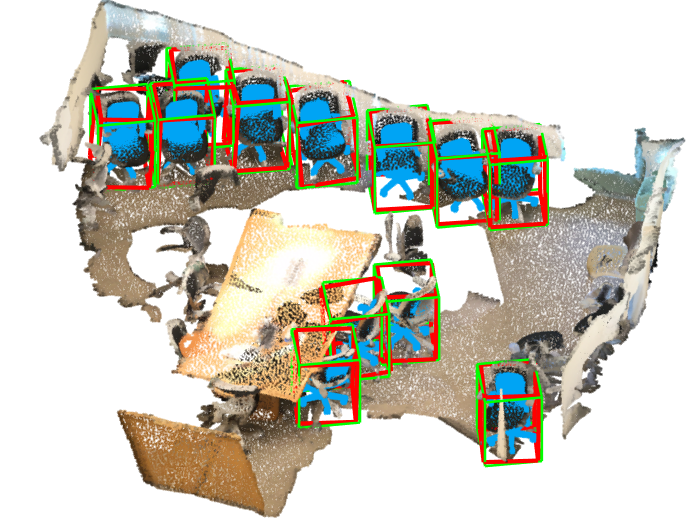}
	\end{minipage}
	\caption{Examples of multi-instance registrations on the Scan2CAD dataset.}
	\label{fig:scan2cad_results}
\end{figure}

To facilitate qualitative comparisons with competing methods, we present visual registration results for three sets of indoor point cloud scenes, each containing 8, 10, and 12 identical instances, as shown in Fig. \ref{fig:scan2cad_results}. As the number of instances increases, our MRG method retains its capability to accurately predict the number of instances in the scene and estimate the transformation poses of all source point clouds with minimal error. The PointCLM and ECC methods can generally predict the poses of all instances, albeit with greater pose errors. The three multi-model fitting methods only identify a limited number of easily distinguishable instances and estimate their poses with reduced accuracy.

\begin{figure}[H]
	\centering
	\begin{minipage}{0.2\textwidth}  % 文本部分
		\centering
		T-linkage\cite{bib17}
	\end{minipage}%
	\begin{minipage}{0.26\textwidth}  % 图片1部分
		\centering
		\includegraphics[width=\textwidth]{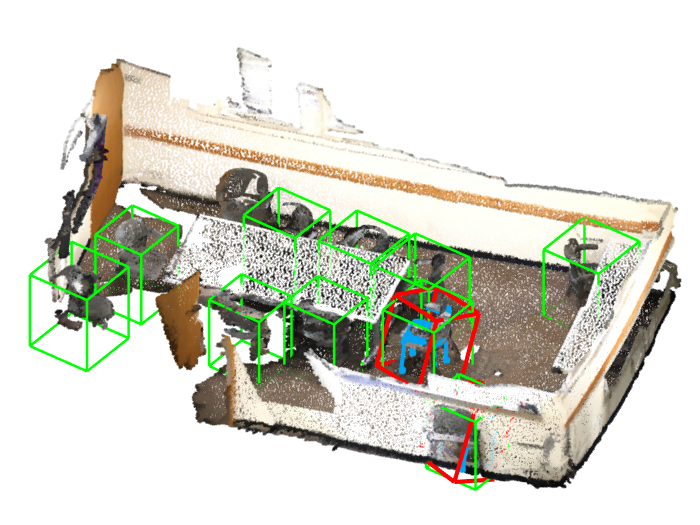}
	\end{minipage}%
	\begin{minipage}{0.26\textwidth}  % 图片2部分
		\centering
		\includegraphics[width=\textwidth]{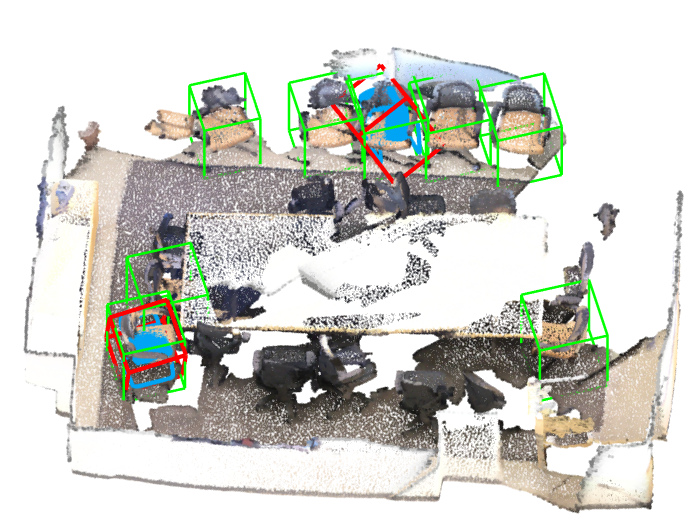}
	\end{minipage}%
	\begin{minipage}{0.26\textwidth}  % 图片3部分
		\centering
		\includegraphics[width=\textwidth]{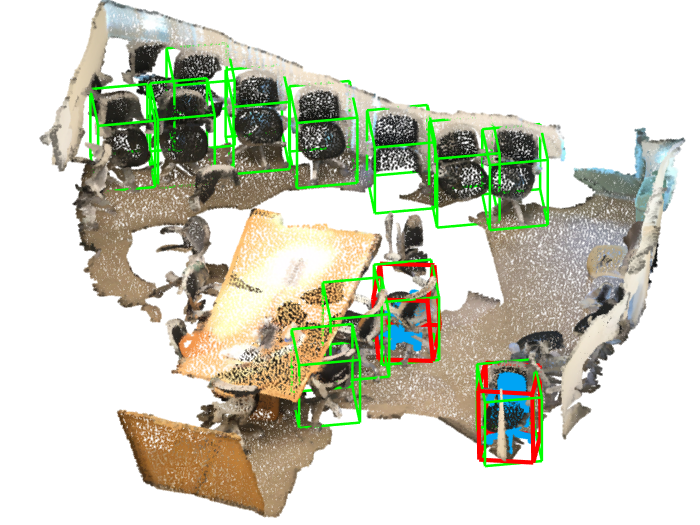}
	\end{minipage}
\end{figure}

\vspace{-10mm}

\begin{figure}[H]
	\centering
	\begin{minipage}{0.2\textwidth}  % 文本部分
		\centering
		RansaCov\cite{bib15}
	\end{minipage}%
	\begin{minipage}{0.26\textwidth}  % 图片1部分
		\centering
		\includegraphics[width=\textwidth]{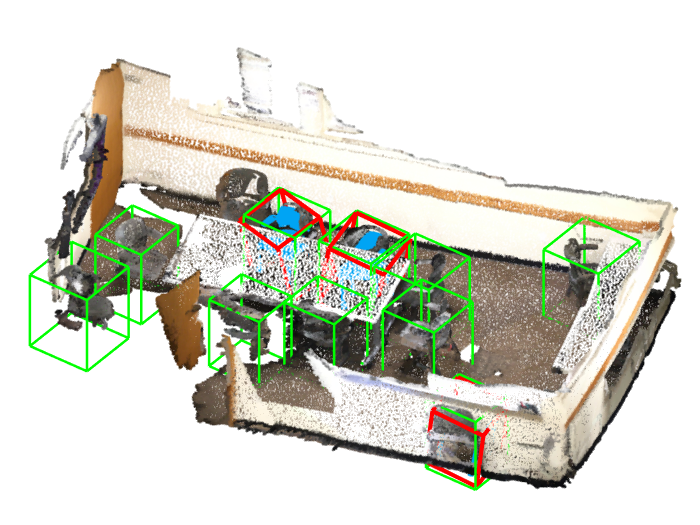}
	\end{minipage}%
	\begin{minipage}{0.26\textwidth}  % 图片2部分
		\centering
		\includegraphics[width=\textwidth]{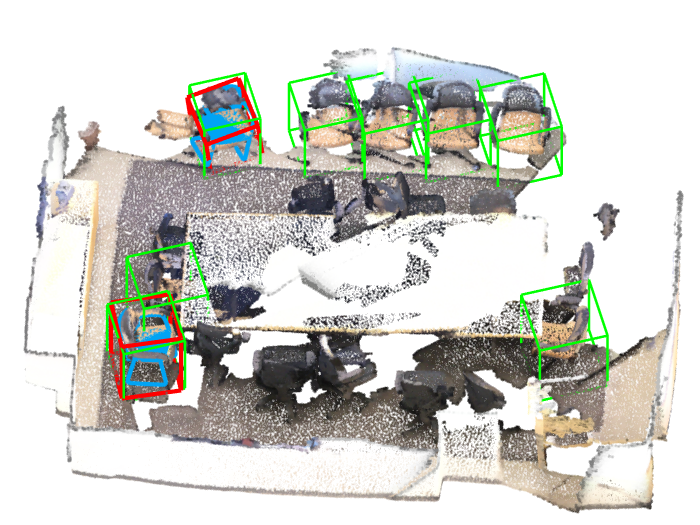}
	\end{minipage}%
	\begin{minipage}{0.26\textwidth}  % 图片3部分
		\centering
		\includegraphics[width=\textwidth]{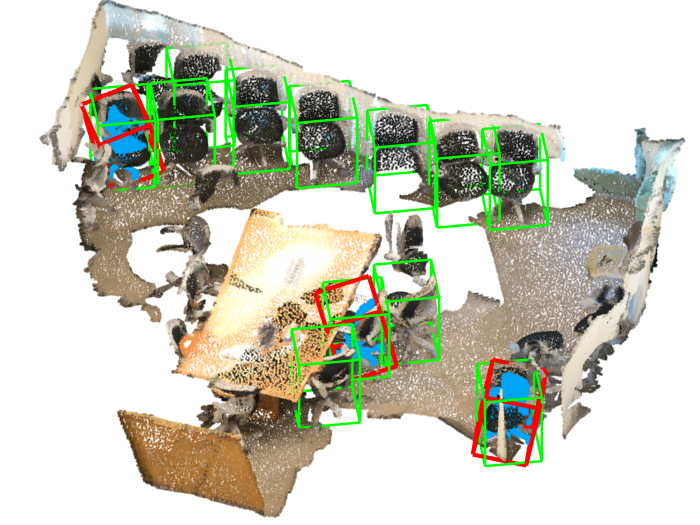}
	\end{minipage}
\end{figure}

\vspace{-10mm}

\begin{figure}[H]
	\centering
	\begin{minipage}{0.2\textwidth}  % 文本部分
		\centering
		CONSAC\cite{bib16}
	\end{minipage}%
	\begin{minipage}{0.26\textwidth}  % 图片1部分
		\centering
		\includegraphics[width=\textwidth]{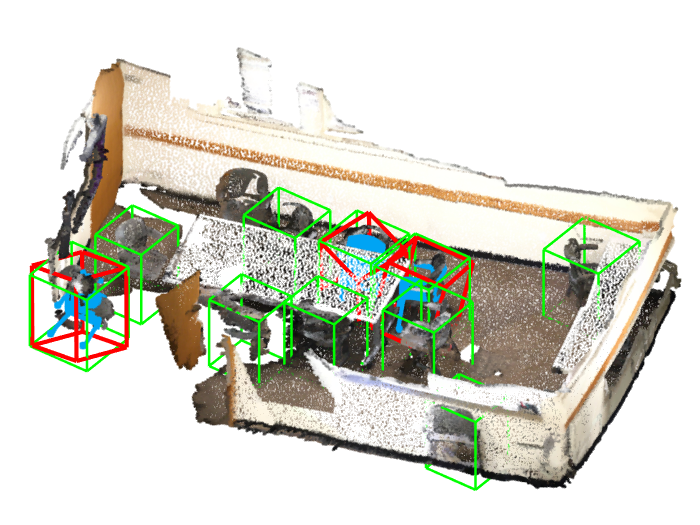}
	\end{minipage}%
	\begin{minipage}{0.26\textwidth}  % 图片2部分
		\centering
		\includegraphics[width=\textwidth]{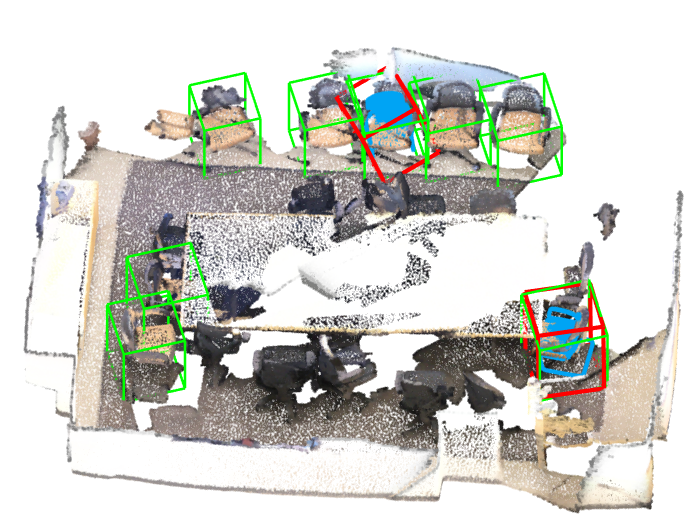}
	\end{minipage}%
	\begin{minipage}{0.26\textwidth}  % 图片3部分
		\centering
		\includegraphics[width=\textwidth]{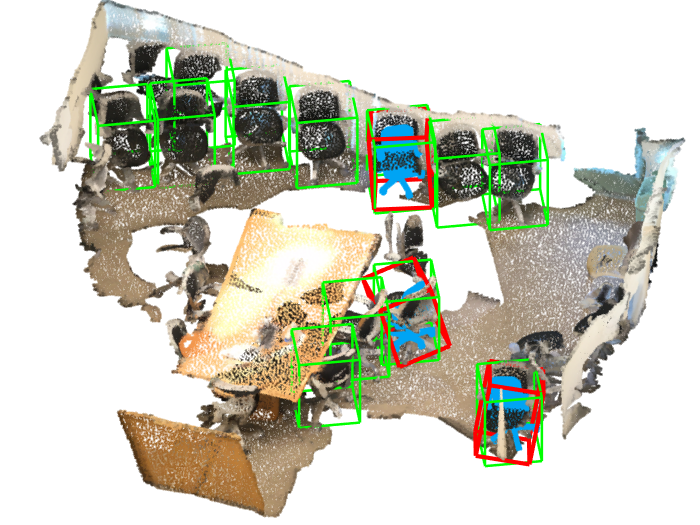}
	\end{minipage}
\end{figure}

\vspace{-10mm}

\begin{figure}[H]
	\centering
	\begin{minipage}{0.2\textwidth}  % 文本部分
		\centering
		ECC\cite{bib19}
	\end{minipage}%
	\begin{minipage}{0.26\textwidth}  % 图片1部分
		\centering
		\includegraphics[width=\textwidth]{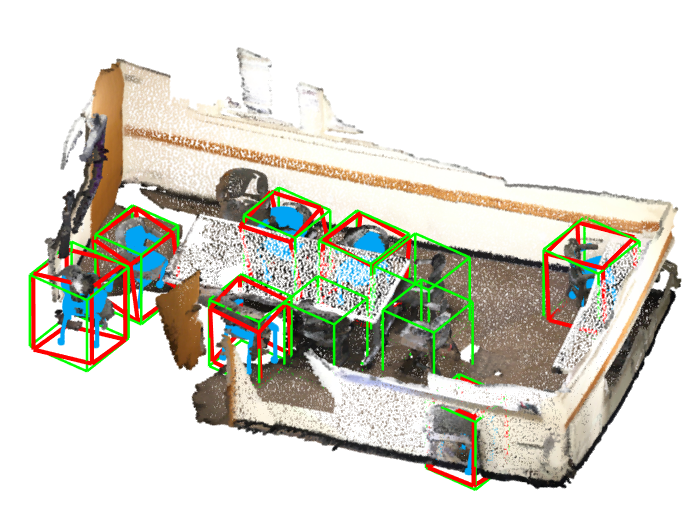}
	\end{minipage}%
	\begin{minipage}{0.26\textwidth}  % 图片2部分
		\centering
		\includegraphics[width=\textwidth]{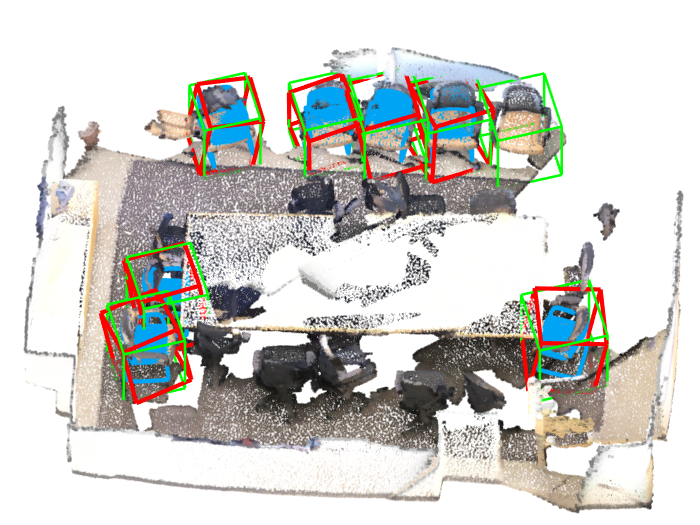}
	\end{minipage}%
	\begin{minipage}{0.26\textwidth}  % 图片3部分
		\centering
		\includegraphics[width=\textwidth]{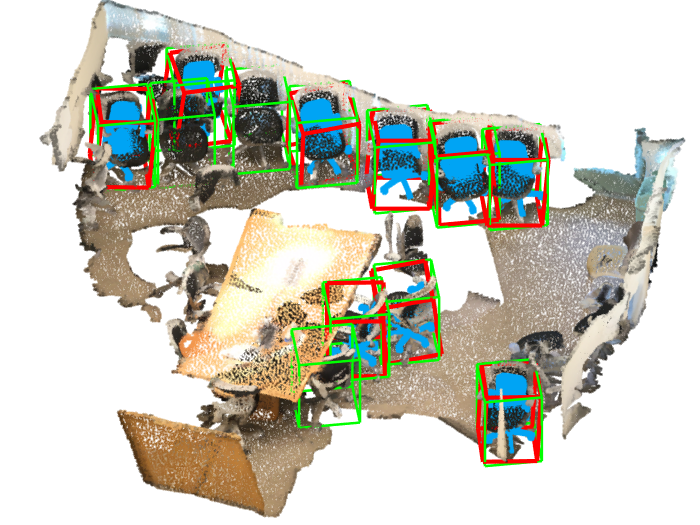}
	\end{minipage}
\end{figure}

\vspace{-10mm}

\begin{figure}[H]
	\centering
	\begin{minipage}{0.2\textwidth}  % 文本部分
		\centering
		PointCLM\cite{bib33}
	\end{minipage}%
	\begin{minipage}{0.26\textwidth}  % 图片1部分
		\centering
		\includegraphics[width=\textwidth]{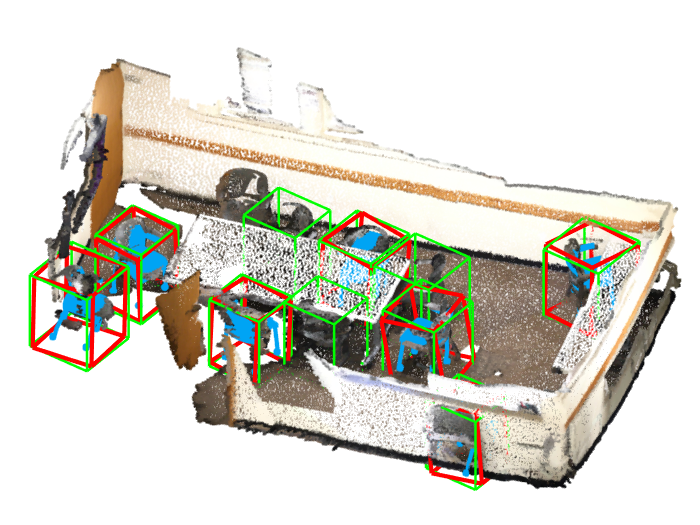}
	\end{minipage}%
	\begin{minipage}{0.26\textwidth}  % 图片2部分
		\centering
		\includegraphics[width=\textwidth]{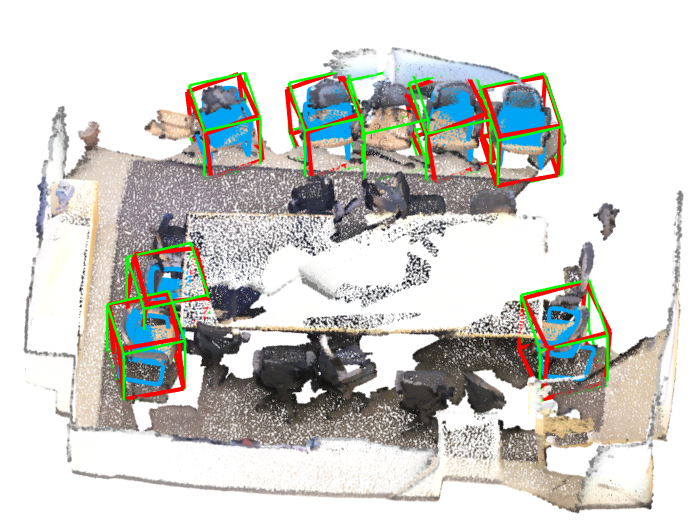}
	\end{minipage}%
	\begin{minipage}{0.26\textwidth}  % 图片3部分
		\centering
		\includegraphics[width=\textwidth]{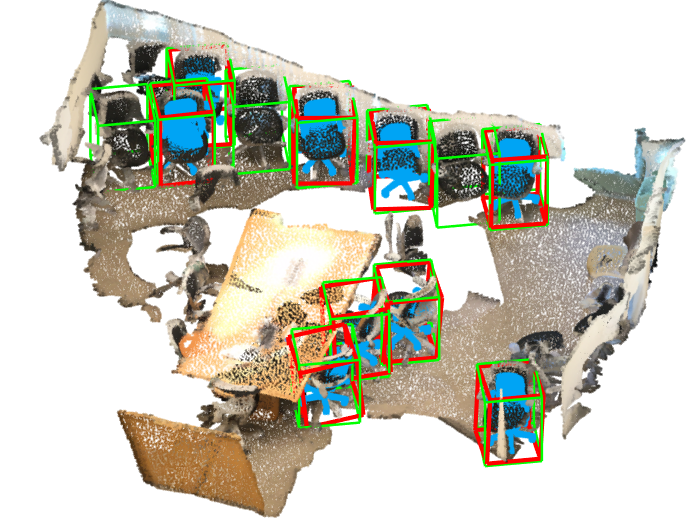}
	\end{minipage}
\end{figure}

\vspace{-10mm}

\subsection{Ablation Experiments}
\begin{figure}[htbp]
	\centering
	\includegraphics[width=\textwidth]{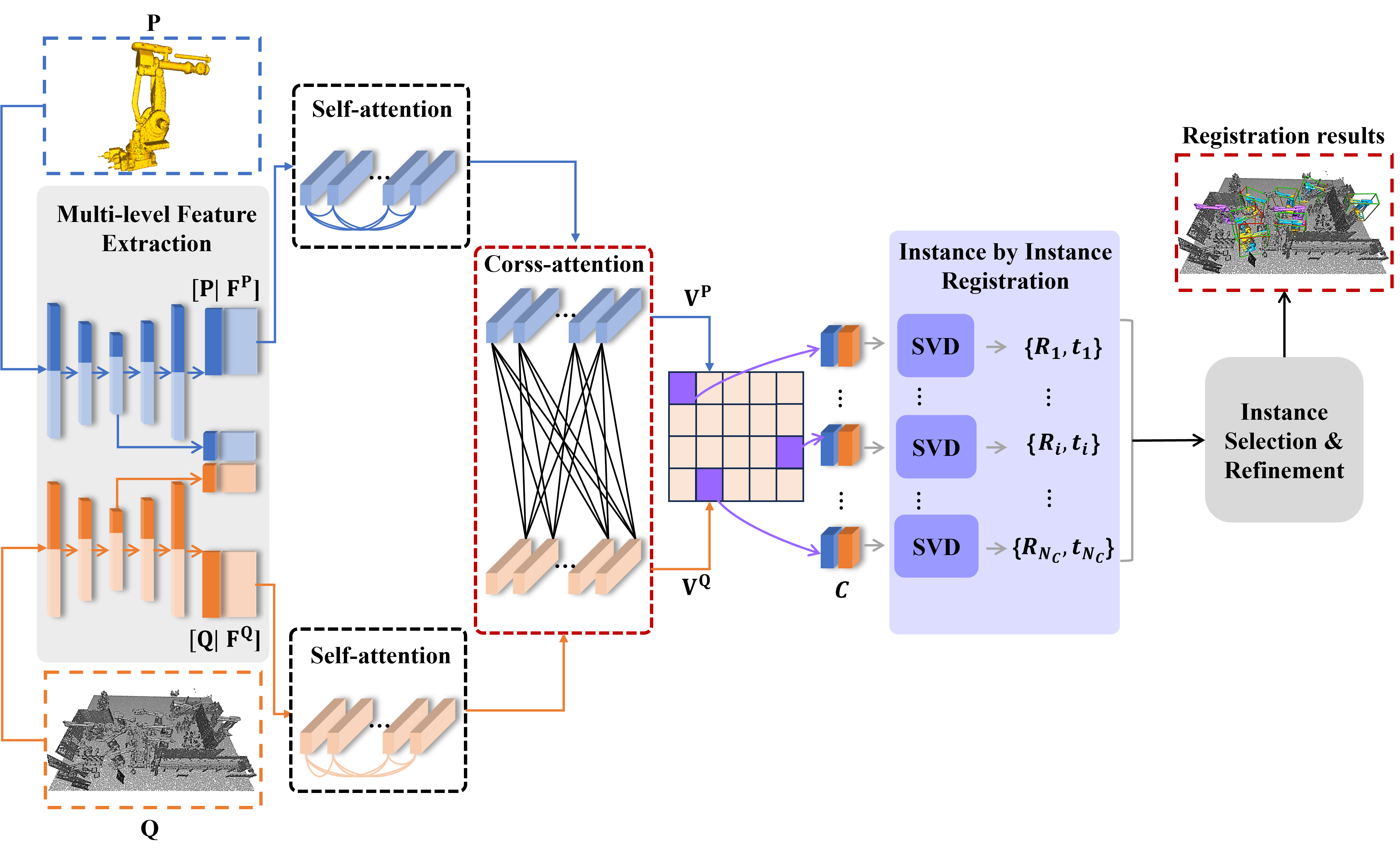}
	\caption{Initial backbone network structure.}
	\label{network_backbone}
\end{figure}

\textbf{Experiments to Verify Each Module.} We conducted a series of ablation experiments using the Welding-Station dataset to assess the individual contributions of each component within the proposed MRG framework. Our primary objective was to evaluate the impact of specific modules on registration accuracy. We began with a baseline network architecture (Fig. \ref{network_backbone}) and progressively integrated three essential modules: the regional association module, the neighbor mask module, and the instance hypothesis generation module (as described in Section \ref{method_subsec3}). This stepwise method allowed us to quantify the contribution of each module, thereby validating the design choices made in our methodology.

Table \ref{tab:ablate_each_module} presents the registration accuracy of different combinations of the backbone network and modules on the Welding-Station dataset. The results indicate a consistently positive trend in the impact of the modules on registration accuracy. Specifically, the combination of the regional association module and the neighbor mask module yields the greatest improvement in accuracy during the coarse correspondence extraction phase. Registration accuracy increased by 22.02\%, 25.58\%, and 23.86\% for the three metrics (MR, MP, and MF), respectively. This enhancement is attributed to the iterative process in which the neighbor mask module accurately predicts instance boundaries, while the regional association module strengthens the relationships between local regions of the instance. As a result, the extracted features of the target instance exhibit greater distinguishability, enhancing the ability to differentiate instances with similar local structures and extract precise coarse correspondences. Furthermore, the instance hypothesis generation module contributes to retaining most of the complex instance geometry during the transition from coarse to dense correspondences, thereby improving registration accuracy during the iterative pose estimation process. This module resulted in improvements of 3.7\%, 3.19\%, and 3.48\% in registration accuracy for the MR, MP, and MF metrics, respectively.

\begin{table}[!ht]
	\caption{Registration results of the backbone network and different module combinations. RAM represents the regional association module, NMM represents neighbor mask module, IHG represents instance hypothesis generation module.}
	\label{tab:ablate_each_module}
	\small
	\begin{tabularx}{\textwidth}{>{\centering\arraybackslash}l|>{\centering\arraybackslash}X>{\centering\arraybackslash}X>{\centering\arraybackslash}X>{\centering\arraybackslash}X>{\centering\arraybackslash}X>{\centering\arraybackslash}X>{\centering\arraybackslash}X>{\centering\arraybackslash}X}
		\toprule
		ID & Backbone & RAM & NMM & IHG & MR(\%)$\uparrow$         & MP(\%)$\uparrow$         & MF(\%)$\uparrow$         & Time(s)$\downarrow$       \\ \midrule
		1  & \checkmark  &     &     &     & 56.35          & 52.33          & 54.27          & \textbf{0.51} \\ 
		2  & \checkmark  & \checkmark  &     &     & 65.36          & 61.58          & 63.41          & $\underline{0.83}$ \\ 
		3  & \checkmark  &     & \checkmark  &     & 68.89          & 66.38          & 67.61          & 0.98          \\ 
		4  & \checkmark  & \checkmark  & \checkmark  &     & $\underline{72.26}$ & $\underline{70.32}$ & $\underline{71.28}$ & 1.10          \\ 
		5  & \checkmark  & \checkmark  & \checkmark  & \checkmark  & \textbf{75.10} & \textbf{72.64} & \textbf{73.85} & 1.33          \\ 
		\bottomrule
	\end{tabularx}
\end{table}
\textbf{Validate the Advantages of the Instance-Focused Transformer.} The results of this experiment demonstrate that using the Instance-Focused Transformer (IFT) module throughout the entire network yields significant advantages. As shown in Table \ref{ablation_transformer_module}, with the integration of IFT module, MRG’s mean recall rate on the Welding-Station dataset increased by 29.87\%, the mean precision increased by 22.49\%, and the mean F1 score increased by 26.31\%. This is because, in the absence of the IFT, the features of the target instance are contaminated by surrounding noise, resulting in an increased number of mismatched points in the extracted coarse correspondences. Furthermore, additional noise is introduced when extending to dense correspondences, leading to poorer registration results.
\begin{table}[!ht]
	\caption{Ablation experiments to validate the Instance-focused transformer.}
	\label{ablation_transformer_module}
	\small
	\begin{tabularx}{\textwidth}{>{\centering\arraybackslash}l|>{\centering\arraybackslash}X>{\centering\arraybackslash}X>{\centering\arraybackslash}X}
		\toprule
		\multicolumn{1}{c|}{Model}           & MR(\%)$\uparrow$         & MP(\%)$\uparrow$         & MF(\%)$\uparrow$         \\ \midrule
		MRG with IFT    & \textbf{75.10} & \textbf{72.64} & \textbf{73.85} \\ 
		MRG without IFT & 52.67          & 56.30          & 54.42          \\ 
		\bottomrule
	\end{tabularx}
\end{table}
Secondly, to explore the effect of geometric embedding on the registration results within the IFT, we investigated the geometric embedding of the neighbor mask and self-attention modules. As shown in Table \ref{tab:ablate_embedding}, GEO represents geometric embedding as described in Reference \cite{bib14}, GDE represents geodesic embedding, and AGE represents aggregate geometric embedding. When combined with GEO and AGE, MRG’s mean recall on the Welding-Station dataset increased by 12.98\%, mean precision increased by 17.24\%, and mean F1 score increased by 15.19\%. This improvement is attributed to AGE integrating geodesic distance, curvature, and normal vector information, which more accurately represents the geometric features of robot instance surfaces.
\begin{table}[!ht]
	\caption{Registration results of different geometric embedding.}
	\label{tab:ablate_embedding}
	\small
	\begin{tabularx}{\textwidth}{>{\centering\arraybackslash}X|>{\centering\arraybackslash}X>{\centering\arraybackslash}X>{\centering\arraybackslash}X}
		\toprule
		Model           & MR(\%)$\uparrow$         & MP(\%)$\uparrow$         & MF(\%)$\uparrow$         \\ \midrule
		None \& None    & 65.35          & 60.12          & 62.63          \\ 
		GEO \& None     & 68.59          & 67.66          & 68.12          \\ 
		GEO \& GEO      & 70.03          & 68.35          & 69.18          \\ 
		GEO \& GDE      & 71.08          & 69.59          & 70.33          \\ 
		GDE \& GDE      & 72.19          & 70.39          & 71.28          \\ 
		GDE \& AGE      & 74.13          & 72.00          & 73.05          \\ 
		AGE \& AGE      & $\underline{74.96}$    & $\underline{72.06}$    & $\underline{73.48}$    \\ 
		GEO \& AGE      & \textbf{75.10} & \textbf{72.64} & \textbf{73.85} \\ 
		\bottomrule
	\end{tabularx}
\end{table}

Finally, to select the optimal neighborhood range for accurately predicting the neighbor mask, we conducted experiments by varying the number of $K$-neighbors within the module. As shown in Table \ref{tab:neighbors_num}, decreasing the number of neighbors leads to a gradual decline in model performance, particularly affecting the MP metric. Reducing the number of neighbors causes the instance hypothesis generation module to generate more instances but introduces more false matches, resulting in an increase in MP and a decrease in registration accuracy. Conversely, increasing the number of neighbors weakens the module’s ability to sense and predict masks, leading to decreased registration accuracy.
\begin{table}[!ht]
	\caption{Ablation experiments of the number of neighbors on Welding-Station dataset.}
	\label{tab:neighbors_num}
	\small
	\begin{tabularx}{\textwidth}{>{\centering\arraybackslash}X|>{\centering\arraybackslash}X>{\centering\arraybackslash}X>{\centering\arraybackslash}X}
		\toprule
		Neighbors           & MR(\%)$\uparrow$         & MP(\%)$\uparrow$         & MF(\%)$\uparrow$         \\ \midrule
		4               & 76.29          & 62.98          & 69.00          \\ 
		8               & \textbf{77.53} & 65.33          & 70.91          \\ 
		16              & 76.69          & 69.39          & 72.86          \\ 
		32              & 75.10          & \textbf{72.64} & \textbf{73.85} \\ 
		48              & 74.79          & $\underline{71.05}$    & $\underline{72.87}$    \\ 
		64              & 74.50          & 70.60          & 72.50          \\ 
		\bottomrule
	\end{tabularx}
\end{table}

\textbf{Validate the Design Advantages of the Instance Hypothesis Generation.} The results of this experiment demonstrate the benefits of utilizing the Instance Hypothesis Generation (IHG) module across the entire network. As shown in Table \ref{tab:ablate_IHG}, when integrated with the IHG, MRG’s mean recall, mean precision, and mean F1 score on the Welding-Station dataset increased by 3.34\%, 4.36\%, and 3.86\%, respectively. This indicates that retaining additional geometric feature information helps improve registration accuracy.

\begin{table}[!ht]
	\caption{Ablation experiments to validate the Instance Hypothesis Generation.}
	\label{tab:ablate_IHG}
	\small
	\begin{tabularx}{\textwidth}{>{\centering\arraybackslash}l|>{\centering\arraybackslash}X>{\centering\arraybackslash}X>{\centering\arraybackslash}X}
		\toprule
		\multicolumn{1}{c|}{Model}           & MR(\%)$\uparrow$         & MP(\%)$\uparrow$         & MF(\%)$\uparrow$         \\ \midrule
		MRG with IHG    & \textbf{75.10} & \textbf{72.64} & \textbf{73.85} \\ 
		MRG without IHG & 72.59          & 69.47          & 71.00          \\ 
		\bottomrule
	\end{tabularx}
\end{table}
%\vspace{-2mm}
% scene_1
\begin{figure}[H]
	\centering
	\begin{minipage}{\textwidth}
		\centering
		\includegraphics[width=\textwidth]{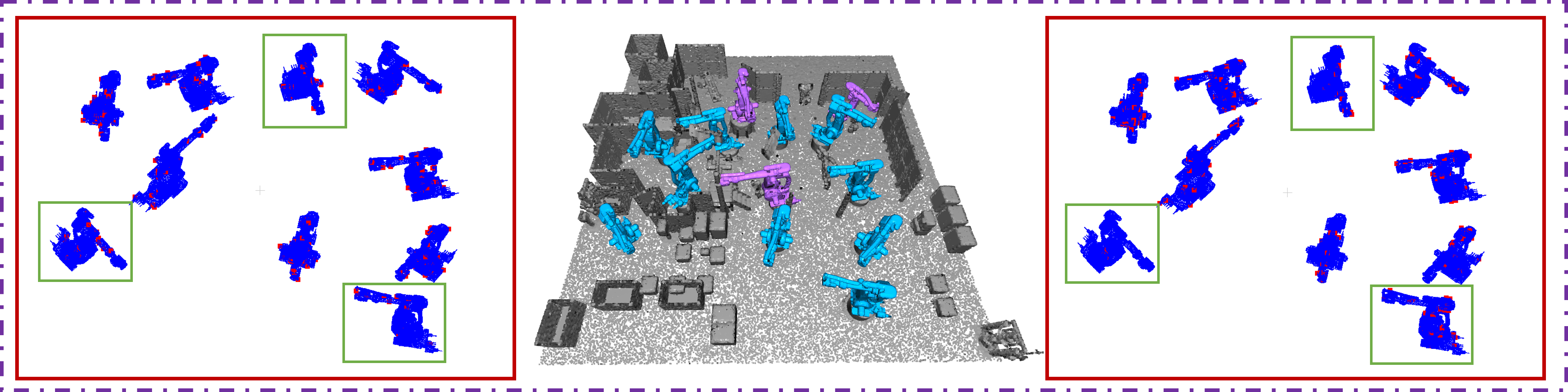}
		
		(a) % 放置在图片正下方
	\end{minipage}
\end{figure}
% scene_2
\vspace{-2mm}
\begin{figure}[H]
	\centering
	\begin{minipage}{\textwidth}  % 文本部分
		\centering
		\includegraphics[width=\textwidth]{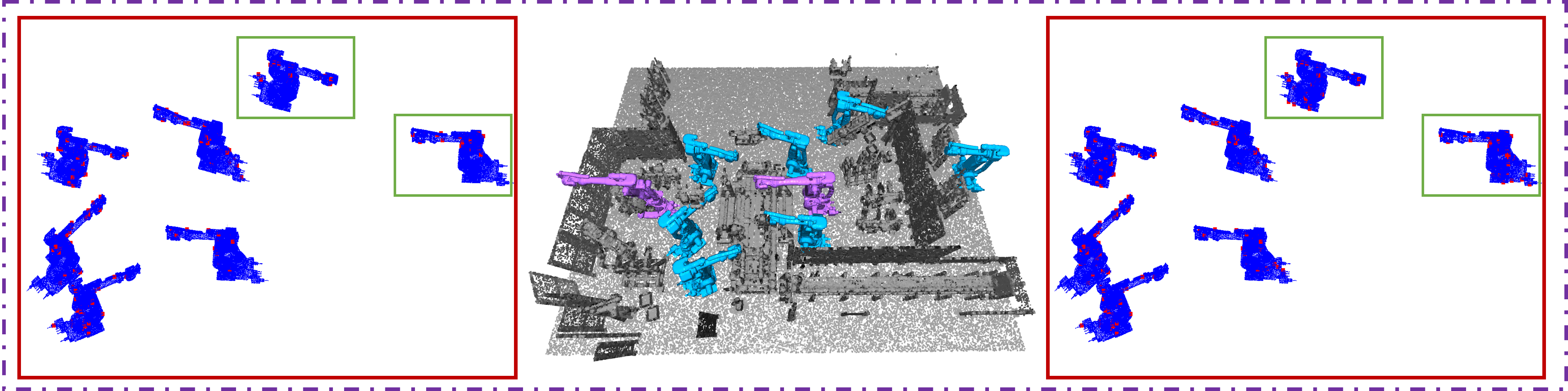}
		
		(b)
	\end{minipage}
\end{figure}
% scene_3
\vspace{-2mm}
\begin{figure}[H]
	\centering
	\begin{minipage}{\textwidth}  % 文本部分
		\centering
		\includegraphics[width=\textwidth]{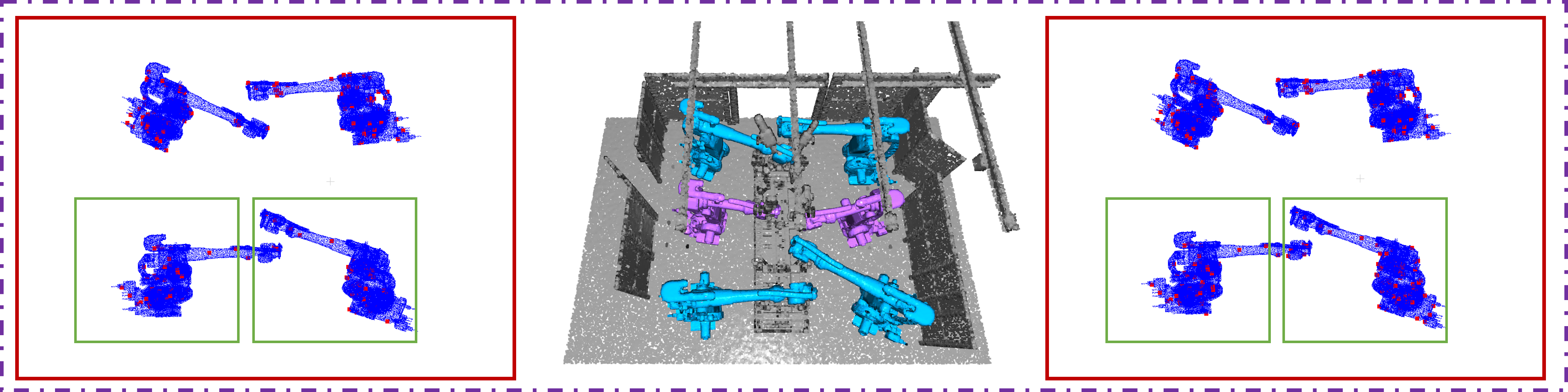}
		
		(c)
	\end{minipage}
\end{figure}
\vspace{-2mm}
% scene_4
\begin{figure}[H]
	\centering
	\begin{minipage}{\textwidth}  % 文本部分
		\centering
		\includegraphics[width=\textwidth]{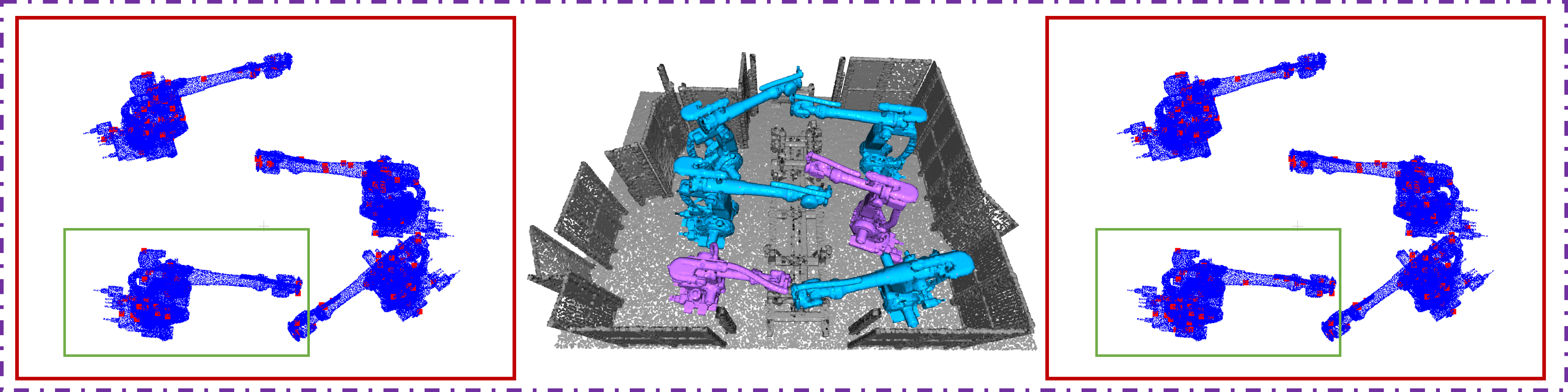}
		
		(d)
	\end{minipage}
\end{figure}
\vspace{-2mm}
% scene_5
\begin{figure}[H]
	\centering
	\begin{minipage}{\textwidth}  % 文本部分
		\centering
		\includegraphics[width=\textwidth]{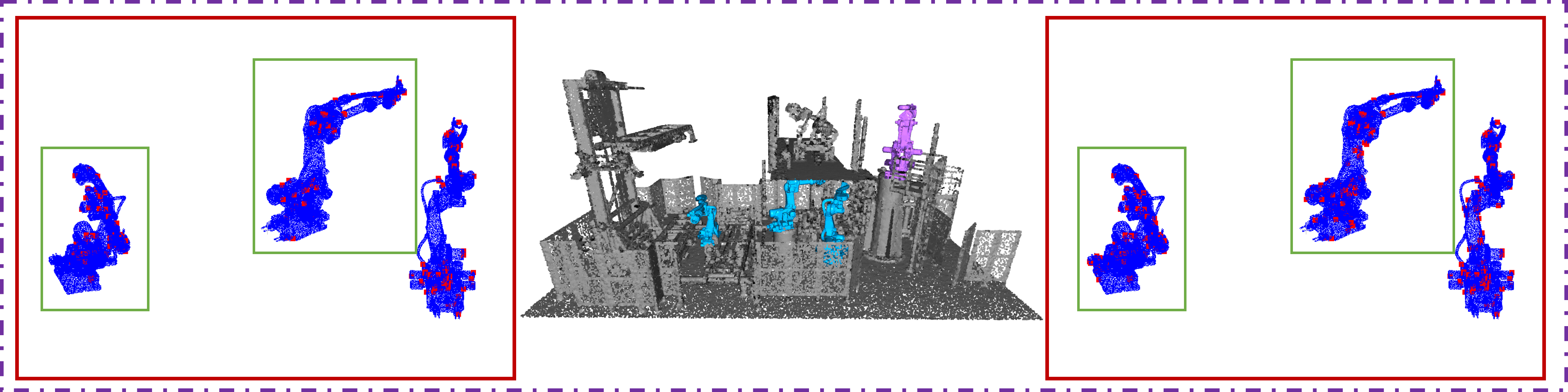}
		
		(e)
	\end{minipage}%
	\caption{Results of various coarse correspondence extraction methods on the Welding-Station dataset. The left is based on TOP-K, the center is raw scene, and the right is MRG. The red points represent coarse correspondence points.}
	\label{fig:comparison_coarse_methods}
\end{figure}
% 此处需要重新改图

As shown in Figure \ref{fig:comparison_coarse_methods}, we present the extraction process of coarse correspondences for robot instances in different welding stations. As the number of instances increases, the geometric features extracted by the TOP-K method become increasingly insufficient, thereby compromising pose estimation accuracy. The IHG module integrates neighbor mask information into the coarse correspondence extraction process. This ensures that keypoints are evenly distributed across each instance, thereby ensuring sufficient geometric information for accurate pose estimation.
%\vspace{-2mm}
% scene_1

\textbf{Validate the Design Advantages of the Instance Filtering and Optimization.}To quantitatively assess the advantages of the proposed instance filtering and optimization module in complex point cloud registration tasks, comparative experiments were performed. In the MRG framework, the instance filtering and optimization module was replaced with three alternative methods: the spectral clustering method from PointCLM \cite{bib33}, the distance-invariance clustering from ECC \cite{bib19}, and multi-model fitting techniques \cite{bib15, bib16, bib17}. The comparative results are presented in Table \ref{tab:different_model_fitting}.

\begin{table}[!ht]
	\caption{Registration results of different multi-model fitting methods combined with MRG point matching.}
	\label{tab:different_model_fitting}
	\small
	\begin{tabularx}{\textwidth}{l|>{\centering\arraybackslash}X>{\centering\arraybackslash}X>{\centering\arraybackslash}X>{\centering\arraybackslash}X}
		\toprule
		\multicolumn{1}{c|}{Method}    & MR(\%)$\uparrow$       & MP(\%)$\uparrow$       & MF(\%)$\uparrow$       & Time(s)$\downarrow$       \\ \midrule
		MRG + T-linkage\cite{bib17}                  & 31.56                   & 40.42                   & 35.44                   & 5.90                      \\
		MRG + RansaCov\cite{bib15}                   & 52.96                   & 41.25                   & 46.38                   & \underline{0.98}          \\
		MRG + CONSAC\cite{bib16}                     & 54.36                   & 50.79                   & 52.51                   & \textbf{0.65}             \\
		MRG + PointCLM\cite{bib33}                   & 69.59                   & 68.37                   & 68.97                   & 1.56                      \\
		MRG + ECC\cite{bib19}                        & \textbf{76.35}                   & \underline{69.16}       & \underline{72.58}       & 3.00                      \\
		MRG(full pipeline)             & \underline{75.10}       & \textbf{72.64}          & \textbf{73.85}          & 1.33                      \\
		\bottomrule
	\end{tabularx}
\end{table}

As shown in Table \ref{tab:different_model_fitting}, the proposed full pipeline significantly outperforms all baseline multi-model fitting methods. Specifically, the proposed method achieves average improvements of 24.15\%, 25.66\%, and 25.29\% in the MR, MP, and MF metrics, respectively. Although the ECC-based model demonstrates slightly better performance in MR compared to the proposed full pipeline, its performance in MP is substantially weaker. This observation suggests that ECC attains a higher MR by predicting an excessive number of registrations. This behavior consistently outperforms PointCLM across all three registration metrics, underscoring its robustness and effectiveness.

Secondly, we analyzed instance filtering strategies involving outlier removal using two methods: “point-to-point” and “point-to-plane”. As shown in Table \ref{different_filtering_strategies}, compared to the “point-to-point” strategy, the “point-to-plane” strategy demonstrated improvements of 1.88\%, 4.56\%, and 3.26\% in the MR, MP, and MF metrics, respectively. This suggests that the “point-to-plane” strategy is more effective at mitigating errors induced by noise and outliers during registration, thereby enhancing overall matching quality. Notably, although the “point-to-plane” strategy considers the local fitting degree when filtering points with large errors, its computational complexity remains comparable to that of the “point-to-point” strategy. As a result, its time consumption aligns with that of the “point-to-point” strategy.

\begin{table}[!ht]
	\caption{Registration results on different filtering strategies. The point-to-point strategy removes corresponding point pairs with distances smaller than a predefined distance threshold, while point-to-plane strategy excludes pairs where the nearest distance from the transformed source point cloud to the target point is below the threshold.}
	\label{different_filtering_strategies}
	\small
	\begin{tabularx}{\textwidth}{l|>{\centering\arraybackslash}X>{\centering\arraybackslash}X>{\centering\arraybackslash}X>{\centering\arraybackslash}X}
		\toprule
		Strategy            & MR(\%)$\uparrow$       & MP(\%)$\uparrow$       & MF(\%)$\uparrow$       & Time(s)$\downarrow$       \\ \midrule
		point-to-point      & 73.69                   & 69.33                   & 71.44                   & \textbf{1.20}             \\
		point-to-plane      & \textbf{75.10}         & \textbf{72.64}         & \textbf{73.85}         & 1.33                      \\
		\bottomrule
	\end{tabularx}
\end{table}

\begin{figure}[H]
	\centering
	% 第三行图片
	\begin{minipage}{0.3\textwidth}
		\centering
		\includegraphics[width=\textwidth]{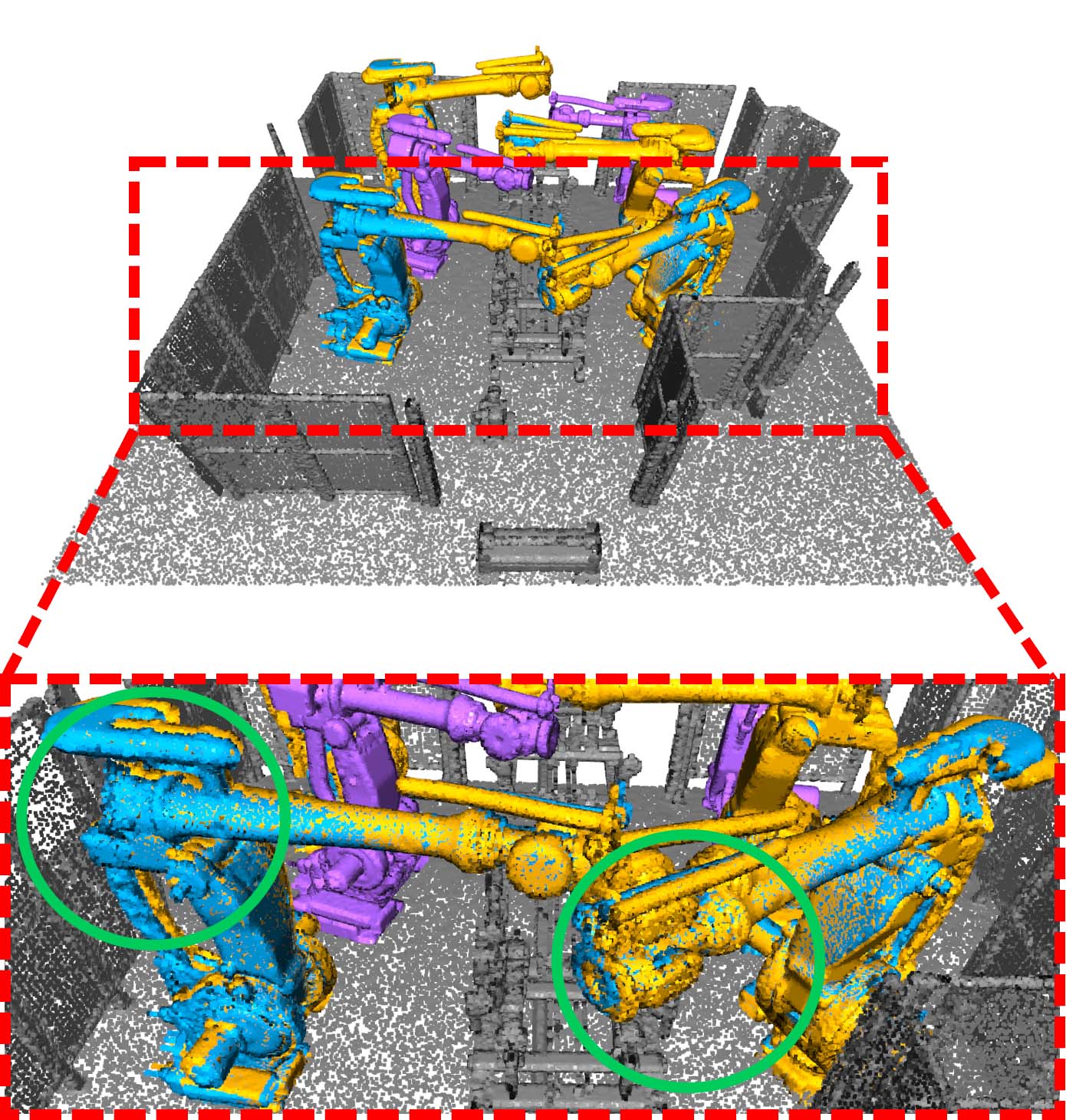}
	\end{minipage}
	\begin{minipage}{0.3\textwidth}
		\centering
		\includegraphics[width=\textwidth]{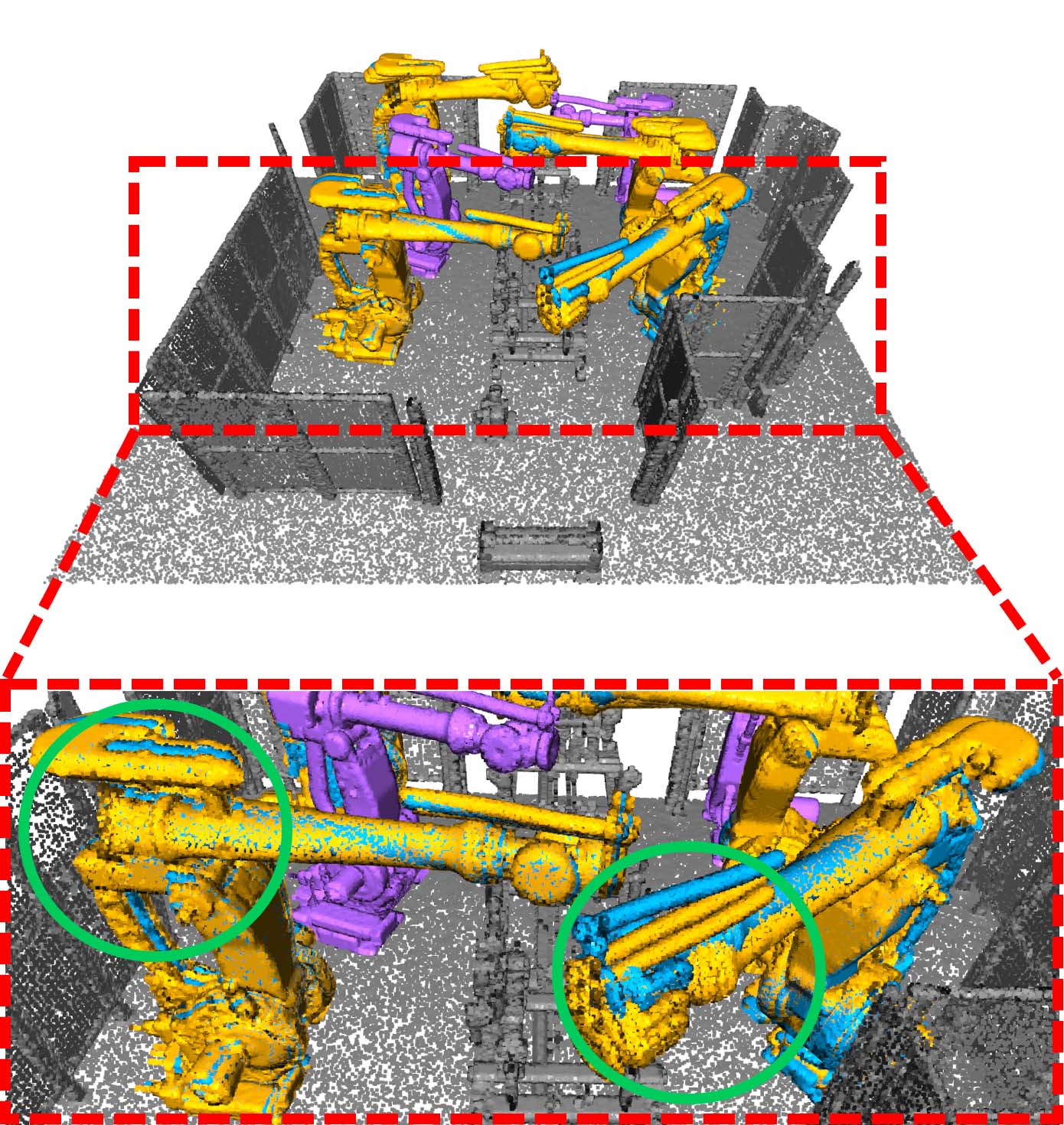}
	\end{minipage}
	\begin{minipage}{0.3\textwidth}
		\centering
		\includegraphics[width=\textwidth]{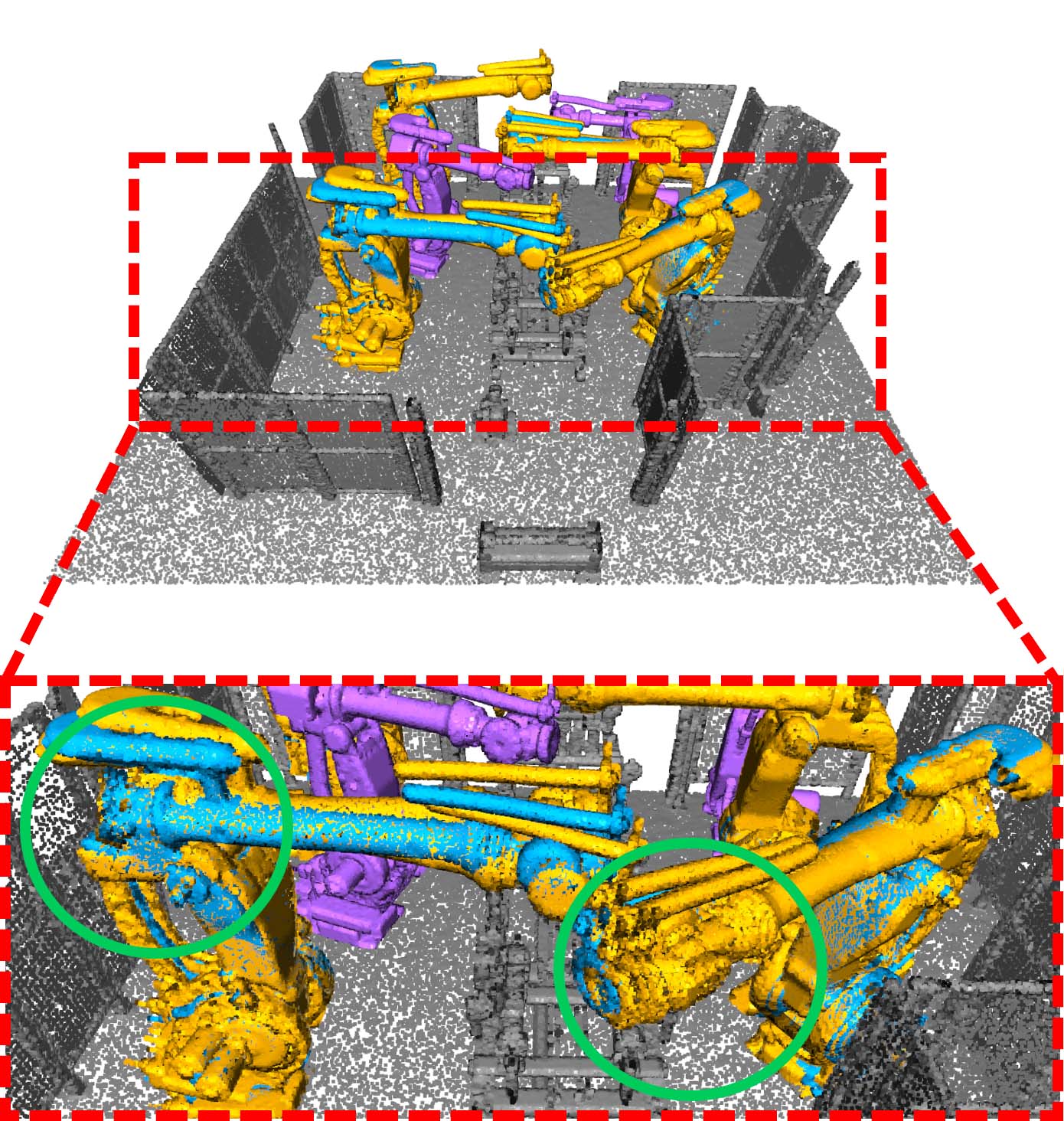}
	\end{minipage}
	
	% 第四行文本
	\vspace{5pt} % 添加一些间距
	\begin{minipage}{0.3\textwidth}
		\centering
		0.6
	\end{minipage}
	\begin{minipage}{0.3\textwidth}
		\centering
		0.5
	\end{minipage}
	\begin{minipage}{0.3\textwidth}
		\centering
		0.4
	\end{minipage}
	\caption{Visualization of registration performance on different thresholds.}
	\label{fig:visual_different_sim_thresholds}
\end{figure}

Finally, to enhance the distinction of local regions across different instances, we conducted a sensitivity analysis on the instance similarity threshold the. The threshold the defines the inclusion criteria for local regions within the same instance. As shown in Table \ref{table:different_sim_thresholds}, as the threshold the decreases from 0.9 to 0.4, MP increases by 17.57\%, while MR decreases by 22.42\%. This suggests that reducing the threshold leads to repeated detections of the same instance and multiple pose estimations. Fig. \ref{fig:visual_different_sim_thresholds} illustrates the visualization results across various threshold values.

\begin{table}[!h]
	\caption{Registration Results on Different Thresholds.}
	\label{table:different_sim_thresholds}
	\small
	\begin{tabularx}{\textwidth}{c|>{\centering\arraybackslash}X>{\centering\arraybackslash}X>{\centering\arraybackslash}X>{\centering\arraybackslash}X}
		\toprule
		$\theta$           & MR(\%)$\uparrow$       & MP(\%)$\uparrow$       & MF(\%)$\uparrow$       & Time(s)$\downarrow$       \\ \midrule
		0.9                 & \textbf{76.35}         & 69.56                   & 72.80                   & 1.33                      \\
		0.8 (Ours)          & \underline{75.10}      & 72.64                   & \textbf{73.85}         & 1.33                      \\
		0.7                 & 69.29                   & 77.39                   & \underline{73.12}      & 1.33                      \\
		0.6                 & 67.60                   & 79.98                   & \textbf{73.85}         & 1.33                      \\
		0.5                 & 63.39                   & \underline{81.10}      & 71.18                   & 1.33                      \\
		0.4                 & 59.23                   & \textbf{84.39}         & 69.61                   & 1.33                      \\
		\bottomrule
	\end{tabularx}
\end{table}
\vspace{-4mm} % 添加一些间距
\begin{figure}[H]
	\centering
	% 第一行图片
	\begin{minipage}{0.3\textwidth}
		\centering
		\includegraphics[width=\textwidth]{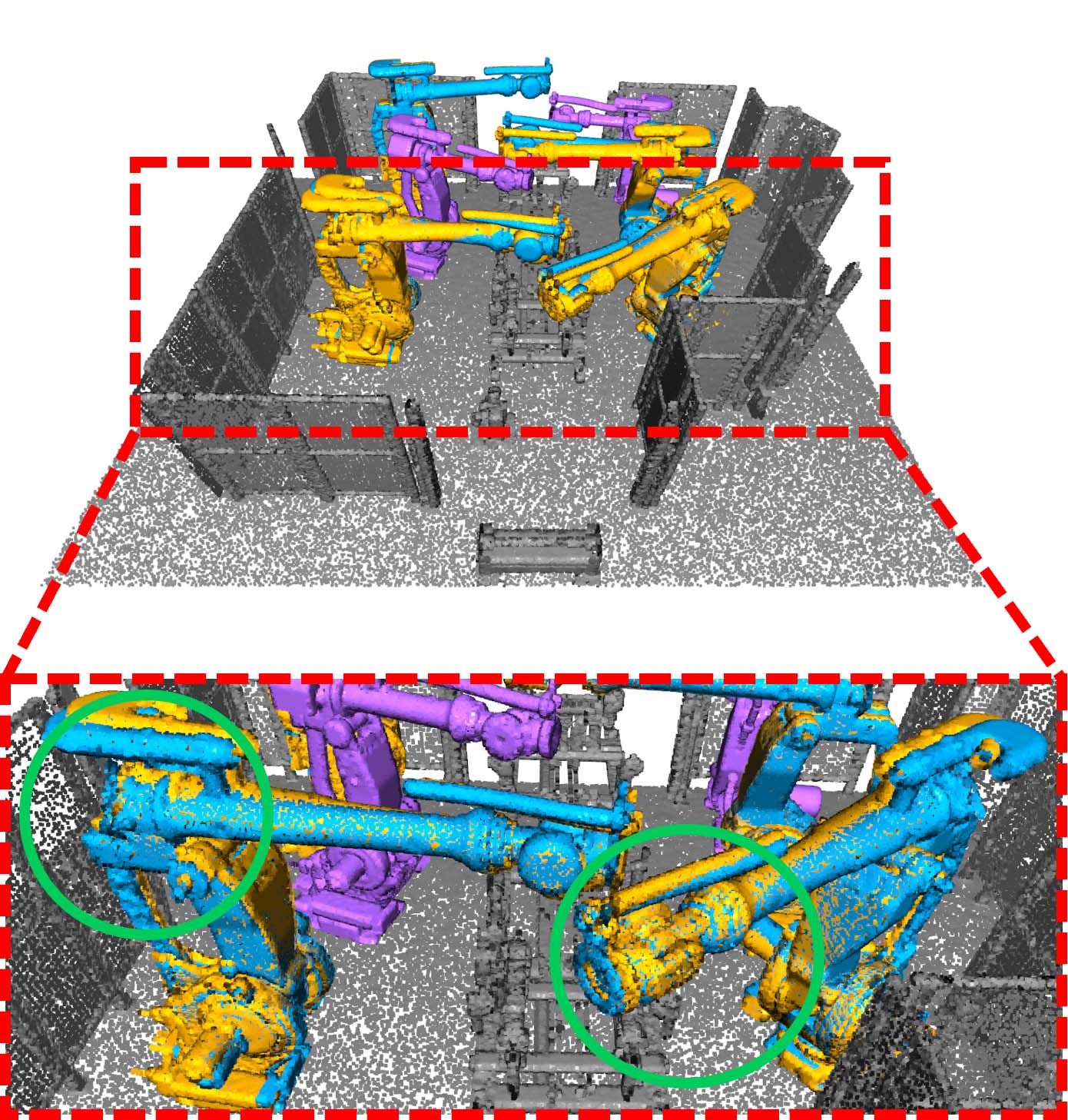}
	\end{minipage}
	\begin{minipage}{0.3\textwidth}
		\centering
		\includegraphics[width=\textwidth]{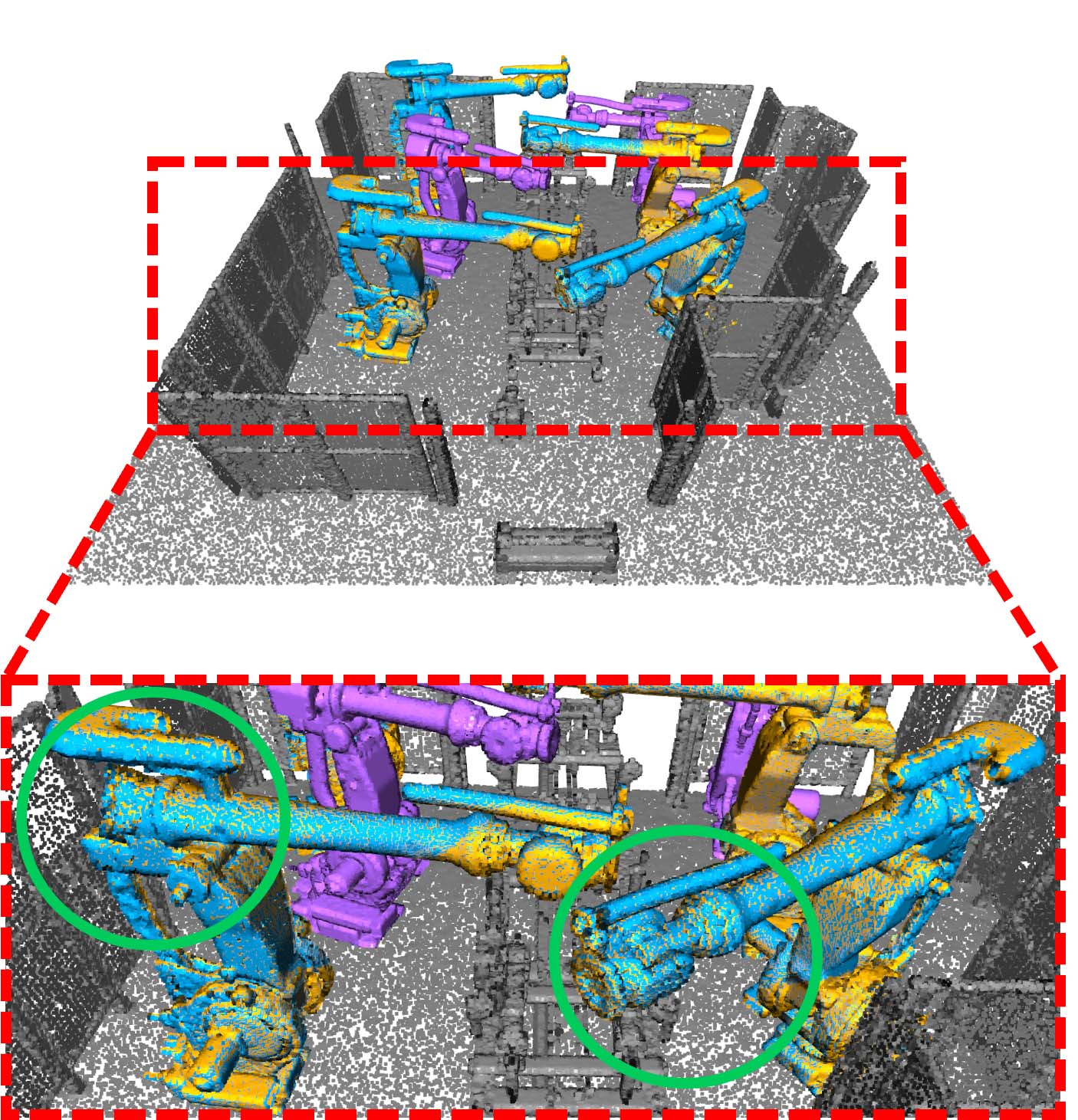}
	\end{minipage}
	\begin{minipage}{0.3\textwidth}
		\centering
		\includegraphics[width=\textwidth]{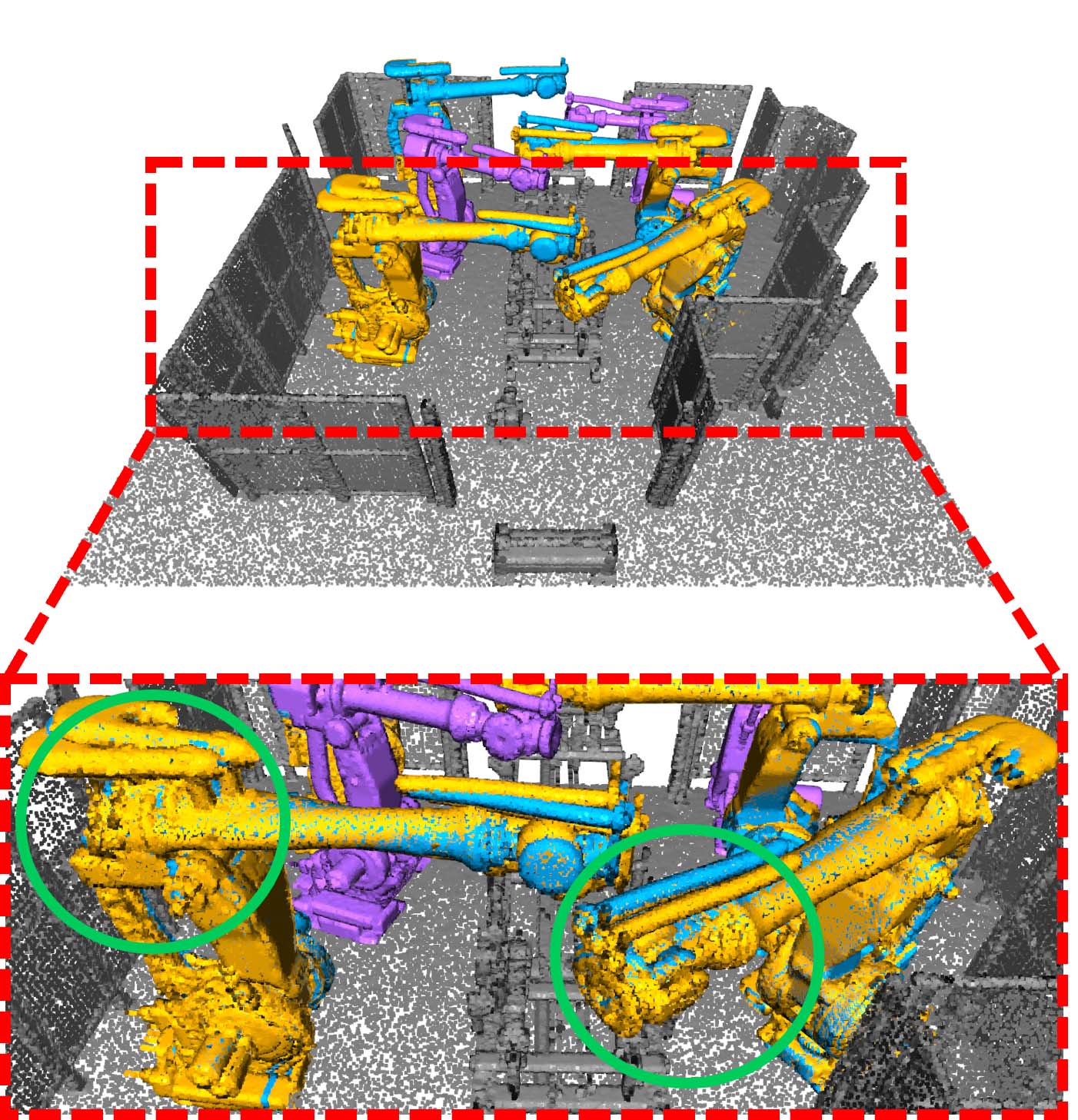}
	\end{minipage}
	
	% 第二行文本
	\vspace{5pt} % 添加一些间距
	\begin{minipage}{0.3\textwidth}
		\centering
		0.9
	\end{minipage}
	\begin{minipage}{0.3\textwidth}
		\centering
		0.8
	\end{minipage}
	\begin{minipage}{0.3\textwidth}
		\centering
		0.7
	\end{minipage}
\end{figure}
\vspace{-10mm} % 添加一些间距

\begin{figure}[H]
	\centering
	\begin{minipage}{0.7\textwidth}
		\centering
		\includegraphics[width=0.7\textwidth]{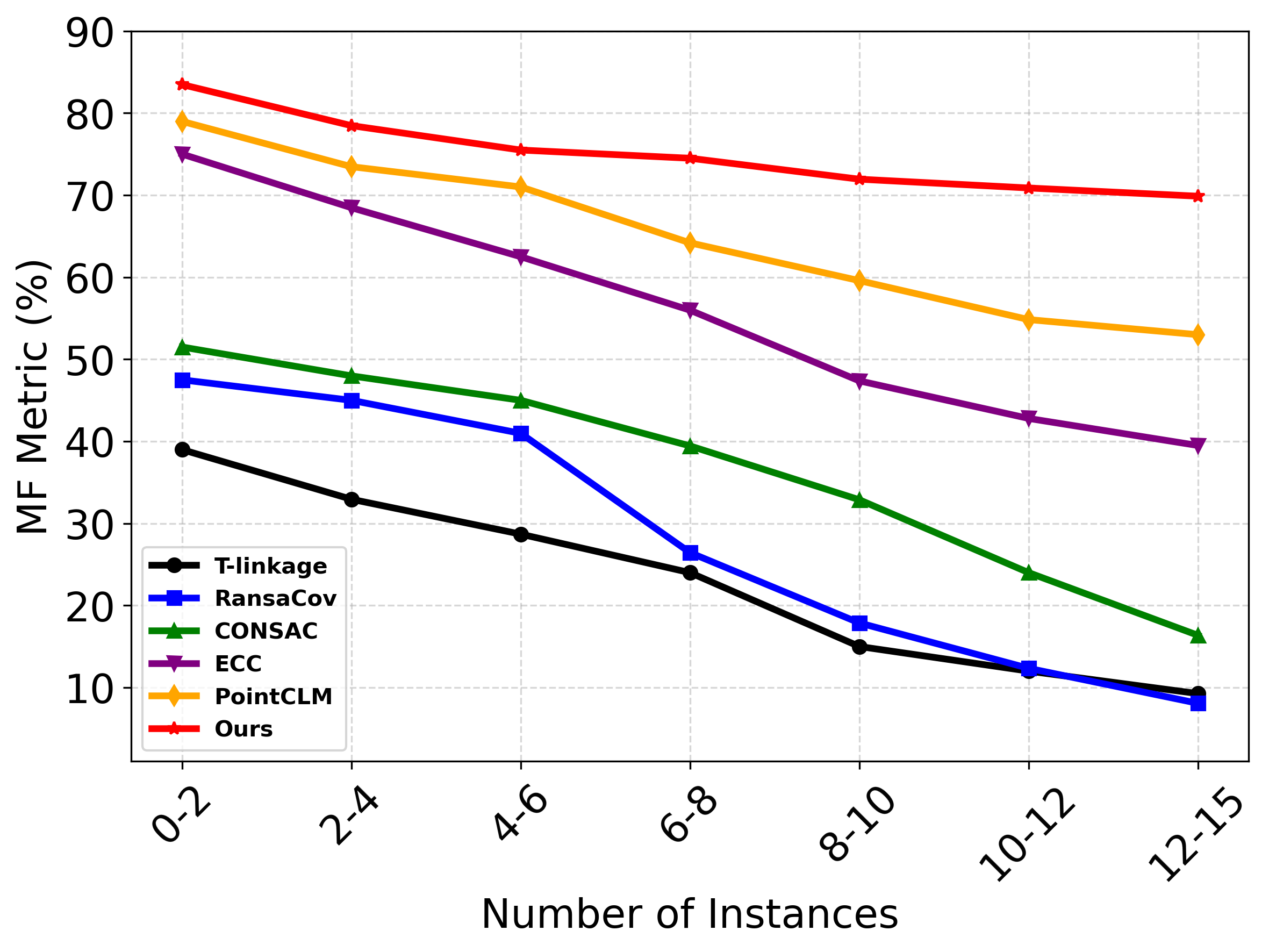}
		
		(c)
	\end{minipage}
	\caption{Registration results for different instances. (a)-(c) represents MR, MP, and MF metrics respectively.}
	\label{tab:different_num_instances}
\end{figure}

\textbf{Validate the Design of Advantages of our method in the scenes with a large number of instances.} To quantitatively assess the registration performance of MRG in scenes with a high number of instances, we compared the MR of PointCLM \cite{bib33}, ECC \cite{bib19}, and MRG methods across varying instance counts. As shown in Figure \ref{tab:different_num_instances}, the accuracy of PointCLM and ECC decreases as the number of instances increases. In contrast, MRG can still maintain high levels of accuracy.
\begin{figure}[H]
	\centering
	\begin{minipage}{0.7\textwidth}
		\centering
		\includegraphics[width=0.7\textwidth]{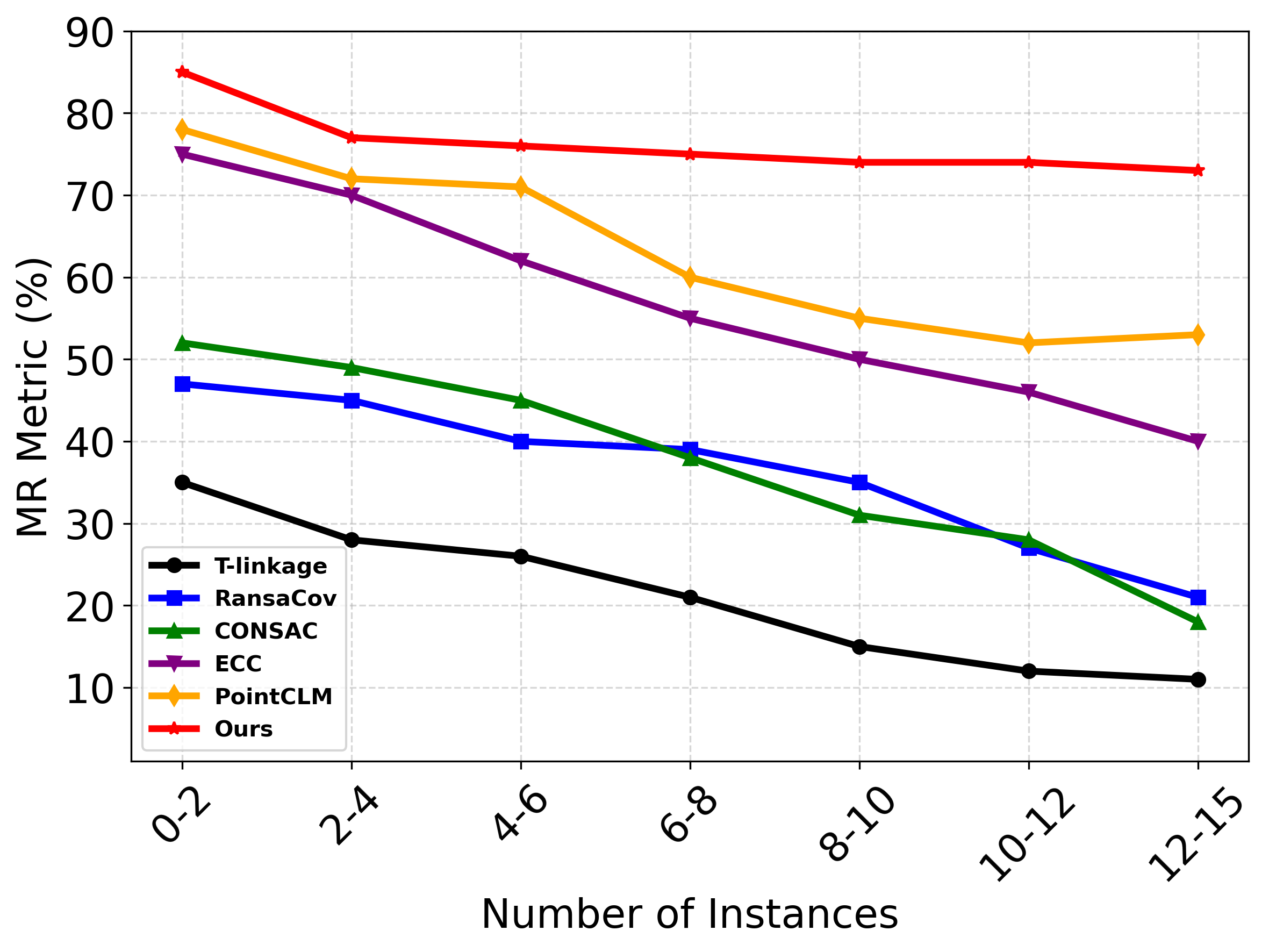}
		
		(a)
	\end{minipage}
\end{figure}
\vspace{-8mm}
\begin{figure}[H]
	\centering
	\begin{minipage}{0.7\textwidth}
		\centering
		\includegraphics[width=0.7\textwidth]{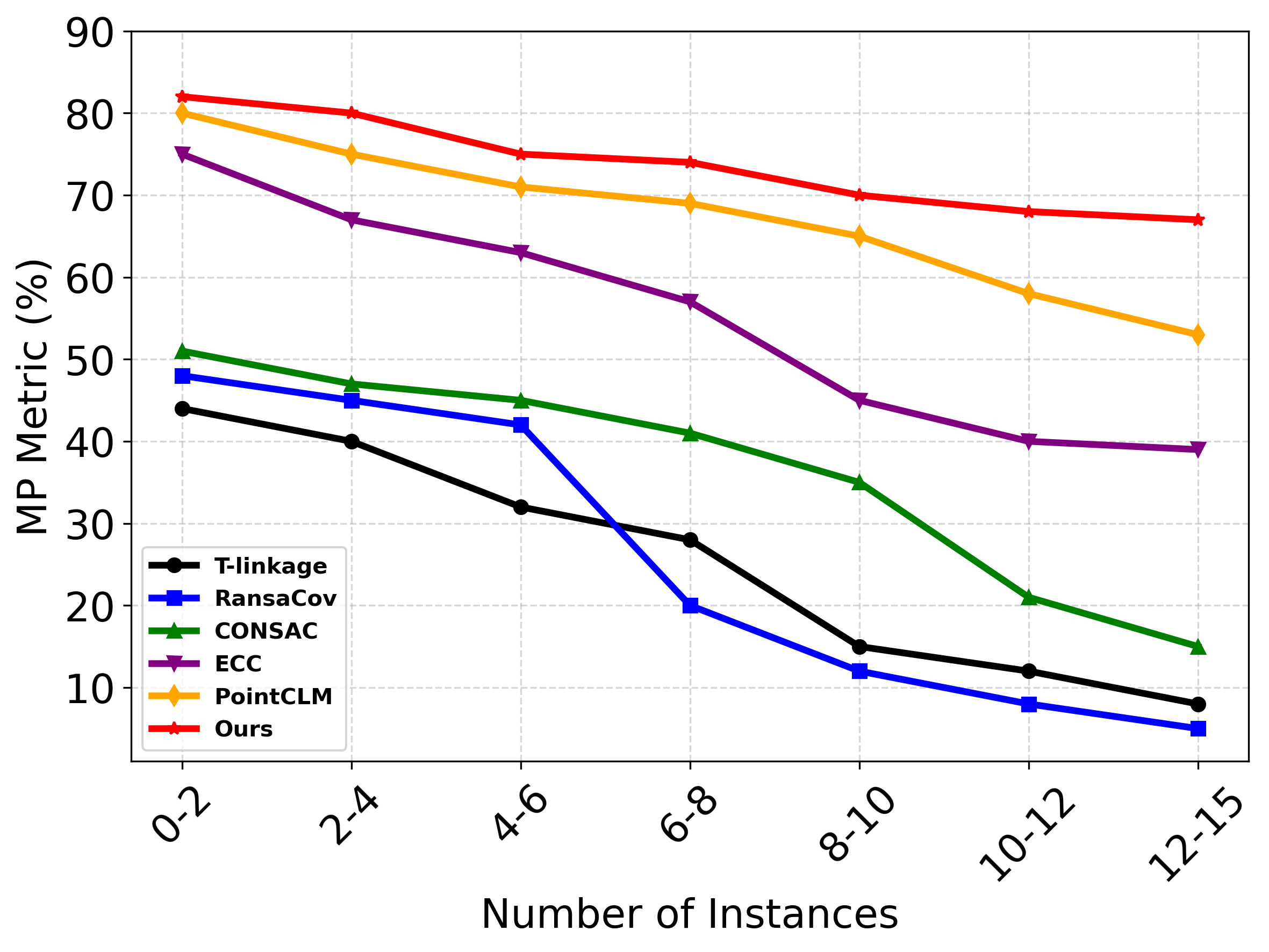}
		
		(b)
	\end{minipage}
\end{figure}
\vspace{-8mm}

\begin{figure}[H]
	\centering
	\includegraphics[width=0.8\textwidth]{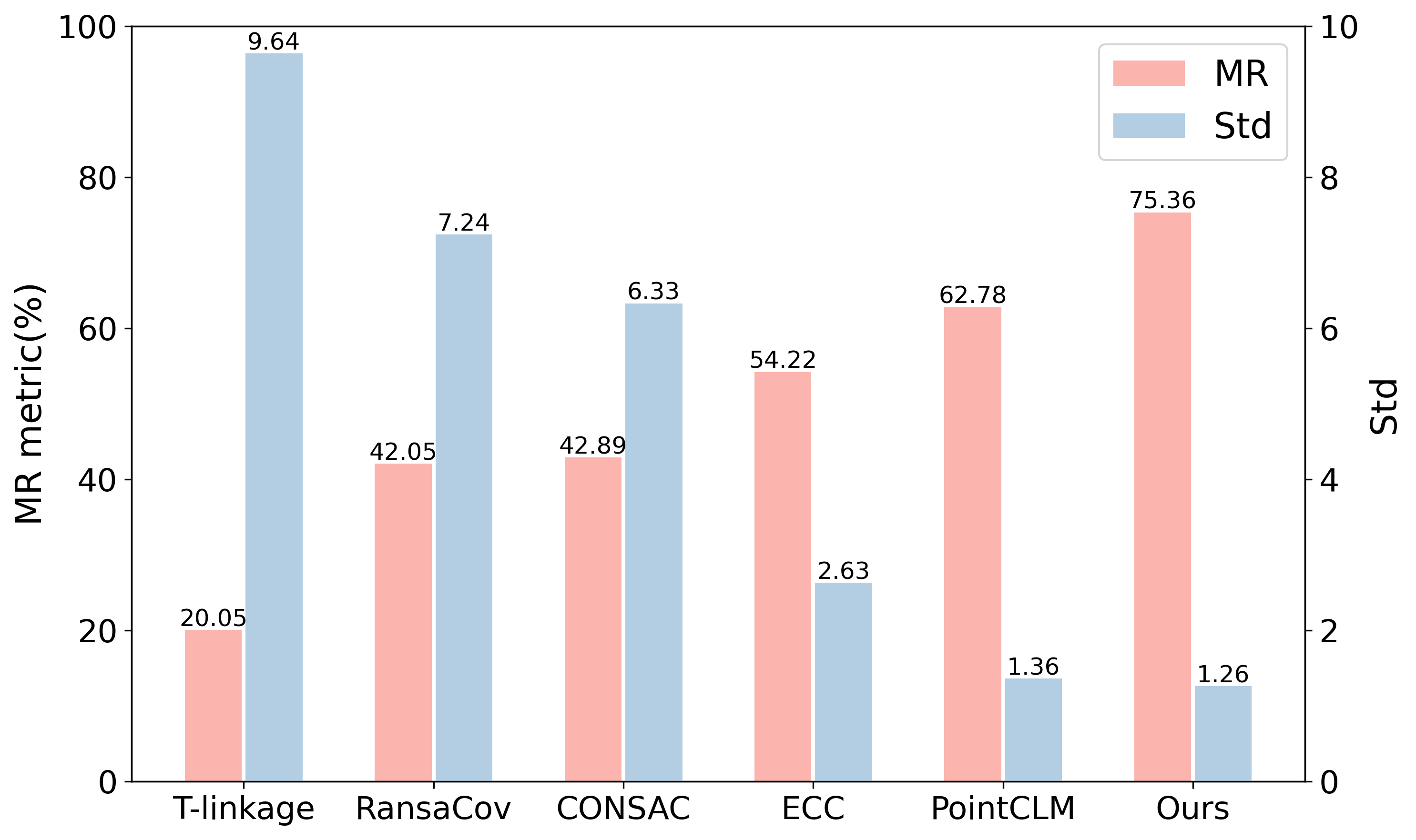}
	\caption{The MR and standard deviations of different methods on the Welding-Station dataset.}
	\label{tab:method_robustness}
\end{figure}

\textbf{Validate the robustness of our method.} To quantitatively assess the robustness of MRG, we randomly selected 100 scenes from the test set and conducted 15 experimental trials. The average recall rate and standard deviation of the five methods in welding scenes are presented in Fig. \ref{tab:method_robustness}, demonstrating that the proposed method achieves superior accuracy and stability.

\section{Conclusions}
\label{sec5}
By constructing high-precision digital models of industrial manufacturing scenes, achieving precise simulation of physical operational results has emerged as the mainstream method for accurate simulation in the development of the manufacturing industry. The advantages of simulation systems include accelerating the manufacturing process and reducing manufacturing costs. However, inconsistencies between the simulation and physical environments reduce the credibility of simulation results, limiting their utility in guiding actual production. Unlike traditional step-by-step "segmentation-registration" generation methods, this paper presents a novel Multi-Robot Manufacturing Digital Scene Generation (MRG) method for the first time. This method employs multi-instance point cloud registration to replace key robot instances in the scene with digital models while retaining point cloud data of other non-production-related operations, thereby enhancing the credibility of the simulation environment and improving generation efficiency, making it especially suitable for manufacturing scenes. Addressing the characteristics of industrial robots and manufacturing environments, an instance-focused Transformer module was developed to delineate instance boundaries and capture correlations between local regions. Additionally, an instance generation module was proposed to extract target instances while preserving key features. Finally, an efficient filtering and optimization algorithm was designed to refine the final registration results. Experimental evaluations on the Scan2CAD and Welding-Station datasets demonstrate that: 1) this method outperforms existing multi-instance point cloud registration techniques; 2) compared to state-of-the-art methods, the Scan2CAD dataset shows improvements of 12.15\% and 17.79\% in MR and MP, respectively. This work represents the first application of multi-instance point cloud registration in manufacturing scenes, significantly enhancing the accuracy and reliability of digital simulation environments for industrial applications.

\section*{Acknowledgements}
This work was supported by the Shanghai Pujiang Program (22PJD048) and the Open Research Project of the Guangxi Key Laboratory of Automobile Components and Vehicle Technology (2023GKLACVTKF05, 2023GKLACVTKF10). In addition, we would like to express our sincere gratitude to SAIC General Motors for generously providing the valuable data that significantly contributed to the success of this research. 
% references
%\begin{thebibliography}{00}
%% For numbered reference style
%\bibitem{lamport94}
%  \textit{\LaTeX: a document preparation system},
%  Addison Wesley, Massachusetts,
%\end{thebibliography}
\bibliographystyle{elsarticle-num}  % 指定参考文献样式
\bibliography{references}

\begin{thebibliography}{10}
\expandafter\ifx\csname url\endcsname\relax
  \def\url#1{\texttt{#1}}\fi
\expandafter\ifx\csname urlprefix\endcsname\relax\def\urlprefix{URL }\fi
\expandafter\ifx\csname href\endcsname\relax
  \def\href#1#2{#2} \def\path#1{#1}\fi

\bibitem{bib1}
B.~Zhou, R.~Zhou, Y.~Gan, F.~Fang, Y.~Mao, Multi-robot multi-station
  cooperative spot welding task allocation based on stepwise optimization: An
  industrial case study, Robotics and Computer-Integrated Manufacturing 73
  (2022) 102197.

\bibitem{bib2}
M.~Kamali, B.~Atazadeh, A.~Rajabifard, Y.~Chen, Advancements in 3d digital
  model generation for digital twins in industrial environments: Knowledge gaps
  and future directions, Advanced Engineering Informatics 62 (2024) 102929.

\bibitem{bib3}
Z.~Zhao, Z.~Zhang, Q.~Nie, C.~Liu, H.~Zhu, K.~Chen, D.~Tang, Probing a point
  cloud based expeditious approach with deep learning for constructing digital
  twin models in shopfloor, Advanced Engineering Informatics 62 (2024) 102748.

\bibitem{bib4}
Y.~Wu, Q.~Yao, X.~Fan, M.~Gong, W.~Ma, Q.~Miao, Panet: A point-attention based
  multi-scale feature fusion network for point cloud registration, IEEE
  Transactions on Instrumentation and Measurement 72 (2023) 1--13.

\bibitem{bib5}
D.~Cattaneo, M.~Vaghi, A.~Valada, Lcdnet: Deep loop closure detection and point
  cloud registration for lidar slam, IEEE Transactions on Robotics 38~(4)
  (2022) 2074--2093.

\bibitem{bib6}
J.~Hu, P.~R. Pagilla, View planning for object pose estimation using point
  clouds: An active robot perception approach, IEEE Robotics and Automation
  Letters 7~(4) (2022) 9248--9255.

\bibitem{bib7}
J.~Wang, Z.~Gong, B.~Tao, Z.~Yin, A 3-d reconstruction method for large
  freeform surfaces based on mobile robotic measurement and global
  optimization, IEEE Transactions on Instrumentation and Measurement 71 (2022)
  1--9.

\bibitem{bib8}
P.~Besl, N.~D. McKay, A method for registration of 3-d shapes, IEEE
  Transactions on Pattern Analysis and Machine Intelligence 14~(2) (1992)
  239--256.
\newblock \href {https://doi.org/10.1109/34.121791}
  {\path{doi:10.1109/34.121791}}.

\bibitem{bib9}
Q.-Y. Zhou, J.~Park, V.~Koltun, Fast global registration, in: Computer
  Vision--ECCV 2016: 14th European Conference, Amsterdam, The Netherlands,
  October 11-14, 2016, Proceedings, Part II 14, Springer, 2016, pp. 766--782.

\bibitem{bib10}
S.~Chen, H.~Ma, C.~Jiang, B.~Zhou, W.~Xue, Z.~Xiao, Q.~Li, Ndt-loam: A
  real-time lidar odometry and mapping with weighted ndt and lfa, IEEE Sensors
  Journal 22~(4) (2021) 3660--3671.

\bibitem{bib11}
H.~Courtois, N.~Aouf, K.~Ahiska, M.~Cecotti, Ndt rc: Normal distribution
  transform occupancy 3d mapping with recentering, IEEE Transactions on
  Intelligent Vehicles 9~(1) (2023) 2999--3009.

\bibitem{bib12}
Y.~Aoki, H.~Goforth, R.~A. Srivatsan, S.~Lucey, Pointnetlk: Robust \& efficient
  point cloud registration using pointnet, in: Proceedings of the IEEE/CVF
  conference on computer vision and pattern recognition, 2019, pp. 7163--7172.

\bibitem{bib13}
H.~Yu, F.~Li, M.~Saleh, B.~Busam, S.~Ilic, Cofinet: Reliable coarse-to-fine
  correspondences for robust pointcloud registration, Advances in Neural
  Information Processing Systems 34 (2021) 23872--23884.

\bibitem{bib14}
Z.~Qin, H.~Yu, C.~Wang, Y.~Guo, Y.~Peng, S.~Ilic, D.~Hu, K.~Xu, Geotransformer:
  Fast and robust point cloud registration with geometric transformer, IEEE
  Transactions on Pattern Analysis and Machine Intelligence 45~(8) (2023)
  9806--9821.

\bibitem{bib15}
L.~Magri, A.~Fusiello, Multiple model fitting as a set coverage problem, in:
  Proceedings of the IEEE conference on computer vision and pattern
  recognition, 2016, pp. 3318--3326.

\bibitem{bib16}
F.~Kluger, E.~Brachmann, H.~Ackermann, C.~Rother, M.~Y. Yang, B.~Rosenhahn,
  Consac: Robust multi-model fitting by conditional sample consensus, in:
  Proceedings of the IEEE/CVF conference on computer vision and pattern
  recognition, 2020, pp. 4634--4643.

\bibitem{bib17}
L.~Magri, A.~Fusiello, T-linkage: A continuous relaxation of j-linkage for
  multi-model fitting, in: Proceedings of the IEEE conference on computer
  vision and pattern recognition, 2014, pp. 3954--3961.

\bibitem{bib18}
S.~Huang, Z.~Gojcic, M.~Usvyatsov, A.~Wieser, K.~Schindler, Predator:
  Registration of 3d point clouds with low overlap, in: Proceedings of the
  IEEE/CVF Conference on computer vision and pattern recognition, 2021, pp.
  4267--4276.

\bibitem{bib19}
W.~Tang, D.~Zou, Multi-instance point cloud registration by efficient
  correspondence clustering, in: Proceedings of the IEEE/CVF conference on
  computer vision and pattern recognition, 2022, pp. 6667--6676.

\bibitem{bib20}
R.~B. Rusu, N.~Blodow, M.~Beetz, Fast point feature histograms (fpfh) for 3d
  registration, in: 2009 IEEE international conference on robotics and
  automation, IEEE, 2009, pp. 3212--3217.

\bibitem{bib21}
F.~Tombari, S.~Salti, L.~Di~Stefano, Unique signatures of histograms for local
  surface description, in: Computer Vision--ECCV 2010: 11th European Conference
  on Computer Vision, Heraklion, Crete, Greece, September 5-11, 2010,
  Proceedings, Part III 11, Springer, 2010, pp. 356--369.

\bibitem{bib22}
A.~Zeng, S.~Song, M.~Nie{\ss}ner, M.~Fisher, J.~Xiao, T.~Funkhouser, 3dmatch:
  Learning local geometric descriptors from rgb-d reconstructions, in:
  Proceedings of the IEEE conference on computer vision and pattern
  recognition, 2017, pp. 1802--1811.

\bibitem{bib23}
H.~Deng, T.~Birdal, S.~Ilic, Ppfnet: Global context aware local features for
  robust 3d point matching, in: Proceedings of the IEEE conference on computer
  vision and pattern recognition, 2018, pp. 195--205.

\bibitem{bib24}
C.~R. Qi, H.~Su, K.~Mo, L.~J. Guibas, Pointnet: Deep learning on point sets for
  3d classification and segmentation, in: Proceedings of the IEEE conference on
  computer vision and pattern recognition, 2017, pp. 652--660.

\bibitem{bib25}
Z.~Gojcic, C.~Zhou, J.~D. Wegner, A.~Wieser, The perfect match: 3d point cloud
  matching with smoothed densities, in: Proceedings of the IEEE/CVF conference
  on computer vision and pattern recognition, 2019, pp. 5545--5554.

\bibitem{bib26}
M.~A. Fischler, R.~C. Bolles, Random sample consensus: a paradigm for model
  fitting with applications to image analysis and automated cartography,
  Communications of the ACM 24~(6) (1981) 381--395.

\bibitem{bib27}
H.~M. Le, T.-T. Do, T.~Hoang, N.-M. Cheung, Sdrsac: Semidefinite-based
  randomized approach for robust point cloud registration without
  correspondences, in: Proceedings of the IEEE/CVF conference on computer
  vision and pattern recognition, 2019, pp. 124--133.

\bibitem{bib28}
D.~Barath, J.~Matas, Graph-cut ransac, in: Proceedings of the IEEE conference
  on computer vision and pattern recognition, 2018, pp. 6733--6741.

\bibitem{bib29}
Y.~Wu, X.~Hu, Y.~Zhang, M.~Gong, W.~Ma, Q.~Miao, Sacf-net: Skip-attention based
  correspondence filtering network for point cloud registration, IEEE
  Transactions on Circuits and Systems for Video Technology 33~(8) (2023)
  3585--3595.

\bibitem{bib30}
C.~Choy, W.~Dong, V.~Koltun, Deep global registration, in: Proceedings of the
  IEEE/CVF conference on computer vision and pattern recognition, 2020, pp.
  2514--2523.

\bibitem{bib31}
K.~S. Arun, T.~S. Huang, S.~D. Blostein, Least-squares fitting of two 3-d point
  sets, IEEE Transactions on pattern analysis and machine intelligence~(5)
  (1987) 698--700.

\bibitem{bib32}
Y.~Kanazawa, H.~Kawakami, Detection of planar regions with uncalibrated stereo
  using distributions of feature points., in: BMVC, Citeseer, 2004, pp. 1--10.

\bibitem{bib33}
M.~Yuan, Z.~Li, Q.~Jin, X.~Chen, M.~Wang, Pointclm: A contrastive
  learning-based framework for multi-instance point cloud registration, in:
  European Conference on Computer Vision, Springer, 2022, pp. 595--611.

\bibitem{bib34}
M.~Leordeanu, M.~Hebert, A spectral technique for correspondence problems using
  pairwise constraints, in: Tenth IEEE International Conference on Computer
  Vision (ICCV'05) Volume 1, Vol.~2, IEEE, 2005, pp. 1482--1489.

\bibitem{bib35}
H.~Thomas, C.~R. Qi, J.-E. Deschaud, B.~Marcotegui, F.~Goulette, L.~J. Guibas,
  Kpconv: Flexible and deformable convolution for point clouds, in: Proceedings
  of the IEEE/CVF international conference on computer vision, 2019, pp.
  6411--6420.

\bibitem{bib36}
Y.~Wang, J.~M. Solomon, Deep closest point: Learning representations for point
  cloud registration, in: Proceedings of the IEEE/CVF international conference
  on computer vision, 2019, pp. 3523--3532.

\bibitem{bib37}
C.~R. Qi, L.~Yi, H.~Su, L.~J. Guibas, Pointnet++: Deep hierarchical feature
  learning on point sets in a metric space, Advances in neural information
  processing systems 30 (2017).

\bibitem{bib38}
S.~Hinterstoisser, V.~Lepetit, S.~Ilic, S.~Holzer, G.~Bradski, K.~Konolige,
  N.~Navab, Model based training, detection and pose estimation of texture-less
  3d objects in heavily cluttered scenes, in: Computer Vision--ACCV 2012: 11th
  Asian Conference on Computer Vision, Daejeon, Korea, November 5-9, 2012,
  Revised Selected Papers, Part I 11, Springer, 2013, pp. 548--562.

\bibitem{bib39}
P.-E. Sarlin, D.~DeTone, T.~Malisiewicz, A.~Rabinovich, Superglue: Learning
  feature matching with graph neural networks, in: Proceedings of the IEEE/CVF
  conference on computer vision and pattern recognition, 2020, pp. 4938--4947.

\bibitem{bib40}
F.~Milletari, N.~Navab, S.-A. Ahmadi, V-net: Fully convolutional neural
  networks for volumetric medical image segmentation, in: 2016 fourth
  international conference on 3D vision (3DV), Ieee, 2016, pp. 565--571.

\bibitem{bib41}
A.~Avetisyan, M.~Dahnert, A.~Dai, M.~Savva, A.~X. Chang, M.~Nie{\ss}ner,
  Scan2cad: Learning cad model alignment in rgb-d scans, in: Proceedings of the
  IEEE/CVF Conference on computer vision and pattern recognition, 2019, pp.
  2614--2623.

\bibitem{bib42}
A.~Dai, A.~X. Chang, M.~Savva, M.~Halber, T.~Funkhouser, M.~Nie{\ss}ner,
  Scannet: Richly-annotated 3d reconstructions of indoor scenes, in:
  Proceedings of the IEEE conference on computer vision and pattern
  recognition, 2017, pp. 5828--5839.

\bibitem{bib43}
J.~Yang, Y.~Xiao, Z.~Cao, W.~Yang, Ranking 3d feature correspondences via
  consistency voting, Pattern Recognition Letters 117 (2019) 1--8.

\end{thebibliography}
\end{sloppypar}
\end{document}